\documentclass[11pt]{book} 
\usepackage{latexsym}
\usepackage{pdfpages}

\usepackage{amsmath}
\usepackage{varioref}

\usepackage{array}
\usepackage{colortbl}

\usepackage{mynatbib}
\usepackage{authorindex,multicol} 

\usepackage[T1]{fontenc}
\usepackage{palatino}

\usepackage{amsfonts,eucal,amsbsy,amsthm,amsopn,amssymb,amsmath}
\usepackage{bm}
\usepackage{url}
\usepackage{graphicx}
\usepackage{caption}

\usepackage{color}

\usepackage{pgfplots}

\usepackage{tikz-qtree}

\usepackage{makeidx}
\usepackage[font=small]{idxlayout}
\makeindex

 \renewenvironment{thebibliography}[1]{%
    \begin{oldthebibliography}{#1}%
      \setlength{\parskip}{0.4ex}%
      \setlength{\itemsep}{0ex}%
  }%
  {%
    \end{oldthebibliography}%
}

\renewcommand{\baselinestretch}{1.1}
\usepackage{tikz}
\usetikzlibrary{positioning,shadows,arrows}
\definecolor{lightblue}{rgb}{.8,.8,1}
\definecolor{mediumlightblue}{rgb}{.5,.5,1}
\definecolor{lightyellow}{rgb}{1,1,.5}
\definecolor{lightorange}{rgb}{1,.9,.7}
\definecolor{darkorange}{rgb}{1,.75,.2}
\definecolor{sortadarkorange}{rgb}{.8,.5,.1}
\definecolor{verydarkorange}{rgb}{.5,.3,0}
\definecolor{darkblue}{rgb}{0,0,0.8}
\definecolor{verydarkgreen}{rgb}{0,0.4,0}
\definecolor{darkgreen}{rgb}{0,0.8,0}
\definecolor{darkred}{rgb}{0.8,0,0}
\definecolor{verydarkred}{rgb}{0.6,0,0}
\definecolor{lightgreen}{rgb}{.8,1,.8}
\definecolor{lightred}{rgb}{1,.8,.8}
\definecolor{darkgrey}{rgb}{0.5,0.5,0.5}
\definecolor{purple}{rgb}{0.6,0,0.6}
\definecolor{red}{rgb}{1,0,0}
\definecolor{orange}{rgb}{.8,.6,0}
\definecolor{cyan}{rgb}{0,.6,.6}
\definecolor{reddishgreen}{rgb}{0.4,0.6,0}
\definecolor{lightgray}{rgb}{.8,.8,.8}
\usepackage{wrapfig}
\usepackage[framemethod=tikz]{mdframed}
\newenvironment{WrapText}[1][r]
  {\setlength{\intextsep}{0pt}%
\wrapfigure{#1}{0.5\textwidth}\vspace{-12pt}\mdframed[backgroundcolor=green!10,skipabove=0pt,skipbelow=0pt]}
  {\endmdframed\endwrapfigure}

\usepackage{amsfonts,eucal,amsbsy,amsthm,amsopn,amssymb}
\usepackage{xcolor}
\usepackage{titlesec}
\usepackage{hyperref}
\definecolor{TitleGreen}{rgb}{.104,.441,.253}
\definecolor{TitleLightGreen}{rgb}{0,.6,.3}
\titleformat
{\chapter}[display]{\bfseries\Huge\sffamily\color{TitleGreen}} 
{Chapter \thechapter} 
{0.5ex} 
{\vspace{0.5ex}} 
[\vspace{0.5ex}] 

\titleformat*{\section}{\Large\bfseries\sffamily\color{TitleGreen}}
\titleformat*{\subsection}{\large\bfseries\sffamily\color{TitleLightGreen}}
\titleformat*{\subsubsection}{\bfseries\sffamily\color{TitleLightGreen}}
\titleformat*{\paragraph}{\bfseries\sffamily\color{TitleGreen}}
\newcommand{\furtherreadings}[1]{{\small #1

}}

\newcommand{\NoteMarginal}[1]{}

\begin{document}
\title{{\huge \bf Statistical Machine Translation}\\[2cm]
{\LARGE
{\bf Draft of Chapter 13: Neural Machine Translation}\\[3cm] \phantom{.}}}
\author{{\LARGE \bf Philipp Koehn}\\[2mm]
{\LARGE Center for Speech and Language Processing}\\[2mm]
{\LARGE Department of Computer Science}\\[2mm]
{\LARGE Johns Hopkins University}\\[7mm]
1st public draft\\[-0.5mm]
August 7, 2015\\[5mm]
2nd public draft\\[-7mm]}

\maketitle

\tableofcontents

\setcounter{chapter}{12}
\chapter{Neural Machine Translation}

A major recent development in statistical machine translation is the adoption of neural networks. Neural network models promise better sharing of statistical evidence between similar words and inclusion of rich context. This chapter introduces several neural network modeling techniques and explains how they are applied to problems in machine translation

\section{A Short History}\index{history}
Already during the last wave of neural network research in the 1980s and 1990s, machine translation was in the sight of researchers exploring these methods \citep{janus}. In fact, the models proposed by \cite{forcada1997recursive} and \cite{castano-tmi-1997} are striking similar to the current dominant neural machine translation approaches.  However, none of these models were trained on data sizes large enough to produce reasonable results for anything but toy examples. 
The computational complexity involved by far exceeded the computational resources of that era, and hence the idea was abandoned for almost two decades. 

During this hibernation period, data-driven approaches such as phrase-based statistical machine translation rose from obscurity to dominance and made machine translation a useful tool for many applications, from information gisting to increasing the productivity of professional translators.

The modern resurrection of neural methods in machine translation started with the integration of neural language models into traditional statistical machine translation systems. The pioneering work by \cite{schwenk:csl:2007} showed large improvements in public evaluation campaigns. However, these ideas were only slowly adopted, mainly due to computational concerns. The use of GPUs for training also posed a challenge for many research groups that simply lacked such hardware or the experience to exploit it.

Moving beyond the use in language models, neural network methods crept into other components of traditional statistical machine translation, such as providing additional scores or extending translation tables \citep{schwenk:2012:POSTERS,lu-chen-xu:2014:P14-1}, reordering \citep{kanouchi-sudoh-komachi:2016:WAT2016,li-EtAl:2014:Coling3} and pre-ordering models \citep{degispert-iglesias-byrne:2015:NAACL-HLT}, and so on. For instance, the joint translation and language model by \cite{devlin-EtAl:2014:P14-1} was influential since it showed large quality improvements on top of a very competitive statistical machine translation system.

More ambitious efforts aimed at pure neural machine translation, abandoning existing statistical approaches completely. Early steps were the use of convolutional models \citep{kalchbrenner-blunsom:2013:EMNLP} and sequence-to-sequence models \citep{NIPS2014_5346,cho-EtAl:2014:SSST-8}. These were able to produce reasonable translations for short sentences, but fell apart with increasing sentence length. The addition of the attention mechanism finally yielded competitive results \citep{bahdanau:ICLR:2015,jean-EtAl:2015:WMT}. With a few more refinements, such as byte pair encoding and back-translation of target-side monolingual data, neural machine translation became the new state of the art.

Within a year or two, the entire research field of machine translation went neural. To give some indication of the speed of change: At the shared task for machine translation organized by the  Conference on Machine Translation (WMT), only one pure neural machine translation system was submitted in 2015. It was competitive, but outperformed by traditional statistical systems. A year later, in 2016, a neural machine translation system won in almost all language pairs. In 2017, almost all submissions were neural machine translation systems.

At the time of writing, neural machine translation research is progressing at rapid pace. There are many directions that are and will be explored in the coming years, ranging from core machine learning improvements such as deeper models to more linguistically informed models. More insight into the strength and weaknesses of neural machine translation is being gathered and will inform future work. 

There is an extensive proliferation of toolkits available for research, development, and deployment of neural machine translation systems. At the time of writing, the number of toolkits is multiplying, rather than consolidating. So, it is quite hard and premature to make specific recommendations. Nevertheless, some of the promising toolkits are:
\begin{itemize} \itemsep 0mm
\item Nematus (based on Theano): {\tt https://github.com/EdinburghNLP/nematus}
\item Marian (a C++ re-implementation of Nematus): {\tt https://marian-nmt.github.io/}
\item OpenNMT (based on Torch/pyTorch): {\tt http://opennmt.net/}
\item xnmt (based on DyNet): {\tt https://github.com/neulab/xnmt}
\item Sockeye (based on MXNet): {\tt https://github.com/awslabs/sockeye} 
\item T2T (based on Tensorflow): {\tt https://github.com/tensorflow/tensor2tensor}
\end{itemize}

\section{Introduction to Neural Networks}\label{sec:nn:introduction}
A neural network is a machine learning technique that takes a number of inputs and predicts outputs. In many ways, they are not very different from other machine learning methods but have distinct strengths.

\subsection{Linear Models}\index{linear model}
Linear models are a core element of statistical machine translation. A potential translation $x$ of a sentence is represented by a set of features $h_i(x)$. Each feature is weighted by a parameter $\lambda_i$ to obtain an overall score. Ignoring the exponential function that we used previously to turn the linear model into a log-linear model, the following formula sums up the model.

\begin{equation}
\text{score}(\lambda,x) = \sum_j \lambda_j \; h_j(x) 
\end{equation}

\begin{figure}
\begin{center}
\includegraphics[scale=1]{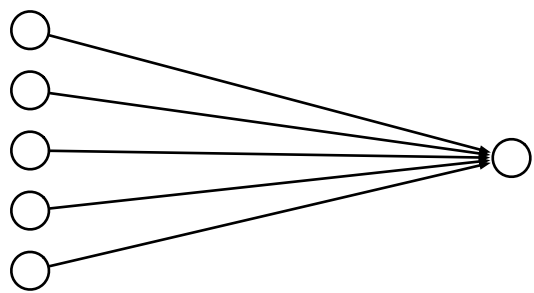}
\end{center}
\caption{Graphical illustration of a linear model as a network: feature values are input nodes, arrows are weights, and the score is an output node.}
\label{fig:nn:unlabelled-perceptron}
\end{figure}

Graphically, a linear model can be illustrated by a network, where feature values are input nodes, arrows are weights, and the score is an output node (see Figure~\ref{fig:nn:unlabelled-perceptron}).

Most prominently, we use linear models to combine different components of a machine translation system, such as the language model, the phrase translation model, the reordering model, and properties such as the length of the sentence, or the accumulated jump distance between phrase translations. Training methods assign a weight value $\lambda_i$ to each such feature $h_i(x)$, related to their importance in contributing to scoring better translations higher. In statistical machine translation, this is called {\bf tuning}\index{tuning}.

However, linear models do not allow us to define more complex relationships between the features. Let us say that we find that for short sentences the language model is less important than the translation model, or that average phrase translation probabilities higher than 0.1 are similarly reasonable but any value below that is really terrible. The first hypothetical example implies dependence between features and the second example implies non-linear relationship between the feature value and its impact on the final score. Linear models cannot handle these cases.

A commonly cited counter-example to the use of linear models is XOR, i.e., the boolean operator $\oplus$ with the truth table $0\oplus0=0$, $1\oplus0=1$, $0\oplus1=1$, and $1\oplus1=0$. For a linear model with two features (representing the inputs), it is not possible to come up with weights that give the correct output in all cases. Linear models assume that all instances, represented as points in the feature space, are linearly separable. This is not the case with XOR, and may not be the case for type of features we use in machine translation.

\subsection{Multiple Layers}
Neural networks modify linear models in two important ways. The first is the use of multiple layers. Instead of computing the output value directly from the input values, a {\bf hidden layer}\NoteMarginal{hidden layer}\index{hidden layer} is introduced. It is called hidden, because we can observe inputs and outputs in training instances, but not the mechanism that connects them --- this use of the concept {\em hidden} is similar to its meaning in hidden Markov models. 

\begin{figure}
\begin{center}
\includegraphics[scale=1]{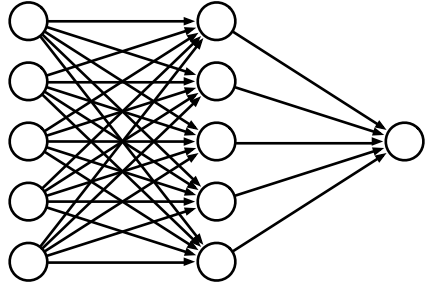}
\end{center}
\caption{A neural network with a hidden layer.}
\label{fig:nn:2-layer-perceptron}
\end{figure}

See Figure~\ref{fig:nn:2-layer-perceptron} for on illustration.  The network is processed in two steps. First, a linear combination of weighted input node is computed to produce each hidden node value. Then a linear combination of weighted hidden nodes is computed to produce each output node value.

At this point, let us introduce mathematical notations from the neural network literature. A neural network with a hidden layer consists of
\begin{itemize} \itemsep 0mm
\item a vector of input nodes with values $\vec{x} = (x_1,x_2,x_3,...x_n)^T$
\item a vector of hidden nodes with values $\vec{h} = (h_1,h_2,h_3,...h_m)^T$
\item a vector of output nodes with values $\vec{y} = (y_1,y_2,y_3,...y_l)^T$
\item a matrix of weights connecting input nodes with hidden nodes $W = \{ w_{ij} \}$
\item a matrix of weights connecting hidden nodes with output nodes $U = \{ u_{ij} \}$
\end{itemize}

The computations in a neural network with a hidden layer, as sketched out so far, are
\begin{eqnarray}
h_j = \sum_i x_i w_{ji}\\
y_k = \sum_j h_j u_{kj}
\end{eqnarray}

Note that we snuck in the possibility of multiple output nodes $y_k$, although our figures so far only showed one.

\subsection{Non-Linearity}
If we carefully think about the addition of a hidden layer, we realize that we have not gained anything so far to model input/output relationships. We can easily do away with the hidden layer by multiplying out the weights
\begin{equation}
\begin{split}
y_k & = \sum_j h_j u_{kj}\\
& = \sum_j \sum_i x_i w_{ji} u_{kj}\\
& = \sum_i x_i \left( \sum_j  u_{kj} w_{ji}  \right)
\end{split}
\end{equation}

Hence, a salient element of neural networks is the use of a {\bf non-linear activation function}\NoteMarginal{non-linear activation function}\index{activation function}\index{non-linear}. After computing the linear combination of weighted feature values $s_j = \sum_i x_i w_{ji}$, we obtain the value of a node only after applying such a function $h_j=f(s_j)$.

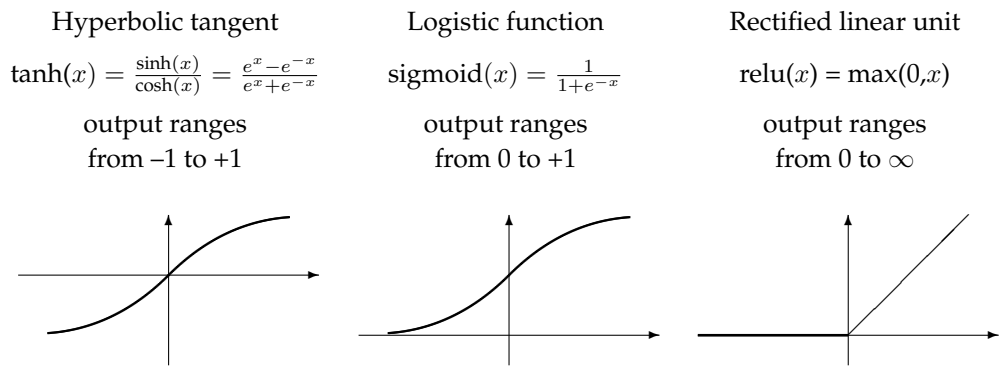
\begin{figure}
{\small
\begin{center}
\begin{tabular}{ccc}
Hyperbolic tangent & 
Logistic function & 
Rectified linear unit\\[2mm]

tanh($x)=\frac{\text{sinh}(x)}{\text{cosh}(x)}=\frac{e^x-e^{-x}}{e^x+e^{-x}}$ & 
sigmoid$(x) = \frac{1}{1+e^{-x}}$ &
relu($x$) = max(0,$x$)\\[2mm]

output ranges & 
output ranges &
output ranges \\

from --1 to +1 & 
from 0 to +1 &
from 0 to $\infty$\\[-7mm]

\setlength{\unitlength}{0.8cm}
\begin{picture}(5,5)(-2.5,-2.5)
\thinlines
\put(-2.5,0){\vector(1,0){5}}
\put(0,-1.5){\vector(0,1){2.5}}
\thicklines
\qbezier(0,0)(0.8853,0.8853)
(2,0.9640)
\qbezier(0,0)(-0.8853,-0.8853)
(-2,-0.9640)
\end{picture}
&
\setlength{\unitlength}{0.8cm}
\begin{picture}(5,5)(-2.5,-2.5)
\thinlines
\put(-2.5,-1){\vector(1,0){5}}
\put(0,-1.5){\vector(0,1){2.5}}
\thicklines
\qbezier(0,0)(0.8853,0.8853)
(2,0.9640)
\qbezier(0,0)(-0.8853,-0.8853)
(-2,-0.9640)
\end{picture}
&
\setlength{\unitlength}{0.8cm}
\begin{picture}(5,5)(-2.5,-2.5)
\linethickness{0.5pt}
\thinlines
\put(-2.5,-1){\vector(1,0){5}}
\put(0,-1.5){\vector(0,1){2.5}}
\linethickness{1pt}
\put(-2.5,-1){\line(1,0){2.5}}
\put(0,-1){\line(1,1){2}}
\end{picture}
\end{tabular}
\vspace{-1cm}
\end{center}
}
\caption{Typical activation functions in neural networks.}
\label{fig:nn:activation-functions}
\end{figure}

Popular choices are the {\bf hyperbolic tangent}\NoteMarginal{hyperbolic tangent}\index{hyperbolic tangent}\index{tangent} tanh(x) and the {\bf logistic function}\NoteMarginal{logistic function}\index{logistic function} sigmoid(x). See Figure~\ref{fig:nn:activation-functions} for more details on these functions. A good way to think about these activation functions is that they segment the range of values for the linear combination $s_j$ into
\begin{itemize} \itemsep 0mm
\item a segment where the node is turned off (values close to 0 for tanh, or --1 for sigmoid)
\item a transition segment where the node is partly turned on
\item a segment where the node is turned on (values close to 1)
\end{itemize}

A different popular choice is the activation function for the {\bf rectified linear unit}\NoteMarginal{rectified linear unit}\index{rectified linear unit} (ReLU)\index{ReLU}. It does not allow for negative values and floors them at 0, but does not alter the value of positive values. It is simpler and faster to compute than tanh($x$) or sigmoid($x$).

You could view each hidden node as a feature detector. For a certain configurations of input node values, it is turned on, for others it is turned off. Advocates of neural networks claim that the use of hidden nodes obviates (or at least drastically reduces) the need for feature engineering: Instead of manually detecting useful patterns in input values,  training of the hidden nodes discovers them automatically.

We do not have to stop at a single hidden layer. The currently fashionable name {\bf deep learning}\NoteMarginal{deep learning}\index{deep learning} for neural networks stems from the fact that often better performance can be achieved by deeply stacking together layers and layers of hidden nodes.

\subsection{Inference}\label{sec:nn:inference}
\begin{figure}
\begin{center}
\includegraphics[scale=1.5]{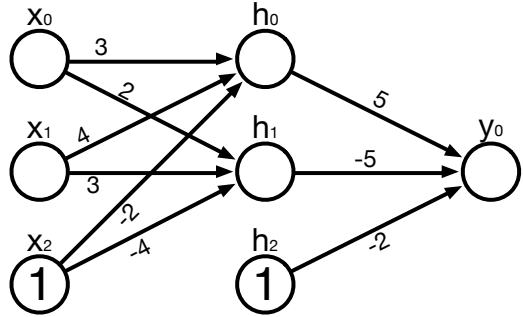}
\end{center}
\caption{A simple neural network with bias nodes in input and hidden layers.}
\label{fig:nn:example-feed-forward}
\end{figure}

Let us walk through neural network {\b inference}\index{inference} (i.e., how output values are computed from input values) with a concrete example. Consider the neural network in Figure~\ref{fig:nn:example-feed-forward}. This network has one additional innovation that we have not presented so far: bias units. These are nodes that always have the value 1. Such bias units give the network something to work with in the case that all input values are 0. Otherwise, the weighted sum $s_j$ would be 0 no matter the weights.

Let us use this neural network to process some input, say the value 1 for the first input node $x_0$ and 0 for the second input node $x_1$. The value of the bias input node (labelled $x_2$) is fixed to 1. To compute the value of the first hidden node $h_0$, we have to carry out the following calculation.

\begin{equation}
\begin{split}
h_0 &= \text{sigmoid}\left( \sum_i x_i w_{1i} \right)\\
&= \text{sigmoid}\left(  1 \times 3 + 0 \times 4 + 1 \times -2\right)\\
&= \text{sigmoid}\left(  1 \right)\\
&= 0.73
\end{split}
\end{equation}

The calculations for the other nodes are summarized in Table~\ref{tab:nn:example-feed-forward-calculations}. The output value in node $y_0$ for the input (0,1) is 0.743. If we expect binary output, we would understand this result as the value 1, since it is over the threshold of 0.5 in the range of possible output values [0;1]. 

\begin{table}
{\small
\begin{center}
\begin{tabular}{cclc}\hline\hline
Layer & Node & Summation & Activation\\ \hline
hidden & $h_0$ & $1 \times 3 + 0 \times 4 + 1 \times -2 = 1$ & 0.731\\
hidden & $h_1$ & $1 \times 2 + 0 \times 3 + 1 \times -4 = -2$ & 0.119\\
output & $y_0$ & $0.731 \times 5 + 0.119 \times -5 + 1 \times -2 = 1.060$ & 0.743\\\hline
\end{tabular}
\end{center}
}
\caption{Calculations for input (1,0) to the network in Figure~\ref{fig:nn:example-feed-forward}.}
\label{tab:nn:example-feed-forward-calculations}
\end{table}

Here, the output for all possible binary inputs:
{\small
\begin{center}
\begin{tabular}{cc|cc|c}
Input $x_0$ & Input $x_1$  & Hidden $h_0$  & Hidden $h_1$  & Output $y_0$ \\ \hline
0 & 0 & 0.119 & 0.018 & 0.183 $\rightarrow$ 0\\
0 & 1 & 0.881 & 0.269 & 0.743 $\rightarrow$ 1\\
1 & 0 & 0.731 & 0.119 & 0.743 $\rightarrow$ 1\\
1 & 1 & 0.993 & 0.731 & 0.334 $\rightarrow$ 0\\
\end{tabular}
\end{center}
}

Our neural network computes {\sc xor}. How does it do that? If we look at the hidden nodes $h_0$ and $h_1$, we notice that $h_0$ acts like the Boolean {\sc or}: Its value is high if at least of the two input values is 1 ($h_0$ = 0.881, 0.731, 0.993, for the three configurations),  it otherwise has a low value (0.119). The other hidden node $h_1$ acts like the Boolean {\sc and} --- it only has a high value (0.731) if both inputs are 1.
{\sc xor} is effectively implemented as the subtraction of the {\sc and} from the {\sc or} hidden node. 

Note that the non-linearity is key here. Since the value for the {\sc or} node $h_0$ is not that much higher for the input of (1,1) opposed to a single 1 in the input (0.993 vs. 0.881 and 0.731), the distinct high value for the {\sc and} node $h_1$ in this case (0.731) manages to push the final output $y_0$ below the threshold. This would not be possible if the values of the inputs would be simply summed up as in linear models.

As mentioned before, recently the use of the name {\bf deep learning} for neural networks has become fashionable. It emphasizes that often higher performance can be achieved by using networks with multiple hidden layers. Our {\sc xor} example hints at where this power comes from.  With a single input-output layer network it is possible to mimic basic Boolean operations such as {\sc and} and {\sc or} since they can be modeled with linear classifiers. {\sc xor} can be expressed as $x \; \text{\sc and} \; y - x \; \text{\sc or} \; y$, and our neural network example implements the Boolean operations {\sc and} and {\sc or} in the first layer, and the subtraction in the second layer. For functions that require more intricate computations, more operations may be chained together, and hence a neural network architecture with more hidden layers may be needed. It may be possible (with sufficient training data) to build neural networks for any computer program, if the number of hidden layers matches the depth of the computation. There is a line of research under the banner {\bf neural Turing machines}\index{neural Turing machines} that explores what kind of architectures are needed to implement basic algorithms \citep{DBLP:journals/corr/GemiciHSWMRAL17}. For instance, a neural network with two hidden layers is sufficient to implement an algorithm that sorts $n$-bit numbers.

\subsection{Back-Propagation Training}\label{sec:backprop}
Training neural networks requires the optimization of weight values so that the network predicts the correct output for a set of training examples. We repeatedly feed the input from the training examples into the network, compare the computed output of the network with the correct output from the training example, and update the weights. Typically, several passes over the training data are carried out. Each pass over the data is called an {\bf epoch}\index{epoch}.

The most common training method for neural networks is called {\bf back-propagation}\NoteMarginal{back-propagation}\index{back-propagation}, since it first updates the weights to the output layer, and propagates back error information to earlier layers. Whenever a training example is processed, then for each node in the network, an error term is computed which is the basis for updating the values for incoming weights.

The formulas used to compute updated values for weights follows principles of {\bf gradient descent}\index{gradient descent}\NoteMarginal{gradient descent} training. The error for a specific node is understood as a function of the incoming weights. To reduce the error given this function, we compute the gradient of the error function with respect to each of the weights, and move against the gradient to reduce the error.

\begin{figure}
\begin{center}
\includegraphics[width=10cm]{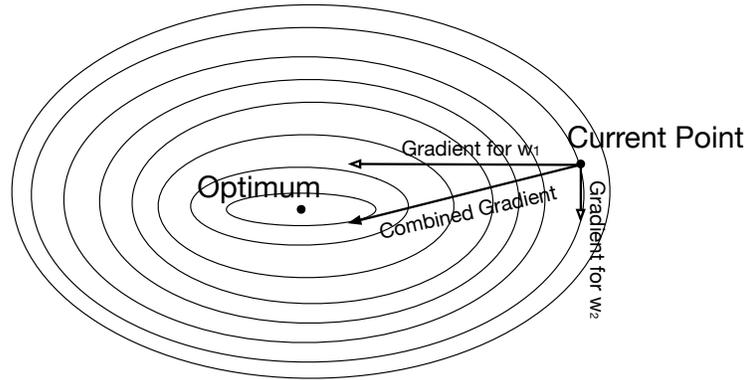}
\end{center}
\caption{Gradient descent training: We compute the gradient with regard to every dimension. In this case the gradient with respect to weight $w_2$ smaller than the gradient with respect to the weight $w_1$, so we move more to the left than down (note: arrows point in negative gradient direction, pointing to the minimum).}
\label{fig:nn:gradient-descent}
\end{figure}

Why is moving alongside the gradient a good idea? Consider that we optimize multiple dimensions at the same time. If you are looking for the lowest point in an area (maybe you are looking for water in a desert), and the ground falls off steep to the west of you, and also slightly south of you, then you would go in a direction that is mainly west --- and only slightly south. In other words, you go alongside the gradient. See Figure~\ref{fig:nn:gradient-descent} for an illustration.

In the following two sections, we will derive the formulae for updating weights for our example network. If you are less interested in the {\em why} and more in the {\em how}, you can skip these sections and continue reading when we summarize the update formulae \vpageref{sec:nn:training-summary}.

\subsubsection{Weights to the output nodes}
Let us first review and extend our notation. At an output node $y_i$, we first compute a linear combination of weight and hidden node values.
\begin{equation}
s_i=\sum_j w_{i\leftarrow j} h_j
\label{eqn:nn:s_i}
\end{equation}

This sum $s_i$ is passed through an activation function such as sigmoid to compute the output value $y$.
\begin{equation}
y_i = \text{sigmoid}(s_i)
\end{equation}

We compare the computed output values $y_i$ against the target output values $t_i$ from the training example. There are various ways to compute an error value $E$ from these values. Let us use the L2 norm.
\begin{equation}
E = \sum_i \frac{1}{2} (t_i-y_i)^2
\label{eqn:nn:l2-loss}
\end{equation}

As we stated above, our goal is to compute the gradient of the error $E$ with respect to the weights $w_k$ to find out in which direction (and how strongly) we should move the weight value. We do this for each weight $w_k$ separately. We first break up the computation of the gradient into three steps, essentially unfolding the Equations~\ref{eqn:nn:s_i} to~\ref{eqn:nn:l2-loss}.
\begin{equation}
\frac{d E}{d w_{i\leftarrow j}} = \frac{d E}{d y_i} \frac{d y_i}{d s_i} \frac{d s_i}{d w_{i\leftarrow j}} 
\label{eqn:dE_dw}
\end{equation}

Let us now work through each of these three steps.

\begin{itemize}
\item Since we defined the error $E$ in terms of the output values $y_i$, we can compute the first component as follows.
\begin{equation}
\frac{d E}{d y_i} = \frac{d}{d y_i} \; \frac{1}{2} (t_i-y_i)^2 = - (t_i-y_i)
\label{eqn:nn:dE_dy}
\end{equation}

\item The derivative of the output value $y_i$ with respect to $s_i$ (the linear combination of weight and hidden node values)  depends on the activation function. In the case of sigmoid, we have:
\begin{equation}
\frac{d y_i}{d s_i} = \frac{d \; \text{sigmoid}(s_i)}{d s_i} = \text{sigmoid}(s_i) ( 1-\text{sigmoid}(s_i)) = y_i(1-y_i)
\label{eqn:nn:dy_ds}
\end{equation}

To keep our treatment below as general as possible and not commit to the sigmoid as an activation function, we will use the shorthand $y'_i$ for $\frac{d y_i}{d s_i}$ below. Note that for any given training example and any given differentiable activation function, this value can always be computed.

\item Finally, we compute the derivative of $s_i$ with respect to the weight $w_{i\leftarrow j}$, which turns out to be quite simply the value to the hidden node $h_j$.
\begin{equation}
\frac{d s}{d w_{i\leftarrow j}} = \frac{d}{d w_{i\leftarrow j}} \sum_j w_{i\leftarrow j} h_j = h_j
\label{eqn:ds_dw}
\end{equation}
\end{itemize}

Where are we? In Equations~\ref{eqn:nn:dE_dy} to \ref{eqn:ds_dw}, we computed the three steps needed to compute the gradient for the error function given the unfolded laid out in Equation~\ref{eqn:dE_dw}.
Putting it all together, we have
\begin{equation}
\begin{split}
\frac{d E}{d w_{i\leftarrow j}} & = \frac{d E}{d y_i} \frac{d y_i}{d s_i} \frac{d s}{d w_{i\leftarrow j}} \\
& = - (t_i-y_i) \;\;\; y'_i \;\;\; h_j
\end{split}
\end{equation}

Factoring in a {\bf learning rate}\index{learning rate} $\mu$ gives us the following update formula for weight $w_{i\leftarrow j}$. Note that we also remove the minus sign, since we move against the gradient towards the minimum.

\begin{equation*}
\Delta w_{i\leftarrow j} = \mu (t_i-y_i) \; y_i' \; h_j
\end{equation*}

It is useful to introduce the concept of an {\bf error term}\NoteMarginal{error term}\index{error term} $\delta_i$. Note that this term is associated with a node, while the weight updates concern weights. The error term has to be computed only once for the node, and it can be then used for each of the incoming weights.
\begin{equation}
\delta_i = (t_i-y_i) \; y'_i
\end{equation}

This reduces the update formula to:
\begin{equation}
\Delta w_{i\leftarrow j} = \mu \; \delta_i \; h_j
\end{equation}

\subsubsection{Weights to the hidden nodes} 
The computation of the gradient and hence the update formula for hidden nodes is quite analogous.
As before, we first define the linear combination $z_j$ (previously $s_i$) of input values $x_k$ (previously hidden values $h_j$) weighted by weights $u_{j\leftarrow k}$ (previously weights $w_{i\leftarrow j}$).
\begin{equation}
z_j=\sum_k u_{j\leftarrow k} x_k
\end{equation}

This leads to the computation of the value of the hidden node $h_j$.
\begin{equation}
h_j = \text{sigmoid}(z_j)
\end{equation}

Following the principles of gradient descent, we need to compute the derivative of the error $E$ with respect to the weights $u_{j\leftarrow k}$. We decompose this derivative as before.
\begin{equation}
\frac{d E}{d u_{j\leftarrow k}} = \frac{d E}{d h_j} \frac{d h_j}{d z_j} \frac{d z_j}{d u_{j\leftarrow k}} 
\label{eqn:nn:dE_du}
\end{equation}

However, the computation of $\frac{d E}{d h_j}$ is more complex than in the case of output nodes, since the error is defined in terms of output values $y_i$, not values for hidden nodes $h_j$. The idea behind back-propagation is to track how the error caused by the hidden node contributed to the error in the next layer.  Applying the chain rule gives us:
\begin{equation}
\frac{d E}{d h_j} = \sum_i \frac{d E}{d y_i} \frac{d y_i}{d s_i} \frac{d s_i}{d h_j} 
\label{eqn:nn:dE_dh}
\end{equation}

We already encountered the first two terms $\frac{d E}{d y_i}$  (Equation~\ref{eqn:nn:dE_dy}) and $\frac{d y_i}{d s_i}$ (Equation~\ref{eqn:nn:dy_ds}) previously. To recap:
\begin{equation}
\begin{split}
\frac{d E}{d y_i} \; \frac{d y_i}{d s_i} &= \frac{d}{d y_i} \; \sum_{i'} \frac{1}{2} (t_i-y_{i'})^2 \;\;\; y'_i\\
&=  \frac{d}{d y_i} \frac{1}{2} (t_i-y_i)^2 \;\;\; y'_i\\
&= - (t_i-y_i) \;\;\; y'_i\\
& = \delta_i
\end{split}
\label{eqn:nn:dE_dy_ds}
\end{equation}

The third term in  Equation~\ref{eqn:nn:dE_dh} is computed straightforward.
\begin{equation}
\frac{d s_i}{d h_j} = \frac{d}{d h_j} \sum_i w_{i\leftarrow j} h_j = w_{i\leftarrow j}
\label{eqn:nn:ds_dh}
\end{equation}

Putting Equation~\ref{eqn:nn:dE_dy_ds} and Equation~\ref{eqn:nn:ds_dh} together, Equation~\ref{eqn:nn:dE_dh} can be solved as:
\begin{equation}
\frac{d E}{d h_j} = \sum_i \delta_i w_{i\leftarrow j} 
\label{eqn:nn:dE_dh_solved}
\end{equation}

This gives rise to a quite intuitive interpretation. The error that matters at the hidden node $h_j$ depends on the error terms $\delta_i$ in the subsequent nodes $y_i$, weighted by $w_{i\leftarrow j}$, i.e., the impact the hidden node $h_j$ has on the output node $y_i$.

Let us tie up the remaining loose ends.
The missing pieces from Equation~\ref{eqn:nn:dE_du} are the second term
\begin{equation}
\frac{d h_j}{d z_j} =  \frac{d \; \text{sigmoid}(z_j)}{d z_j} = \text{sigmoid}(z_j) ( 1-\text{sigmoid}(z_j)) = h_j(1-h_j) = h'_j
\label{eqn:nn:dh_dz}
\end{equation}

and third term
\begin{equation}
\frac{d z_j}{d u_{j\leftarrow k}} = \frac{d}{d u_{j\leftarrow k}} \sum_k u_{j\leftarrow k} x_k = x_k
\label{eqn:nn:dz_du}
\end{equation}

Putting Equation~\ref{eqn:nn:dE_dh_solved}, Equation~\ref{eqn:nn:dh_dz}, and Equation~\ref{eqn:nn:dz_du} together gives us the gradient 
\begin{equation}
\begin{split}
\frac{d E}{d u_{j\leftarrow k}} &= \frac{d E}{d h_j} \frac{d h_j}{d z_j} \frac{d z_j}{d u_{j\leftarrow k}}\\
&= \sum_i \left( \delta_i w_{i\leftarrow j} \right) \;\; h'_j \;\; x_k
\end{split}
\end{equation}

If we define an error term $\delta_j$ for hidden nodes analogous to output nodes
\begin{equation}
\delta_j =  \sum_i \left( \delta_i w_{i\leftarrow j} \right) \;\; h'_j
\end{equation}

then we have an analogous update formula
\begin{equation}
\Delta u_{j\leftarrow k} = \mu \; \delta_j \; x_k
\end{equation}

\subsubsection{Summary}\label{sec:nn:training-summary}
We train neural networks by processing training examples, one at a time, and update weights each time. What drives weight updates is the gradient towards a smaller error. Weight updates are computed based on error terms $\delta_i$ associated with each non-input node in the network.

For output nodes, the error term $\delta_i$ is computed from the actual output $y_i$ of the node for our current network, and the target output $t_i$ for the node.
\begin{equation}
\delta_i = (t_i-y_i) \; y'_i
\end{equation}

For hidden nodes, the error term $\delta_j$ is computed via back-propagating the error term $\delta_i$ from subsequent nodes connected by weights $w_{i\leftarrow j}$.
\begin{equation}
\delta_j =  \sum_i \left( \delta_i w_{i\leftarrow j} \right) \; h'_j
\end{equation}

Computing $y'_i$ and $h'_j$ requires the derivative of the activation function, to which the weighted sum of incoming values is passed.

Given the error terms, weights $w_{i\leftarrow j}$ (or $u_{j\leftarrow k}$) from each proceeding node $h_j$ (or $x_k$) are updated, tempered by a learning rate $\mu$.
\begin{equation}
\begin{split}
\Delta w_{i\leftarrow j} &= \mu \; \delta_i \; h_j\\
\Delta u_{j\leftarrow k} &= \mu \; \delta_j \; x_k
\end{split}
\label{eqn:weight-update-summary}
\end{equation}

Once weights are updated, the next training example is processed. There are typically several passes over the training set, called epochs.

\subsubsection{Example} 
Given the neural network in Figure~\ref{fig:nn:example-feed-forward}, let us see how the training example (1,0) $\rightarrow$ 1 is processed. 

Let us start with the calculation of the error term $\delta$ for the output node $y_0$. During inference (recall Table~\vref{tab:nn:example-feed-forward-calculations}), we computed the linear combination of weighted hidden node values $s_0=1.060$ and the node value $y_0=0.743$. The target value is $t_0=1$.
\begin{equation}
\delta = (t_0-y_0) \; y'_0 = (1-0.743) \times \text{sigmoid}'(1.060) = 0.257 \times 0.191 = 0.049 
\end{equation}

With this number, we can compute weight updates, such as for weight $w_{0 \leftarrow 0}$.
\begin{equation}
\Delta w_{0 \leftarrow 0} = \mu \; \delta_0 \; h_0 = \mu \times 0.049 \times 0.731 = \mu \times 0.036
\end{equation}

Since the hidden node $h_0$ leads only to one output node $y_0$, the calculation of its error term $\delta_0$ is not more computationally complex.
\begin{equation}
\delta_j =  \sum_i \left( \delta_i u_{i\leftarrow 0} \right) \; h'_0 = \left( \delta \times w_{0\leftarrow 0} \right) \times \text{sigmoid}'(z_0) = 0.049 \times 5 \times 0.197 = 0.049
\end{equation}

Table~\ref{tab:nn:example-updates} summarizes the updates for all weights.

\begin{table}
{\small
\begin{center} \vspace{-0cm}
\includegraphics[scale=1.3]{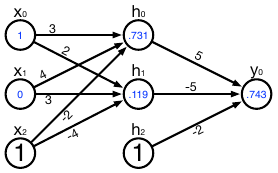}

\vspace{2mm}
\begin{tabular}{cll}\hline
Node & Error term & Weight updates \\ \hline \hline
& $\delta = (t_0-y_0) \; \text{sigmoid}(s_0)$ & $\Delta w_{0 \leftarrow j} = \mu \; \delta \; h_j$ \\ \hline
$y_0$ & $\delta = (1-0.743) \times 0.191 = 0.049$ & $\Delta w_{0\leftarrow 0} = \mu \times 0.049 \times 0.731 = 0.036$\\
&&$\Delta w_{0\leftarrow 1} = \mu \times 0.049 \times 0.119 = 0.006$\\
&&$\Delta w_{0\leftarrow 2} = \mu \times 0.049 \times 1 = 0.049$\\ \hline \hline
& $\delta_j = \delta \; w_{i\leftarrow j} \; \text{sigmoid}(z_j)$ & $\Delta u_{j \leftarrow i} = \mu \; \delta_j \; x_i$ \\ \hline
$h_0$ & $\delta_0 = 0.049 \times 5 \times 0.197 = 0.048$ & $\Delta u_{0\leftarrow 0} = \mu \times 0.048 \times 1 = 0.048$\\
&&$\Delta u_{0\leftarrow 1} = \mu \times 0.048 \times 0 = 0$\\
&&$\Delta u_{0\leftarrow 2} = \mu \times 0.048 \times 1  = 0.048$\\ \hline
$h_1$ & $\delta_1 = 0.049 \times -5 \times 0.105 = -0.026$ & $\Delta u_{1\leftarrow 0} = \mu \times -0.026 \times 1 = -0.026$\\
&&$\Delta u_{1\leftarrow 1} = \mu \times -0.026 \times 0  = 0$\\
&&$\Delta u_{1\leftarrow 2} = \mu \times -0.026 \times 1  = -0.026$\\
\hline\end{tabular}
\end{center}
}
\caption{Weight updates (with unspecified learning rate $\mu$) for the neural network in Figure~\ref{fig:nn:example-feed-forward} (repeated above the table) when the training example (1,0) $\rightarrow$ 1 is presented.}
\label{tab:nn:example-updates}
\end{table}

\subsection{Refinements}\label{sec:gradient-descent-refinements}

\begin{figure}
\begin{center}
\begin{tabular}{ccc}
\includegraphics[scale=0.86]{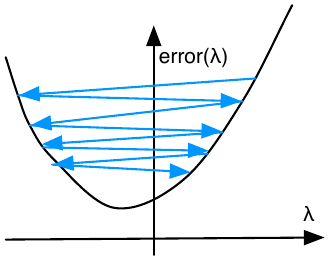} &
\includegraphics[scale=0.86]{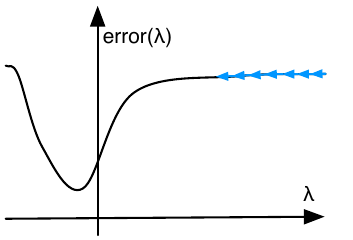} &
\includegraphics[scale=0.86]{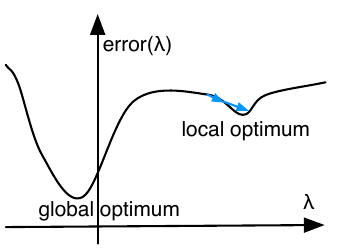}\\
Too high learning rate & Bad initialization & Local optimum
\end{tabular}
\end{center}
\caption{Problems with gradient descent training that motivate some of the refinements detailed in Section~\ref{sec:gradient-descent-refinements}: (a) a too high learning rate may lead to too drastic parameter updates, overshooting the optimum, (b) bad initialization may require many updates to escape a plateau, and (c) the existence of local optima which trap training.}
\label{fig:gradient-descent-problem}
\end{figure}

We conclude our introduction to neural networks with some basic refinements and considerations. 
To motivate some of the refinements, consider Figure~\ref{fig:gradient-descent-problem}. While gradient descent training is a fine idea, it may run into practical problems.
\begin{itemize}
\item Setting the learning rate too high leads to updates that overshoot the optimum. Conversely, a too low learning rate leads to slow convergence.
\item Bad initialization of weights may lead to long paths of many update steps to reach the optimum. This is especially a problem with activation functions like sigmoid which only have a short interval of significant change.
\item The existence of local optima lead the search to get trapped and miss the global optimum.
\end{itemize}

\subsubsection{Validation Set}
Neural network training proceeds for several epochs, i.e., full iterations over the training data. When to stop? 
When we track training progress, we see that the error on the training set continuously decreases. However, at some point {\bf over-fitting}\index{over-fitting} sets in, where the training data is memorized and not sufficiently generalized.

\begin{figure}
\begin{center}
\includegraphics[scale=1]{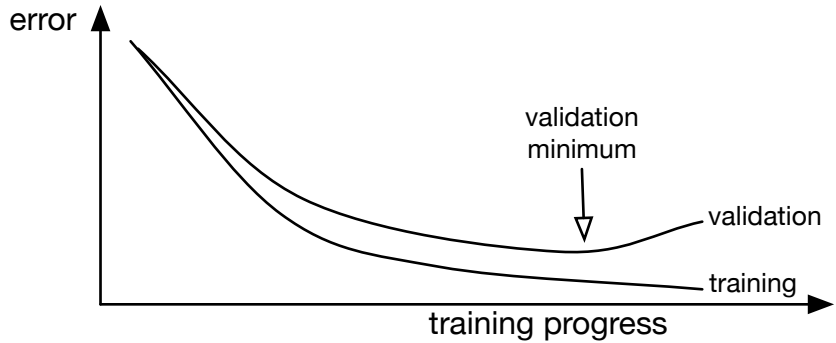}
\end{center}
\caption{Training progress over time. The error on the training set continuously decreases. However, on a validation set (not used for training), at some point the error increases. Training is stopped at the validation minimum before such over-fitting sets in.}
\label{fig:nn-training-validation}
\end{figure}

We can check this with an additional set of examples, called the {\bf validation set}\index{validation set}, that is not used during training. See 
Figure~\ref{fig:nn-training-validation} for an illustration. When we measure the error on the validation set at each point of training, we see that at some point this error increases. Hence, we stop training, when the minimum on the validation set is reached.

\subsubsection{Weight Initialization}\index{initialization}
Before training starts, weights are initialized to random values. The values are choses from a uniform distribution. We prefer  initial weights that lead to node values that are in the transition area for the activation function, and not in the low or high shallow slope where it would take a long time to push towards a change. For instance, for the sigmoid activation function, feeding values in the range of, say, $[-1;1]$ to the activation function leads to activation values in the range of [0.269;0.731].

For the sigmoid activation function, commonly used formula for weights to the final layer of a network are
\begin{equation}
\big[ - \frac{1}{\sqrt{n}},\frac{1}{\sqrt{n}} \big]
\end{equation}
where $n$ is the size of the previous layer. For hidden layers, we chose weights from the range
\begin{equation}
\big[ - \frac{\sqrt{6}}{\sqrt{n_j+n_{j+1}}},\frac{\sqrt{6}}{\sqrt{n_j+n_{j+1}}} \big]
\end{equation}
where $n_j$ is the size of the previous layer, $n_j$ size of next layer.

\subsubsection{Momentum Term}
Consider the case where a weight value is far from its optimum.
Even if most training examples push the weight value in the same direction, it may still take a while for each of these small updates to accumulate until the weight reaches its optimum. A common trick is to use a {\bf momentum term}\index{momentum term}\NoteMarginal{momentum term} to speed up training. 
This momentum term $m_t$ gets updated at each time step $t$ (i.e., for each training example). We combine the previous value of the momentum term $m_{t-1}$ with the current raw weight update value $\Delta w_t$ and use the resulting momentum term value to update the weights.

For instance, with a decay rate of 0.9, the update formula changes to
\begin{equation}
\begin{split}
m_t & = 0.9 m_{t-1} + \Delta w_t\\
w_t & = w_{t-1} - \mu \; m_t
\end{split}
\label{eqn:momentum}
\end{equation}

\subsubsection{Adapting Learning Rate per Parameter}
A common training strategy is to reduce the learning rate $\mu$ over time. At the beginning the parameters are far away from optimal values and have to change a lot, but in later training stages we are concerned with fine tuning, and a large learning rate may cause a parameter to bounce around an optimum. 

But different parameters may be at different stages on the path to their optimal values, so a different learning rate for each parameter may be helpful. One such method, called {\bf Adagrad}\NoteMarginal{Adagrad}\index{Adagrad}, records the gradients that were computed for each parameter and accumulates their square values over time, and uses this sum to adjust the learning rate.

The Adagrad update formula is based on the sum of gradients of the error $E$ with respect to the weight $w$ at all time steps $t$, i.e.,  $g_t = \frac{d E_t}{d w}$. We divide the learning rate $\mu$ for this weight this accumulated sum.
\begin{equation}
\Delta w_t = \frac{\mu}{\sqrt{\sum_{\tau=1}^{t} g_\tau^2}} \; g_t
\end{equation}

Intutively, big changes in the parameter value (corresponding to big gradients $g_t$), lead to a reduction of the learning rate of the weight parameter.

Combining the idea of momentum term and adjusting parameter update by their accumulated change is the inspiration of {\bf Adam}\index{Adam}, another method to transform the raw gradient into a parameter update.

First, there is the idea of momentum, which is computed as in Equation~\ref{eqn:momentum} above. 
\begin{equation}
m_t = \beta_1 m_{t-1} + (1-\beta_1)g_t
\end{equation}

Then, there is the idea of the squares of gradients (as in Adagrad) for adjusting the learning rate. Since raw accumulation does run the risk of becoming too large and hence permanently depressing the learning rate, Adam uses exponential decay, just like for the momentum term.
\begin{equation}
v_t = \beta_2 v_{t-1} + (1-\beta_2)g_t^2
\end{equation}

The hyper parameters $\beta_1$ and $\beta_2$ are set typically close to 1, but this also means that early in training the values for $m_t$ and $v_t$ are close to their initialization values of 0. To adjust for that, they are corrected for this bias.
\begin{equation}
\hat{m}_t = \frac{m_t}{1 - \beta^t_1},  \hspace{1cm}
\hat{v}_t = \frac{v_t}{1 - \beta^t_2} 
\end{equation}

With increasing training time steps $t$, this correction goes away: $\text{lim}_{t\rightarrow\infty}\frac{1}{1-\beta^t} \rightarrow 1$.

Having these pieces in hand (learning rate $\mu$, momentum $\hat{m}_t$, accumulated change $\hat{v}_t$), weight update per Adam is computed as
\begin{equation}
\Delta w_t = \frac{\mu}{\sqrt{\hat{v}_t} + \epsilon} \hat{m}_t
\end{equation}

Common values for the hyper parameters are $\beta_1=0.9$, $\beta_2=0.999$ and $\epsilon=10^{-8}$.

There are various other adaptation schemes. This is an active area of research. For instance, the second order gradient gives some useful information about the rate of change. However, it is often expensive to compute, so other shortcuts are taken.

\subsubsection{Dropout}
The parameter space in which back-propagation learning and its variants are operating is littered with local optima. The hill-climbing algorithm may just climb a mole hill and be stuck there, instead of moving towards a climb of the highest mountain. 
Various methods have been proposed to get training out of these local optima. One currently popular method in neural machine translation is called {\bf drop-out}\index{drop-out}. 

It sounds a bit simplistic and wacky. During training, some of the nodes of the neural network are ignored. Their values are set to 0, and their associated parameters are not updated. These dropped-out nodes are chosen at random, and may account for as much as 10\%, 20\% or even more of all the nodes. Training resumes for some number of iterations without the nodes, and then a different set of drop-out nodes are selected.

The dropped-out nodes played some useful role in the model trained up to the point where they are ignored. After that, other nodes have to pick up the slack. The end result is a more robust model where several nodes share similar roles.

\subsubsection{Layer Normalization}\label{sec:layer-normalization}\index{layer normalization} 
Layer normalization addresses a problem that arises especially in the deep neural networks that we are using in neural machine translation, where computing proceeds through a large sequence of layers. For some training examples, average values at one layer may become very large, which feed into the following layer, also producing large output values, and so on. This is especially a problem with activation functions that do not limit the output to a narrow interval, such as rectified linear units. For other training examples the average values at the same layers may be very small. 
This causes a problem for training. Recall from Equation~\ref{eqn:weight-update-summary}, that gradient updates are strongly effected by node values. Too large node values lead to exploding gradients and too small node values lead to diminishing gradients. 

To remedy this, the idea is to normalize the values on a per-layer basis. This is done by adding additional computational steps to the neural network. Recall that a feed-forward layer consists of the the matrix multiplication of the weight matrix $W$ with the node values from the previous layer $\vec{h^{l-1}}$, resulting in a weighted sum $\vec{s^l}$, followed by an activation function such as sigmoid.
\begin{equation}
\begin{aligned}
\vec{s^l} &= W \; \vec{h^{l-1}}\\
\vec{h^l} &= \text{sigmoid}(\vec{h^l})
\end{aligned}
\end{equation}

We can compute the mean and variance of the values in the weighted sum vector $\vec{s^l}$ by
\begin{equation}
\begin{aligned}
\mu^l &= \frac{1}{H} \sum_{i-1}^{H} s_i^l\\
\sigma^l &= \sqrt{\frac{1}{H} \sum_{i-1}^{H} (s_i^l-\mu^l)^2}
\end{aligned}
\end{equation}

Using these values, we normalize the vector $\vec{s^l}$ using two additional bias vectors $\vec{g}$ and $\vec{b}$
\begin{equation}
\vec{\hat{s^l}} = \frac{\vec{g}}{\sigma^l} \cdot (\vec{s^l} - \mu^l) + \vec{b}
\end{equation}
where $\cdot$ is element-wise multiplication and the difference subtracts the scalar average from each vector element.

The formula first normalizes the values in $\vec{s^l}$ by shifting them against their average value, hence ensuring that their average afterwards is 0. The resulting vector is then divided by the variance $\sigma^l$. The additional bias vectors give some flexibility, they may be shared across multiple layers of the same type, such as multiple time steps in a recurrent neural network (we will introduce these in Section~\vref{sec:rnn}).

\subsubsection{Mini Batches}\label{sec:nn:mini-batch}
Each training example yields a set of weight updates $\Delta w_i$. We may first process all the training examples and only afterwards apply all the updates. But neural networks have the advantage that they can immediately learn from each training example. A training method that updates the model with each training example is called {\bf online learning}\index{online learning}\index{learning!online}\NoteMarginal{online learning}. The online learning variant of gradient descent training is called {\bf stochastic gradient descent}\NoteMarginal{stochastic gradient descent}\index{stochastic gradient descent}. 

Online learning generally takes fewer passes over the training set (called {\bf epochs}\NoteMarginal{epoch})  for convergence.
However, since training constantly changes the weights, it is hard to parallelize. So, instead, we may want to process the training data in batches, accumulate the weight updates, and then apply them collectively. These smaller sets of training examples are called {\bf mini batches}\NoteMarginal{mini batch}\index{mini batch} to distinguish this approach from {\bf batch training}\NoteMarginal{batch training}\index{batch training} where the entire training set is considered one batch.

There are other variations to organize the processing of the training set, typically motivated by restrictions of parallel processing. If we process the training data in mini batches, then we can parallelize the computation of weight update values $\Delta w$, but have to synchronize their summation and application to the weights. If we want to distribute training over a number of machines, it is computationally more convenient to break up the training data in equally sized parts, perform online learning for each of the parts (optionally using smaller mini batches), and then average the weights.
Surprisingly, breaking up training this way, often leads to better results than straightforward linear processing.

Finally, a scheme called {\bf Hogwild}\index{Hogwild} runs several training threads that immediately update weights, even though other threads still use the weight values to compute gradients. While this is clearly violates the safe guards typically taken in parallel programming, it does not hurt in practical experience.




\subsubsection{Vector and Matrix Operations}\index{vector}\index{matrix}
We can express the calculations needed for handling neural networks as vector and matrix operations.
\begin{itemize} \itemsep 0mm
\item Forward computation: $\vec{s} = W \; \vec{h}$
\item Activation function:  $\vec{y} = \text{sigmoid}(\vec{h})$
\item Error term: $\vec{\delta} = (\vec{t}-\vec{y}) \; \text{sigmoid'}(\vec{s})$
\item Propagation of error term: $\vec{\delta}_i = W \; \vec{\delta}_{i+1} \cdot \text{sigmoid'}(\vec{s}) $
\item Weight updates: $\Delta W = \mu \; \vec{\delta} \; \vec{h}^T$
\end{itemize}

Executing these operations is computationally expensive. If our layers have, say, 200 nodes, then the matrix operation $W \vec{h}$ requires $200\times200=40,000$ multiplications. Such matrix operations are also common in another highly used area of computer science: graphics processing. When rendering images on the screen, the geometric properties of 3-dimensional objects have to be processed to generate the color values of the 2-dimensional image on the screen. Since there is high demand for fast graphics processing, for instance for the use in realistic looking computer games, specialized hardware has become commonplace: {\bf graphics processing units (GPUs)}\NoteMarginal{graphics processing unit}\index{GPU}\index{graphics processing unit}.

These processors have a massive number of cores (for example, the NVIDIA GTX 1080ti GPU provides 3584 thread processors) but a rather lightweight instruction set. GPUs provide instructions that are applied to many data points at once, which is exactly what is needed out the vector space computations listed above. Programming for GPUs is supported by various libraries, such as CUDA for C++, and has become an essential part of developing large scale neural network applications.

The general term for scalars, vectors, and matrices is {\bf tensors}\index{tensor}. A tensor may also have more dimensions: a sequence of matrices can be packed into a 3-dimensional tensor. Such large objects are actually frequently used in today's neural network toolkits.

\paragraph{Further Readings}
\furtherreadings{A good introduction to modern neural network research is the textbook ''Deep Learning'' \citep{Goodfellow-et-al-2016}. There is also book on neural network methods applied to the natural language processing in general \citep{Goldberg17}. 

A number of key techniques that have been recently developed have entered the standard repertoire of neural machine translation research. Training is made more robust by methods such as drop-out \citep{JMLR:v15:srivastava14a}, where during training intervals a number of nodes are randomly masked. To avoid exploding or vanishing gradients during back-propagation over several layers, gradients are typically clipped \citep{DBLP:conf/icml/PascanuMB13}. Layer normalization \citep{2016arXiv160706450L} has similar motivations, by ensuring that node values are within reasonable bounds.

An active topic of research are optimization methods that adjust the learning rate of gradient descent training. Popular methods are Adagrad \citep{duchi2011adaptive}, Adadelta \citep{DBLP:journals/corr/abs-1212-5701}, and currently Adam \citep{ICLR:2015:KingmaBa}.}

\section{Computation Graphs}\label{sec:computation-graphs}
For our example neural network from Section~\ref{sec:backprop}, we painstakingly worked out derivates for gradient computations needed by gradient descent training. After all this hard work, it may come as surprise that you will likely never have to do this again. It can be done automatically, even for arbitrarily complex neural network architectures. There are a number of toolkits that allow you to define the network and it will take care of the rest. In this section, we will take a close look at how this works.

\subsection{Neural Networks as Computation Graphs}
First, we will take a different look at the networks we are building. We previously represented neural networks as graphs consisting of nodes and their connections (recall Figure~\vref{fig:nn:example-feed-forward}), or by mathematical equations such as 
\begin{eqnarray} 
\begin{split}
h = \text{sigmoid} ( W_1 x + b_1 )\\
y =  \text{sigmoid} ( W_2 h + b_2 )
\end{split}
\end{eqnarray}

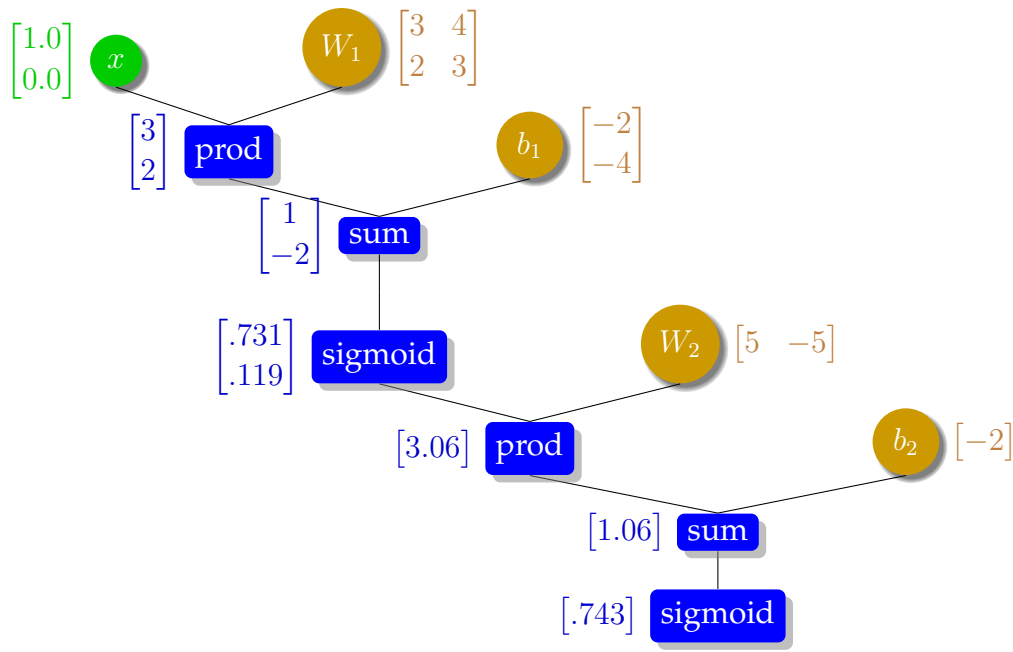
\begin{figure}
\begin{center}
\large
\begin{tikzpicture}[grow = up,
    computation/.style={rectangle, draw=none, rounded corners=1mm, fill=blue, drop shadow,
        text centered, anchor=south, text=white},
    derivative/.style={rectangle, draw=none, rounded corners=1mm, fill=purple, drop shadow,
        text centered, anchor=south, text=white},
    parameter/.style={circle, draw=none, fill=orange, circular drop shadow,
        text centered, anchor=south, text=white},
    data/.style={circle, draw=none, fill=darkgreen, circular drop shadow,
        text centered, anchor=south, text=white},
    level distance=0.5cm, growth parent anchor=north
]
\node (sigmoid2) [computation] {sigmoid} [-]
    child{ [sibling distance=5cm]
        node (sum2) [computation] {sum}
            child { 
                node (b2) [parameter] {$b_2$}
            }
            child{ [sibling distance=4cm]
                node (prod2) [computation] {prod}
                child{
                    node (w2) [parameter] {$W_2$}
                }
                child{ [sibling distance=4cm]
                    node (sigmoid1) [computation] {sigmoid}
                    child{
                        child{
                            node (sum1) [computation] {sum}
                                child {
                                    node (b1) [parameter] {$b_1$}
                                }
                            child{ [sibling distance=3cm]
                                node (prod1) [computation] {prod}
                                child{
                                    node (w1) [parameter] {$W_1$}
                                }
                                child{
                                    node (x) [data] {$x$}
                                }
                            }
                       }
                   }
              }
         }
     };
     
     \node [right=of w1,xshift=-1cm] {\textcolor{brown}{$\begin{bmatrix} 3 & 4 \\ 2 & 3 \end{bmatrix}$}};
     \node [right=of w2,xshift=-1cm] {\textcolor{brown}{$\begin{bmatrix} 5 & -5 \end{bmatrix}$}};
     \node [right=of b1,xshift=-1cm] {\textcolor{brown}{$\begin{bmatrix} -2 \\ -4 \end{bmatrix}$}};
     \node [right=of b2,xshift=-1cm] {\textcolor{brown}{$\begin{bmatrix} -2 \end{bmatrix}$}};   

     \node [left=of x,xshift=1cm] {\textcolor{darkgreen}{$\begin{bmatrix} 1.0 \\ 0.0 \end{bmatrix}$}};
     \node [left=of prod1,xshift=1cm] {\textcolor{darkblue}{$\begin{bmatrix} 3 \\ 2 \end{bmatrix}$}};
     \node [left=of sum1,xshift=1cm] {\textcolor{darkblue}{$\begin{bmatrix} 1 \\ -2 \end{bmatrix}$}};
     \node [left=of sigmoid1,xshift=1cm] {\textcolor{darkblue}{$\begin{bmatrix} .731 \\ .119 \end{bmatrix}$}};
     \node [left=of prod2,xshift=1cm] {\textcolor{darkblue}{$\begin{bmatrix} 3.06 \end{bmatrix}$}};
     \node [left=of sum2,xshift=1cm] {\textcolor{darkblue}{$\begin{bmatrix} 1.06 \end{bmatrix}$}};
     \node [left=of sigmoid2,xshift=1cm] {\textcolor{darkblue}{$\begin{bmatrix} .743 \end{bmatrix}$}};

\end{tikzpicture}
\end{center}
\caption{Two layer feed-forward neural network as a computation graph, consisting of the input value $x$, weight parameters $W_1$, $W_2$, $b_1$, $b_2$, and computation nodes (product, sum, sigmoid). To the right of each parameter node, its value is shown. To the left of input and computation nodes, we show how the input $(1,0)^T$ is processed by the graph.}
\label{fig:computation-graph}
\end{figure}

The equations above describe the feed-forward neural network that we use as our running example. We now represent this math in form of a {\bf computation graph}\index{computation graph}. See Figure~\ref{fig:computation-graph} for an illustration of the computation graph for our network. The graph contains as nodes the parameters of the models (the weight matrices $W_1$, $W_2$ and bias vectors $b_1$, $b_2$), the input $x$ and the mathematical operations that are carried out between them (product, sum, and sigmoid). Next to each parameter, we show their values.

Neural networks, viewed as computation graphs, are any arbitrary connected operations between an input and any number of parameters. Some of these operations may have little to do with any inspiration from neurons in the brain, so we are stretching the term {\em neural networks} quite a bit here. The graph does not have to have a nice tree structure as in our example, but may be any {\bf acyclical directed graph}\index{acyclical directed graph}, i.e., anything goes as long there is a straightforward processing direction and no cycles. Another way to view such a graph is as a fancy way to visualize a sequence of function calls that take as arguments the input, parameters, previously computed values, or any combination thereof, but have no recursion or loops.

Processing an input with the neural network requires placing the input value into the node $x$ and carrying out the computations. In the figure, we show this with the input vector $(1,0)^T$. The resulting numbers should look familiar since they are the same as when previously worked through this example in Section~\ref{sec:nn:inference}.

Before we move on, let us take stock of what each {\bf computation node}\index{computation node} in the graph has to accomplish. It consists of the following:
\begin{itemize} \itemsep 0mm \vspace{-3mm}
\item a function that executes its computation operation
\item links to input nodes
\item when processing an example, the computed value
\end{itemize}

We will add two more items to each node in the following section.

\subsection{Gradient Computations}
So far, we showed how the computation graph can be used process an input value. Now we will examine how it can be used to vastly simply model training. Model training requires an error function and the computation of gradients to derive update rules for parameters.

The first of these is quite straightforward. To compute the error, we need to add another computation at the end of the computation graph. This computation takes the computed output value $y$ and the given correct output value $t$ from the training data and produces an error value. A typical error function is the L2 norm $\frac{1}{2}(t-y)^2$. From the view of training, the result of the execution of the computation graph is an error value.
  
\begin{WrapText}
{\bf Calculus Refresher}\\
In calculus, the {\bf chain rule}\index{chain rule} is a formula for computing the derivative of the composition of two or more functions. That is, if f and g are functions, then the chain rule expresses the derivative of their composition $f \cdot g$ (the function which maps $x$ to $f(g(x))$) in terms of the derivatives of $f$ and $g$ and the product of functions as follows:

$(f\circ g)'=(f'\circ g)\cdot g'.$

This can be written more explicitly in terms of the variable. Let $F = f \cdot g$, or equivalently, $F(x) = f(g(x))$ for all $x$. Then one can also write $F'(x)=f'(g(x))g'(x)$

The chain rule may be written, in Leibniz's notation, in the following way. If a variable z depends on the variable $y$, which itself depends on the variable $x$, so that $y$ and $z$ are therefore dependent variables, then $z$, via the intermediate variable of $y$, depends on $x$ as well. The chain rule then states,

$\frac{dz}{dx}=\frac {dz}{dy} \cdot \frac {dy}{dx}$

The two versions of the chain rule are related, if $z=f(y)$ and $y=g(x)$, then

$\frac {dz}{dx}= \frac {dz}{dy} \cdot \frac {dy}{dx} = f'(y) g'(x)=f'(g(x))g'(x)$

\begin{flushright}
{\small (adapted from Wikipedia)}
\end{flushright}
\end{WrapText}

Now, for the more difficult part --- devising update rules for the parameters. Looking at the computation graph, model updates originate from the error values and propagate back to the model parameter. Hence, we call the computations needed to compute the update values also the {\bf backward pass}\index{backward pass} through the graph, opposed to the {\bf forward pass}\index{forward pass} that computed output and error.

Consider the chain of operations that connect the weight matrix $W_2$ to the error computation.
\begin{equation}
\begin{split}
e &= \text{L2}(y,t)\\
y &= \text{sigmoid}(s)\\
s &= \text{sum}(p,b_2)\\
p &= \text{prod}(h,W_2)
\end{split}
\end{equation}
where $h$ are the values of the hidden layer nodes, resulting from earlier computations. 

To compute the update rule for the parameter matrix $W_2$, we view the error as a function of these parameters and take the derivative with respect to them, in our case $\frac{d L2(W_2)}{d W_2}$. Recall that when we computed this derivate we first broke it up into steps using the chain rule. We now do the same here.

\begin{equation}
\frac{d \text{L2}(W_2)}{d W_2} = \frac{d \text{L2}(\text{sigmoid}(\text{sum}(\text{prod}(h,W_2),b_2)),t)}{d W_2} = \frac{d \text{L2}(y,t)}{d y} \frac{d \text{sigmoid}(s)}{d s} \frac{d \text{sum}(p,b_2)}{d p} \frac{d \text{prod}(h,W_2)}{d W_2}
\end{equation}

Note that for the purpose for computing an update rule for $W_2$, we treat all the other variables in this computation (the  target value $t$, the bias vector $b_2$, the hidden node values $h$) as constants.  This breaks up the derivative of the error with respect to the parameters $W_2$ into a chain of derivatives along the line of the nodes of the computation graph.

Hence, all we have to do for gradient computations is to come up with derivatives for each node in the computation graph. In our example these are 
\begin{equation}
\begin{split}
\frac{d \text{L2}(y,t)}{d y} &= \frac{d \frac{1}{2} (t-y)^2}{d y} = t-y\\
\frac{d \text{sigmoid}(s)}{d s} &= \text{sigmoid}(s) (1-\text{sigmoid}(s)) \\
\frac{d \text{sum}(p,b_2)}{d p} &= \frac{d p+b_2}{d p} = 1\\
\frac{d \text{prod}(h,W_2)}{d W_2} &= \frac{d W_2 h}{d W_2} = h
\end{split}
\end{equation}

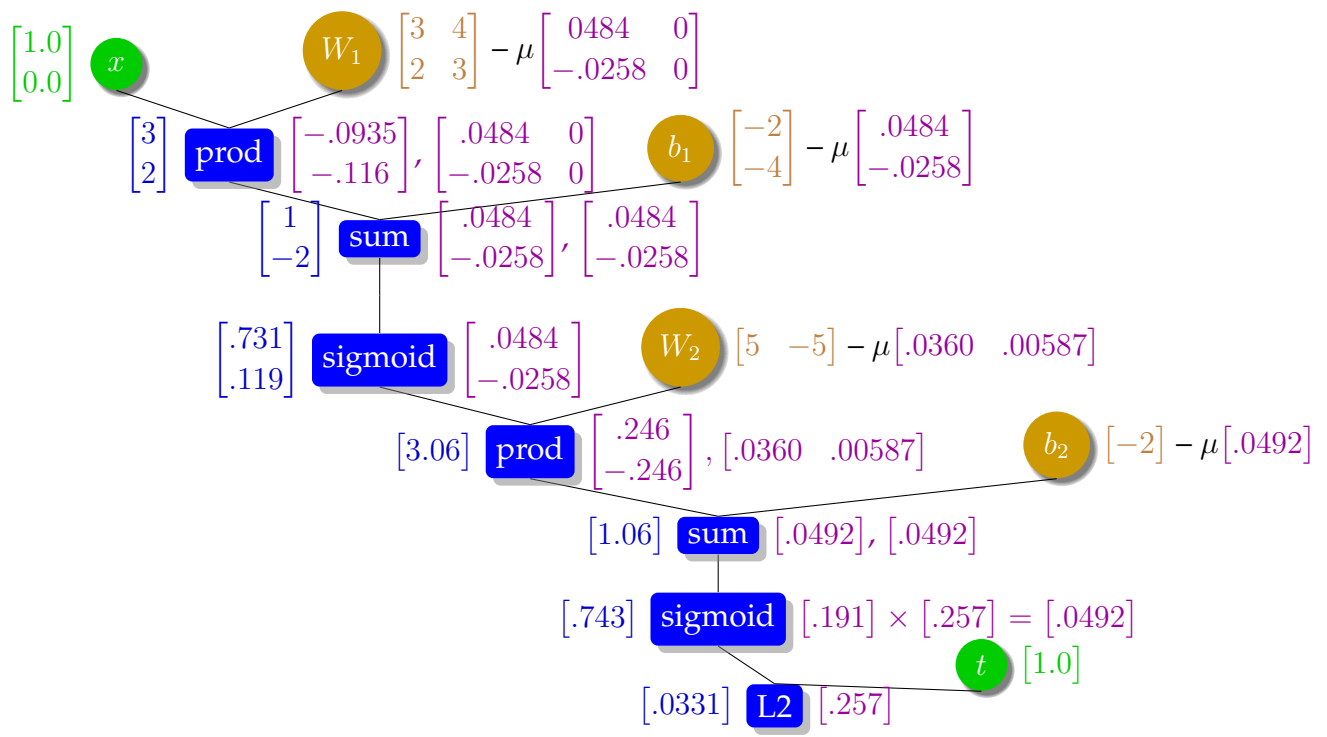
\begin{figure}
\begin{center}
\large
\begin{tikzpicture}[grow = up,
    computation/.style={rectangle, draw=none, rounded corners=1mm, fill=blue, drop shadow,
        text centered, anchor=south, text=white},
    derivative/.style={rectangle, draw=none, rounded corners=1mm, fill=purple, drop shadow,
        text centered, anchor=south, text=white},
    parameter/.style={circle, draw=none, fill=orange, circular drop shadow,
        text centered, anchor=south, text=white},
    data/.style={circle, draw=none, fill=darkgreen, circular drop shadow,
        text centered, anchor=south, text=white},
    level distance=0.5cm, growth parent anchor=north
]
\node (l2)  [computation,sibling distance=8cm] {L2} 
            child { 
                node (t) [data,yshift=-.6cm,xshift=2cm] {$t$}
            }
    child{ [sibling distance=6.5cm]
        node (sigmoid2) [computation]  {sigmoid}
    child{ [sibling distance=5cm]
        node (sum2) [computation] {sum}
            child { 
                node (b2) [parameter,xshift=2cm] {$b_2$}
            }
            child{ [sibling distance=4cm]
                node (prod2) [computation] {prod}
                child{
                    node (w2) [parameter] {$W_2$}
                }
                child{ [sibling distance=4cm]
                    node (sigmoid1) [computation] {sigmoid}
                    child{
                        child{
                            node (sum1) [computation] {sum}
                                child {
                                    node (b1) [parameter,xshift=2cm] {$b_1$}
                                }
                            child{ [sibling distance=3cm]
                                node (prod1) [computation] {prod}
                                child{
                                    node (w1) [parameter] {$W_1$}
                                }
                                child{
                                    node (x) [data] {$x$}
                                }
                            }
                       }
                   }
              }
         }}
     };

     \node [right=of w1,xshift=-1cm] {\textcolor{brown}{$\begin{bmatrix} 3 & 4 \\ 2 & 3 \end{bmatrix}$} -- $\mu$\textcolor{purple}{$\begin{bmatrix} 0484 & 0 \\ -.0258 & 0  \end{bmatrix}$}};
     \node [right=of w2,xshift=-1cm] {\textcolor{brown}{$\begin{bmatrix} 5 & -5 \end{bmatrix}$} -- $\mu$\textcolor{purple}{$\begin{bmatrix}.0360 & .00587  \end{bmatrix}$}};
     \node [right=of b1,xshift=-1cm] {\textcolor{brown}{$\begin{bmatrix} -2 \\ -4 \end{bmatrix}$} -- $\mu$\textcolor{purple}{$\begin{bmatrix} .0484 \\ -.0258 \end{bmatrix}$}};
     \node [right=of b2,xshift=-1cm] {\textcolor{brown}{$\begin{bmatrix} -2 \end{bmatrix}$} -- $\mu$\textcolor{purple}{$\begin{bmatrix} .0492 \end{bmatrix}$}};  

     \node [left=of x,xshift=1cm] {\textcolor{darkgreen}{$\begin{bmatrix} 1.0 \\ 0.0 \end{bmatrix}$}};
     \node [left=of prod1,xshift=1cm] {\textcolor{darkblue}{$\begin{bmatrix} 3 \\ 2 \end{bmatrix}$}};
     \node [left=of sum1,xshift=1cm] {\textcolor{darkblue}{$\begin{bmatrix} 1 \\ -2 \end{bmatrix}$}};
     \node [left=of sigmoid1,xshift=1cm] {\textcolor{darkblue}{$\begin{bmatrix} .731 \\ .119 \end{bmatrix}$}};
     \node [left=of prod2,xshift=1cm] {\textcolor{darkblue}{$\begin{bmatrix} 3.06 \end{bmatrix}$}};
     \node [left=of sum2,xshift=1cm] {\textcolor{darkblue}{$\begin{bmatrix} 1.06 \end{bmatrix}$}};
     \node [left=of sigmoid2,xshift=1cm] {\textcolor{darkblue}{$\begin{bmatrix} .743 \end{bmatrix}$}};
     \node [left=of l2,xshift=1cm] {\textcolor{darkblue}{$\begin{bmatrix} .0331 \end{bmatrix}$}};
     \node [right=of t,xshift=-1cm] {\textcolor{darkgreen}{$\begin{bmatrix} 1.0 \end{bmatrix}$}};

     \node [right=of prod1,xshift=-1cm] {\textcolor{purple}{$\begin{bmatrix} -.0935 \\ -.116 \end{bmatrix}$, $\begin{bmatrix} .0484 & 0 \\ -.0258 & 0 \end{bmatrix}$}};
     \node [right=of sum1,xshift=-1cm] {\textcolor{purple}{$\begin{bmatrix}.0484 \\ -.0258 \end{bmatrix}$, $\begin{bmatrix} .0484 \\ -.0258 \end{bmatrix}$}};
     \node [right=of sigmoid1,xshift=-1cm] {\textcolor{purple}{$\begin{bmatrix} .0484 \\ -.0258 \end{bmatrix}$}};
     \node [right=of prod2,xshift=-1cm] {\textcolor{purple}{$\begin{bmatrix} .246 \\ -.246 \end{bmatrix},\begin{bmatrix} .0360 & .00587 \end{bmatrix}$}};
     \node [right=of sum2,xshift=-1cm] {\textcolor{purple}{$\begin{bmatrix} .0492 \end{bmatrix}$, $\begin{bmatrix} .0492 \end{bmatrix}$}};
     \node [right=of sigmoid2,xshift=-1cm] {\textcolor{purple}{$\begin{bmatrix} .191 \end{bmatrix} \times \begin{bmatrix} .257 \end{bmatrix} = \begin{bmatrix} .0492 \end{bmatrix}$}};
     \node [right=of l2,xshift=-1cm] {\textcolor{purple}{$\begin{bmatrix} .257 \end{bmatrix}$}};

\end{tikzpicture}
\end{center}
\caption{Computation graph with gradients computed in the backward pass for the training example $(0,1)^T \rightarrow 1.0$. Gradients are computed with respect to the input of the nodes, so some nodes that have two inputs also have two gradients. See text for details on the computations of the values.}
\label{fig:computation-graph-with-gradients}
\end{figure}

If we want to compute the gradient update for a parameter such as $W_2$, we compute values in a backward pass, starting from the error term $y$. See Figure~\ref{fig:computation-graph-with-gradients} for an illustration. 

To give more detail on the computation of the gradients in the backward pass, starting at the bottom of the graph:
\begin{itemize}

\item For the {\bf L2} node, we use the formula
\begin{equation}
\frac{d \text{L2}(y,t)}{d y} = \frac{d \frac{1}{2} (t-y)^2}{d y} = t-y
\end{equation}
The given target output value given as training data is $t=1$, while we computed $y=0.743$ in the forward pass. Hence, the gradient for the L2 norm is $1-0.743=0.257$. Note that we are using values computed in the forward pass for these gradient computations.

\item For the lower {\bf sigmoid} node, we use the formula
\begin{equation}
\frac{d \text{sigmoid}(s)}{d s} = \text{sigmoid}(s) (1-\text{sigmoid}(s)) 
\end{equation}
Recall that the formula for the sigmoid is $\text{sigmoid}(s)=\frac{1}{1+e^{-s}}$.
Plugging in the value for $s=1.06$ computed in the forward pass into this formula gives us 0.191. The chain rule requires us to multiply this with the value that we just computed for the {\bf L2} node, i.e., 0.257, which gives us $0.191 \times 0.257 = 0.0492$.

\item For the lower {\bf sum} node, we simply copy the previous value, since the derivate is 1:
\begin{equation}
\frac{d \text{sum}(p,b_2)}{d p} = \frac{d p+b_2}{d p} = 1
\end{equation}
Note that there are two gradients associated with the sum node. One with respect to the output of the {\bf prod} node, and one with the $b_2$ parameter. In both cases, the derivative is 1, so both values are the same.
Hence, the gradient in both cases is 0.0492.

\item For the lower {\bf prod} node, we use the formula
\begin{equation}
\frac{d \text{prod}(h,W_2)}{d W_2} = \frac{d W_2 h}{d W_2} = h
\end{equation}
So far, we dealt with scalar values. Here we encounter vectors for the first time: the value of the hidden nodes $h=(0.731,0.119)^T$. The chain rule requires us to multiply this with the previously computed scalar 0.0492: 
\begin{equation*}
\Big(\begin{bmatrix} 0.731 \\ 0.119 \end{bmatrix} \times 0.0492\Big)^T = \begin{bmatrix} 0.0360 & 0.00587 \end{bmatrix}
\end{equation*}

As for the sum node, there are two inputs and hence two gradients. The other gradient is with respect to the output of the upper {\bf sigmoid} node.
\begin{equation}
\frac{d \text{prod}(h,W_2)}{d h} = \frac{d W_2 h}{d h} = W_2
\end{equation}

Similarly to above, we compute 
\begin{equation*}
(W_2 \times 0.0492)^T =  \Big(\begin{bmatrix} 5 & -5 \end{bmatrix} \times 0.0492\Big)^T= \begin{bmatrix} 0.246 \\ -0.246 \end{bmatrix}
\end{equation*}
\end{itemize}

Having all the gradients in place, we can now read of the relevant values for weight updates. These are the gradients associated with trainable parameters. For the $W_2$ weight matrix, this is the second gradient of the {\bf prod} node. So the new value for $W_2$ at time step $t+1$ is 
\begin{equation}
W_2^{t+1} = W_2^t - \mu \frac{d \text{prod}(x,W_2^t)}{d W_2^t} = \begin{bmatrix} 5 & 5 \end{bmatrix} - \mu \begin{bmatrix}  0.0360 & 0.00587  \end{bmatrix}
\end{equation}

The remaining computations are carried out in very similar fashion, since they form simply another layer of the feed-forward neural network.

Our example did not include one special case: the output of a computation may be used multiple times in subsequent steps of a computation graphs. So, there are multiple output nodes that feed back gradients in the back-propagation pass. In this case, we add up the gradients from these descendent steps to factor in their added impact.

Let us take a second look at what a node in a computation graph comprises:
\begin{itemize} \itemsep 0mm \vspace{-3mm}
\item a function that computes its value
\item links to input nodes (to obtain argument values)
\item when processing an example in the forward pass, the computed value
\item a function that executes its gradient computation
\item links to children nodes (to obtain downstream gradient values)
\item when processing an example in the forward pass, the computed gradient
\end{itemize}

From an object oriented programming view, a node in a computation graph provides a forward and backward function for value and gradient computations. As instantiated in an computation graph, it is connected with specific inputs and outputs, and is also aware of the dimensions of its variables its value and gradient. During forward and backward pass, these variables are filled in.

\subsection{Deep Learning Frameworks}\label{sec:theano}\index{deep learning!framworks}\index{framework}
In the next sections, we will encounter various network architectures. What all of these share, however, are the need for vector and matrix operations, as well as the computation of derivatives to obtain weight update formulas. It would be quite tedious to write almost identical code to deal with each these variants. Hence, a number of frameworks have emerged to support developing neural network methods for any chosen problem. At the time of writing, the most prominent ones are {\bf Theano}\NoteMarginal{Theano}\footnote{\tt http://deeplearning.net/software/theano/}\index{Theano} (a Python library that dymanically generates and compiles C++ code and is build on NumPy), {\bf Torch}\NoteMarginal{Torch}\index{Torch}\footnote{\tt http://torch.ch/} (a machine learning library  and a script language based on the Lua programming language), {\bf pyTorch}\index{pyTorch}\footnote{\tt http://pytorch.org/} (the Python variant of Torch), {\bf DyNet}\index{DyNet}\footnote{\tt http://dynet.readthedocs.io/} (a C++ implementation by natural language processing researchers that can be used as a library in C++ or Python), and {\bf Tensorflow}\index{Tensorflow}\NoteMarginal{Tensorflow}\footnote{\tt http://www.tensorflow.org/} (a more recent entry to the genre from Google).

These frameworks are less geared towards ready-to-use neural network architectures, but provide efficient implementations of the vector space operations and computation of derivatives, with seamless support of GPUs. 
Our example from Section~\ref{sec:nn:introduction} can be implemented in a few lines of Python code, as we will show in this section, using the example of Theano (other frameworks are quite similar).

You can execute the following commands on the Python command line interface if you first installed Theano ({\tt pip install Theano}).

\noindent {\tt \small > import numpy\\
> import theano\\
> import theano.tensor as T}

The mapping of the input layer {\tt x} to the hidden layer {\tt h} uses a weight matrix {\tt W}, a bias vector {\tt b}, and a mapping function which consists of the linear combination {\tt T.dot} and the sigmoid activation function.

\noindent {\tt \small > x = T.dmatrix()\\
> W = theano.shared(value=numpy.array([[3.0,2.0],[4.0,3.0]]))\\
> b = theano.shared(value=numpy.array([-2.0,-4.0]))\\
> h = T.nnet.sigmoid(T.dot(x,W)+b)}

Note that we define {\tt x} as a matrix. This allows us to process several training examples at once (a sequence of vectors).
A good way to think about these definitions of {\tt x} and {\tt h} is in term of a functional programming language. They symbolically define operations. To actually define a function that can be called, the Theano method {\tt function} is used.

\noindent {\tt \small > h\_function = theano.function([x], h)\\
> h\_function([[1,0]])\\
array([[ 0.73105858,  0.11920292]])}

This example call to {\tt h\_function} computes the values for the hidden nodes (compare to the numbers in Table~\vref{tab:nn:example-feed-forward-calculations}).\vspace{2mm}

The mapping from the hidden layer {\tt h} to the output layer {\tt y} is defined in the same fashion.

\noindent {\tt \small W2 = theano.shared(value=numpy.array([5.0,-5.0] ))\\
b2 = theano.shared(-2.0)\\
y\_pred = T.nnet.sigmoid(T.dot(h,W2)+b2)}

Again, we can define a callable function to test the full network.

\noindent {\tt \small > predict = theano.function([x], y\_pred)\\
> predict([[1,0]])\\
array([[ 0.7425526]])}

Model training requires the definition of a cost function (we use the L2 norm). To formulate it, we first need to define the variable for the correct output.
The overall cost is computed as average over all training examples.

\noindent {\tt \small > y = T.dvector()\\
> l2 = (y-y\_pred)**2\\
> cost = l2.mean()}

Gradient descent training requires the computation of the derivative of the cost function with respect to the model parameters (i.e., the values in the weight matrices {\tt W} and {\tt W2} and the bias vectors {\tt b} and {\tt b2}. A great benefit of using Theano is that it computes the derivatives for you. The following is also an example of a function with multiple inputs and multiple outputs.

\noindent{\tt \small > gW, gb, gW2, gb2 = T.grad(cost, [W,b,W2,b2])}

We have now all we need to define training. The function updates the model parameters and returns the current predictions and cost. It uses a learning rate of 0.1.

\noindent{\tt \small > train = theano.function(inputs=[x,y],outputs=[y\_pred,cost], \\
\phantom{> > } updates=((W, W-0.1*gW), (b, b-0.1*gb),\\
\phantom{> > updates=(}(W2, W2-0.1*gW2), (b2, b2-0.1*gb2)))}

Let us define the training data.

\noindent {\tt \small > DATA\_X = numpy.array([[0,0],[0,1],[1,0],[1,1]])\\
> DATA\_Y = numpy.array([0,1,1,0])\\
> predict(DATA\_X)\\
array([ 0.18333462,  0.7425526 ,  0.7425526 ,  0.33430961])\\
> train(DATA\_X,DATA\_Y)\\ 
{[}array([ 0.18333462,  0.7425526 ,  0.7425526 ,  0.33430961]),\\
array(0.06948320612438118)]}

The training function returns the prediction and cost before the updates. If we call the training function again, then the predictions and cost have changed for the better.

\noindent {\tt \small > train(DATA\_X,DATA\_Y)\\ %
{[}array([ 0.18353091,  0.74260499,  0.74321824,  0.33324929]),\\
      array(0.06923193686092949)]}

Typically, we would loop over the training function until convergence. As discussed above, we may also break up the training data into mini-batches and train on one mini-batch at a time.

\section{Neural Language Models}\index{language model}\index{neural language model}
Neural networks are a very powerful method to model conditional probability distributions with multiple inputs $p(a|b,c,d)$. They are robust to unseen data points --- say, an unobserved (a,b,c,d) in the training data. Using traditional statistical estimation methods, we may address such a sparse data problem with back-off and clustering, which require insight into the problem (what part of the conditioning context to drop first?) and arbitrary choices (how many clusters?). 

N-gram language models which reduce the probability of a sentence to the product of word probabilities in the context of a few previous words --- say, $p(w_i|w_{i-4},w_{i-3},w_{i-2},w_{i-1})$. Such models are a prime example for a conditional probability distribution with a rich conditioning context for which we often lack data points and would like to cluster information. In statistical language models, complex discounting and back-off schemes are used to balance rich evidence from lower order models --- say, the bigram model $p(w_i|w_{i-1})$ --- with the sparse estimates from high order models. Now, we turn to neural networks for help.

\subsection{Feed-Forward Neural Language Models}\label{sec:nn:feed-forward-nlm}
\begin{figure}
\begin{center}
\includegraphics[width=10cm]{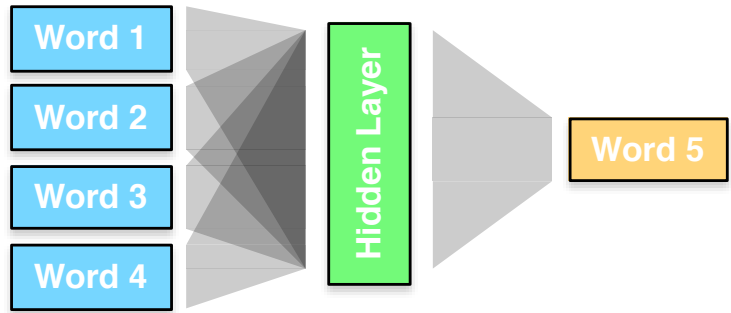}
\end{center}
\caption{Sketch of a neural language model: We predict a word $w_i$ based on its preceding words.}
\label{fig:nn:nnlm-sketch}
\end{figure}

\index{neural network!feed-forward}\index{feed-forward neural network}\index{language model!feed-forward}\index{neural language model!feed-forward}
Figure~\ref{fig:nn:nnlm-sketch} gives a basic sketch of a 5-gram neural network language model. Network nodes representing the context words have connections to a hidden layer, which connects to the output layer for the predicted word.

\paragraph{Representing Words}
We are immediately faced with a difficult question: How do we represent words? Nodes in a neural network carry real-numbered values, but words are discrete items out of a very large vocabulary. We cannot simply use token IDs, since the neural network will assume that token 124,321 is very similar to token 124,322 --- while in practice these numbers are completely arbitrary. The same arguments applies to the idea of using bit encoding for token IDs. The words $(1,1,1,1,0,0,0,0)^T$ and $(1,1,1,1,0,0,0,1)^T$ have very similar encodings but may have nothing to do with each other. While the idea of using such bit vectors is occasionally explored, it does not appear to have any benefits over what we consider next.

Instead, we will represent each word with a high-dimensional vector, one dimension per word in the vocabulary, and the value 1 for the dimension that matches the word, and 0 for the rest. 
The type of vectors are called {\bf one hot vector}\NoteMarginal{one hot vector}\index{one hot vector}\index{1-hot vector}.
For instance:
\begin{itemize} \itemsep 0mm
\item {\em dog} = $(0,0,0,0,1,0,0,0,0,...)^T$
\item {\em cat} = $(0,0,0,0,0,0,0,1,0,...)^T$
\item {\em eat} = $(0,1,0,0,0,0,0,0,0,...)^T$
\end{itemize}

These are very large vectors, and we will continue to wrestle with the impact of this choice to represent words. One stopgap is to limit the vocabulary to the most frequent, say, 20,000 words, and pool all the other words in an {\sc other} token. We could also use word classes (either automatic clusters or linguistically motivated classes such as part-of-speech tags) to reduce the dimensionality of the vectors. We will revisit the problem of large vocabularies later.

To pool evidence between words, we introduce another layer between the input layer and the hidden layer. In this layer, each context word is individually projected into a lower dimensional space. We use the same weight matrix for each of the context words, thus generating a continuous space representation for each word, independent of its position in the conditioning context. This representation is commonly referred to as {\bf word embedding}\NoteMarginal{word embedding}\index{word embedding}\index{embedding!word}. 

Words that occur in similar contexts should have similar word embeddings. 
For instance, if the training data for a language model frequently contains the n-grams
\begin{itemize} \itemsep 0mm
\item {\em but the cute dog jumped}
\item {\em but the cute cat jumped}
\item {\em child hugged the cat tightly}
\item {\em child hugged the dog tightly}
\item {\em like to watch cat videos}
\item {\em like to watch dog videos}
\end{itemize}
then the language model would benefit from the knowledge that {\em dog} and {\em cat}  occur in similar contexts and hence are somewhat interchangeable. If we like to predict from a context where {\em dog} occurs but we have seen this context only with the word {\em cat}, then we would still like to treat this as positive evidence. Word embeddings enable generalizing between words (clustering) and hence having robust predictions in unseen contexts (back-off).

\begin{figure}
\begin{center}
\includegraphics[width=13cm]{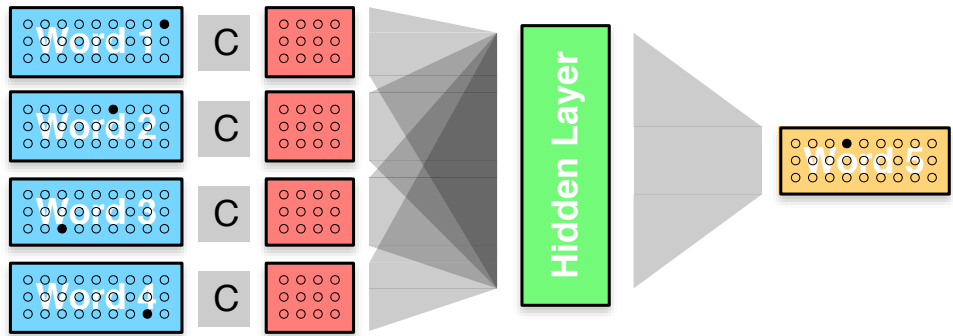}
\end{center}
\caption{Full architecture of a feed-forward neural network language model. Context words ($w_{i-4},w_{i-3},w_{i-2},w_{i-1}$) are represented in a one-hot vector, then projected into continuous space as word embeddings (using the same weight matrix $C$ for all words). The predicted word is computed as a one-hot vector via a hidden layer.}
\label{fig:nn:nnlm-third-sketch} 
\end{figure}

\paragraph{Neural Network Architecture}
See Figure~\ref{fig:nn:nnlm-third-sketch} for a visualization of the architecture the fully fledged feed forward neural network language model, consisting of the context words as one-hot-vector input layer, the word embedding layer, the hidden layer and predicted output word layer.

The context words are first encoded as one-hot vectors. These are then passed through the embedding matrix $C$, resulting in a vector of floating point numbers, the word embedding. This embedding vector has typically in the order of 500 or 1000 nodes. Note that we use the same embedding matrix $C$ for all the context words. 

Also note that mathematically there is not all that much going on here. Since the input to the multiplication to the matrix $C$ is a one hot vector, most of the input values to the matrix multiplication are zeros. So, practically, we are selecting the one column in the matrix that corresponds to the input word ID. Hence, there is no use for an activation function here. In a way, the embedding matrix a lookup table $C(w_j)$ for word embeddings, indexed by the word ID $w_j$. 
\begin{equation}
C(w_j) = C\; w_j
\end{equation}

Mapping to the hidden layer in the model requires concatenation of all context word embeddings $C(w_j)$ as input to a typical feed-forward layer, say, using tanh as activation function.
\begin{equation}
h = \text{tanh} \Big( b_h + \sum_j H_j C(w_j) \Big)
\end{equation}

The output layer is interpreted as a probability distribution over words. As before, first the linear combination $s_i$ of weights $w_{ij}$ and hidden node values $h_j$ is computed for each node $i$.
\begin{equation}
s = W\; h\label{eqn:nlm-hidden-to-out}
\end{equation}

To ensure that it is indeed a proper probability distribution, we use the {\bf softmax}\NoteMarginal{softmax}\index{softmax} activation function to ensure that all values add up to one.
\begin{equation}
p_i = \text{softmax}(s_i,\vec{s}) = \frac{e^{s_i}}{\sum_j e^{s_j}}
\end{equation}

What we described here is close to the neural probabilistic language model proposed by \cite{bengio:JMLR:2003}. This model had one more twist, it added direct connections of the context word embeddings to the output word. So, Equation~\ref{eqn:nlm-hidden-to-out} is replaced by 
\begin{equation}
s = W\; h + \sum_j U \; C(w_j)
\end{equation}

Their paper reports that having such {\bf direct connections}\index{direct connection} from context words to output words speeds up training, although does not ultimately improve performance. We will encounter the idea of short-cutting hidden layers again a bit later when we discuss deeper models with more hidden layers. They are also called {\bf residual connections}\index{residual connection}, {\bf skip connections}\index{skip connection}, or even {\bf highway connections}\index{highway connection}.

\paragraph{Training}
We train the parameters of a neural language model (word embedding matrix, weight matrices, bias vectors) by processing all the n-grams in the training corpus. For each n-gram, we feed the context words into the network and match the network's output against the one-hot vector of the correct word to be predicted. Weights are updated using back-propagation (we will go into details in the next section).

Language models are commonly evaluated by perplexity, which is related to the probability given to proper English text. A language model that likes proper English is a good language model. Hence, the training objective for language models is to increase the likelihood of the training data.

During training, given a context $\text{\bf x}=(w_{n-4},w_{n-3},w_{n-2},w_{n-1})$, we have the correct value for the 1-hot vector $\vec{y}$. For each training example $(\text{\bf x},\vec{y})$, likelihood is defined as 
\begin{equation}
L({\bf x},\vec{y};W) = - \sum_k y_k \; \text{log} \; p_k
\end{equation}
Note that only one value $y_k$ is 1, the others are 0. So this really comes down to the probability $p_k$ given to the correct word $k$. Defining likelihood this way allows us to update all weights,  also the one that lead to the wrong output words.

\subsection{Word Embedding}
Before we move on, it is worth reflecting the role of word embeddings in neural machine translation and many other natural language processing tasks. We introduced them here as compact encoding of words in relatively high-dimensional space, say 500 or 1000 floating point numbers. In the field of natural language processing, at the time of this writing, word embeddings have acquired the reputation of almost magical quality. 

Consider the role they play in the neural language language that we just described. They represent context words to enable prediction the next word in a sequence.

Recall part of our earlier example:
\begin{itemize} \itemsep 0mm
\item {\em but the cute dog jumped}
\item {\em but the cute cat jumped}
\end{itemize}

Since {\em dog} and {\em cat} occur in similar contexts, their influence on predicting the word {\em jumped} should be similar. It should be different from words such as {\em dress} which is unlikely to trigger the completion {\em jumped}. The idea that words that occur in similar contexts are semantically similar is a powerful idea in lexical semantics. 

At this point in the argument, researchers love to cite John Rupert Firth:
\begin{quotation}
\em You shall know a word by the company it keeps.
\end{quotation}

Or, as Ludwig Wittgenstein put it a bit more broadly:
\begin{quotation}
\em The meaning of a word is its use.
\end{quotation}

Meaning and semantics are quite difficult concepts with largely unresolved definition. The idea of {\bf distributional lexical semantics}\index{semantics}\index{distributional lexical semantics} is to define word meaning by their distributional properties, i.e., in which contexts they occur. Words that occur in similar contexts ({\em dog} and {\em cat}) should have similar representations. In vector space models, such as the word embeddings that we use here, similarity can be measured by a distance function, e.g., the cosine distance --- the angle between the vectors.

\begin{figure}
\begin{center}
\includegraphics[width=\textwidth]{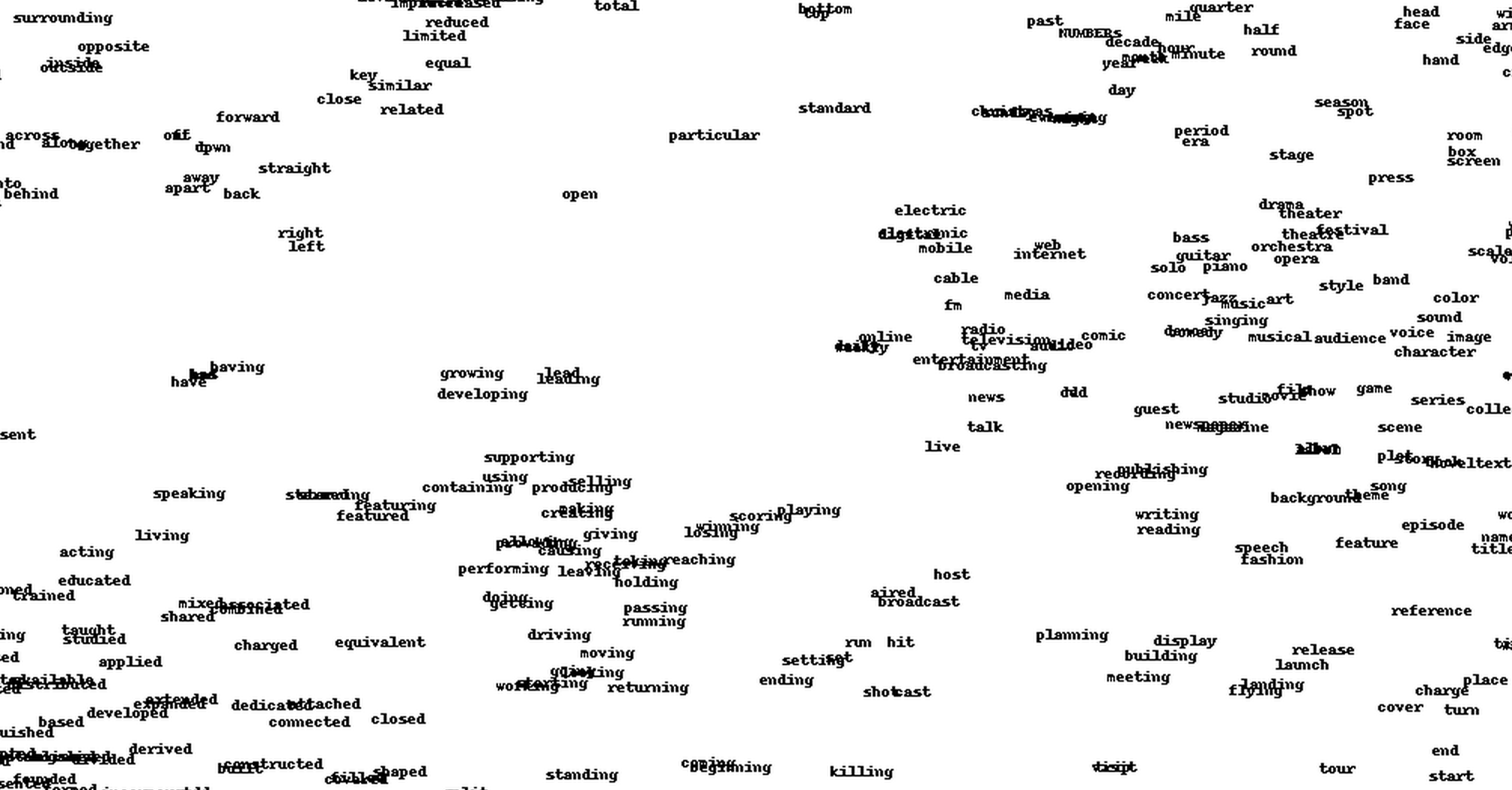}
\end{center}
\caption{Word embeddings projected into 2D. Semantically similar words occur close to each other.}
\label{fig:nn-word-embedding}
\end{figure}

If we project the high-dimensional word embeddings down to two dimensions, we can visualize word embeddings as shown in Figure~\ref{fig:nn-word-embedding}. In this figure, words that are similar ({\em drama, theater, festival}) are clustered together.

But why stop there? We would like to have semantic representations so we can carry out semantic inference such as
{\em 
\begin{itemize}
\item queen = king + (woman -- man)
\item queens = queen + (kings -- king)
\end{itemize}
}

Indeed there is some evidence that word embedding allow just that \citep{mikolov-yih-zweig:2013:NAACL-HLT}. However, we better stop here and just note that word embeddings are a crucial tool in neural machine translation.

\subsection{Efficient Inference and Training}\label{sec:noise-contrastive-estimation}
Training a neural language model is computationally expensive. For billion word corpora, even with the use of GPUs, training takes several days with modern compute clusters. Even using a neural language model as a scoring component in statistical machine translation decoding requires a lot of computation. We could restrict its use only to re-ranking n-best lists or lattices, or consider more efficient methods for inference and training.

\paragraph{Caching for Inference}
However, with a few considerations, it is actually possible to use this neural language model within the decoder.
\begin{itemize}
\item Word embeddings are fixed for the words, so do not actually need to carry out the mapping from one-hot vectors to word embeddings, but just store them beforehand.
\item The computation between embeddings and the hidden layer can be also partly carried out offline. Note that each word can occur in one of the 4 slots for conditioning context (assuming a 5-gram language model). For each of the slots, we can pre-compute the matrix multiplication of word embedding vector and the corresponding submatrix of weights. So, at run time, we only have to sum up these pre-computations at the hidden layer and apply the activation function.
\item Computing the value for each output node is insanely expensive, since there are as many output nodes as vocabulary items. However, we are interested only in the score for a given word that was produced by the translation model. If we only compute its node value, we have a score that we can use. 
\end{itemize}

The last point requires a longer discussion. If we compute the node value only for the word that we want to score with the language model, we are missing an important step. To obtain a proper probability, we need to normalize it, which requires the computation of the values for all the other nodes.

We could simply ignore this problem and use the scores at face value. More likely words given a context will get higher scores than less likely words, and that is the main objective. But since we place no constraints on the scores, we may work with models where some contexts give high scores to many words, while some contexts do not give preference for any.

It would be great, if the node values in the final layer were already normalized probabilities. There are methods to enforce this during training. Let us first discuss training in detail, and then move to these methods in Section~\ref{sec:noise-contrastive-estimation}.

\paragraph{Noise Contrastive Estimation}
We discussed earlier the problem that computing probabilities with a neural language model is very expensive due to the need to normalize the output node values $y_i$ using the softmax function. This requires computing values for all output nodes, even if we are only interested in the score for a particular n-gram. To overcome the need for this explicit normalization step, we would like to train a model that already has $y_i$ values that are normalized.

One way is to include the constraint that the normalization factor $Z(x)=\sum_j e^{s_j}$ is close to 1 in the objective function. So, instead of the just the simple likelihood objective, we may include the L2 norm of the log of this factor. Note that if $\text{log} \; Z(x) \simeq 0$, then $Z(x) \simeq 1$.

\begin{equation}
L({\bf x},\vec{y};W) = - \sum_k y_k \; \text{log} \; p_k - \alpha \; \text{log}^2 \; Z(x)
\end{equation}

Another way to train a self-normalizing model is called noise contrastive estimation. The main idea is to optimize the model so that it can separate correct training examples from artificially created noise examples. This method needs less computation during training, since it does not require the computation of all output node values. 

Formally, we are trying to learn the model distribution $p_m(\vec{y}|{\bf x};W)$. Given a noise distribution $p_n(\vec{y}|{\bf x})$ --- in our case of language modeling a unigram model $p_n(\vec{y})$ is a good choice --- we first generate a set of noise examples $U_n$ in addition to the correct training examples $U_t$. If both sets have the same size $|U_n|=|U_t|$, then the probability that a given example $({\bf x}; \vec{y}) \in U_n \cup U_t$ is predicted to be a correct training example is
\begin{equation}
p(\text{correct}|{\bf x},\vec{y}) = \frac{p_m(\vec{y}|{\bf x};W)}{p_m(\vec{y}|{\bf x};W) + p_n(\vec{y}|{\bf x})}
\end{equation}

The objective of noise contrastive estimation is to maximize $p(\text{correct}|{\bf x},\vec{y})$ for correct training examples $({\bf x}; \vec{y}) \in U_t$ and to minimize it for noise examples $({\bf x}; \vec{y}) \in U_n$. Using log-likelihood, we define the objective function as
\begin{equation}
L = \frac{1}{2|U_t|} \sum_{({\bf x}; \vec{y}) \in U_t} \text{log} \; p(\text{correct}|{\bf x},\vec{y}) \; \;
+ \frac{1}{2|U_n|} \sum_{({\bf x}; \vec{y}) \in U_n} \text{log} \; (1-p(\text{correct}|{\bf x},\vec{y}))
\end{equation}

Returning the the original goal of a self-normalizing model, first note that the noise distribution $p_n(\vec{y}|{\bf x})$ is normalized. Hence, the model distribution is encouraged to produce comparable values. If $p_m(\vec{y}|{\bf x};W)$ would generally overshoot --- i.e., $\sum_{\vec{y}} p_m(\vec{y}|{\bf x};W) > 1$ then it would also give too high values for noise examples. Conversely, generally undershooting would give too low values to correct translation examples.

Training is faster, since we only need to compute the output node value for the given training and noise examples --- there is no need to compute the other values, since we do not normalize with the softmax function. 

Given the definition of the training objective $L$, we have now a complete computation graph that we can implement using standard deep learning toolkits, as we have done before. These toolkits compute the gradients $\frac{dL}{dW}$ for all parameters $W$ and use them for parameter updates via gradient descent training (or its variants).


It may not be immediately obvious why optimizing towards classifying correct against noise examples gives rise to a model that also predicts the correct probabilities for n-grams. But this is a variant of methods that are common in statistical machine translation in the tuning phase. MIRA (margin infused relaxation algorithm) and PRO (pairwise ranked optimization) follow the same principle.


\subsection{Recurrent Neural Language Models}\label{sec:rnn}\index{recurrent neural language model}\index{language model!recurrent}\index{neural language model!recurrent}
The feed-forward neural language model that we described above is able to use longer contexts than traditional statistical back-off models, since it has more flexible means to deal with unknown contexts. Namely, the use of word embeddings to make use of similar words, and the robust handling of unseen words in any context position. Hence, it is possible to condition on much larger contexts than traditional statistical models. In fact, large models, say, 20-gram models, have been reported to be used.

Alternatively, instead of using a fixed context word window, {\bf recurrent neural networks}\NoteMarginal{recurrent neural network}\index{RNN}\index{recurrent neural network}\index{neural network!recurrent} may condition on context sequences of any length. The trick is to re-use the hidden layer when predicting word $w_n$ as additional input to predict word $w_{n-1}$.

\begin{figure}
\begin{center}
\includegraphics[scale=0.87]{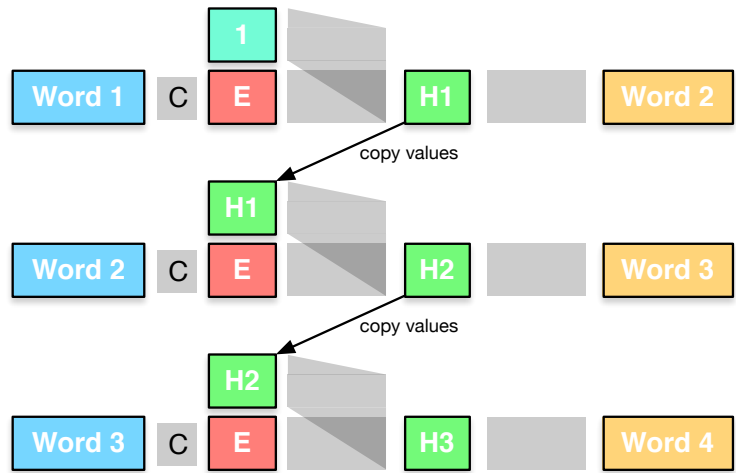}
\end{center}
\caption{Recurrent neural language models: After predicting Word~2 in the context of following Word~1, we re-use this hidden layer (alongside the correct Word~2) to predict Word~3. Again, the hidden layer of this prediction is re-used for the prediction of Word~4.}
\label{fig:nn:rnnlm}
\end{figure}

See Figure~\ref{fig:nn:rnnlm} for an illustration. Initially, the model does not look any different from the feed-forward neural language model that we discussed so far. The inputs to the network is the first word of the sentence $w_1$ and a second set of neurons which at this point indicate the start of the sentence. The word embedding of $w_1$ and the start-of-sentence neurons first map into a hidden layer $h_1$, which is then used to predict the output word $w_2$.

This model uses the same architecture as before: Words (input and output) are represented with one-hot vectors; word embeddings and the hidden layer use, say, 500 real valued neurons. We use a sigmoid activation function at the hidden layer and the softmax function at the output layer.

Things get interesting when we move to predicting the third word $w_3$ in the sequence. One input is the directly preceding (and now known) word $w_2$, as before. However, the neurons in the network that we used to represent start-of-sentence are now filled with values from the hidden layer of the previous prediction of word $w_2$. In a way, these neurons encode the previous sentence context. They are enriched at each step with information about a new input word and are hence conditioned on the full history of the sentence. So, even the last word of the sentence is conditioned in part on the first word of the sentence. Moreover, the model is simpler: it has less weights than a 3-gram feed-forward neural language model.

How do we train such a model with arbitrarily long contexts? 

One idea: At the initial stage (predicting the second word from the first), we have the same architecture and hence the same training procedure as for feed-forward neural networks. We assess the error at the output layer and propagate updates back to the input layer. We could process every training example this way --- essentially by treating the hidden layer from the previous training example as fixed input the current example. However, this way, we never provide feedback to the representation of prior history in the hidden layer.

\begin{figure}
\begin{center}
\includegraphics[scale=0.87]{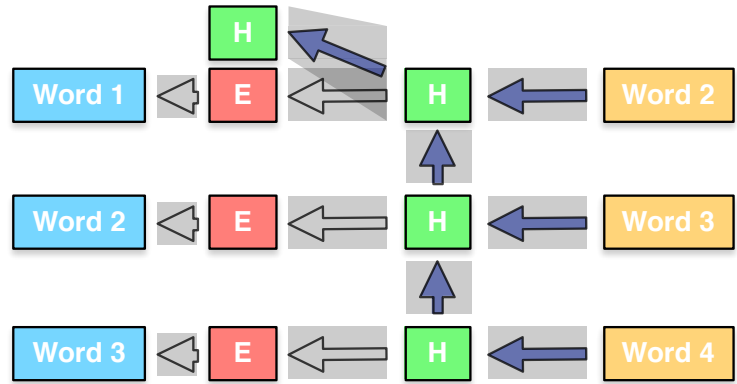}
\end{center}
\caption{Back-propagation through time: By unfolding the recurrent neural network over a fixed number of prediction steps (here: 3), we can derive update formulas based on the training objective of predicting all output words and back-propagation of the error via gradient descent.}
\label{fig:nn:rnnlm-training-bptt}
\end{figure}

The {\bf back-propagation through time}\NoteMarginal{back-propagation through time}\index{back-propagation!through time} training procedure (see Figure~\ref{fig:nn:rnnlm-training-bptt}) unfolds the recurrent neural network over a fixed number of steps, by going back over, say, 5 word predictions. 
Note that, despite limiting the unfolding to 5 time steps, the network is still able to learn dependencies over longer distances.

Back-propagation through time can be either applied for each training example (here called time step), but this is computationally quite expensive. Each time computations have to be carried out over several steps. Instead, we can compute and apply weight updates in mini-batches (recall Section~\ref{sec:nn:mini-batch}). First, we process a larger number of training examples (say, 10-20, or the entire sentence), and then update the weights.

Given modern compute power, fully {\bf unfolding the recurrent neural network}\index{unfolding} has become more common. While recurrent neural networks have in theory arbitrary length, given a specific training example, its size is actually known and fixed, so we can fully construct the computation graph for each given training example, define the error as the sum of word prediction errors, and then carry out back-propagation over the entire sentence. This does require that we can quickly build computation graphs --- so-called {\bf dynamic computation graphs}\index{dynamic computation graph}\index{computation graph!dynamic} --- which is currently supported by some toolkits better than others.

\subsection{Long Short-Term Memory Models}\label{sec:nn:lstm}
Consider the following step during word prediction in a sequential language model:
\begin{quotation}
{\em After much economic progress over the years, the {\bf country} $\rightarrow$ has}
\end{quotation}

The directly preceding word {\em country} will be the most informative for the prediction of the word {\em has}, all the previous words are much less relevant. In general, the importance of words decays with distance. The hidden state in the recurrent neural network will always be updated with the most recent word, and its memory of older words is likely to diminish over time.

But sometimes, more distant words are much more important, as the following example shows:

\begin{quotation}
{\em The {\bf country} which has made much economic progress over the years still $\rightarrow$ has}
\end{quotation}

In this example, the inflection of the verb {\em have} depends on the subject {\em country} which is separated by a long subordinate clause. 

Recurrent neural networks allow modeling of arbitrarily long sequences. Their architecture is very simple. But this simplicity causes a number of problems.
\begin{itemize}
\item The hidden layer plays double duty as memory of the network and as continuous space representation used to predict output words.
\item While we may sometimes want to pay more attention to the directly previous word, and sometimes pay more attention to the longer context, there is no clear mechanism to control that.
\item If we train the model on long sequences, then any update needs to back propagate to the beginning of the sentence. However, propagating through so many steps raises concerns that the impact of recent information at any step drowns out older information.\footnote{Note that there is a corresponding {\bf exploding gradient}\index{exploding gradient}\index{gradient!exploding} problem, where over long distance gradient values become too large. This is typically suppressed by {\bf clipping}\index{clipping}\index{gradient!clipping} gradients, i.e., limiting them to a maximum value set as a hyper parameter.}
\end{itemize}

The rather confusingly named {\bf long short-term memory (LSTM)}\index{LSTM}\index{long short-term memory}\NoteMarginal{long short-term memory} neural network architecture addresses these issues. Its design is quite elaborate, although it is not very difficult to use in practice. 

A core distinction is that the basic building block of LSTM networks, the so-called {\bf cell}\index{cell (LSTM)}\NoteMarginal{cell}, contains an explicit memory state. The memory state in the cell is motivated by digital memory cells in ordinary computers. Digital memory cells offer operations to read, write, and reset. While a digital memory cell may store just a single bit, a LSTM cell stores a real number.

\begin{figure}
\begin{center}
\includegraphics[scale=0.75]{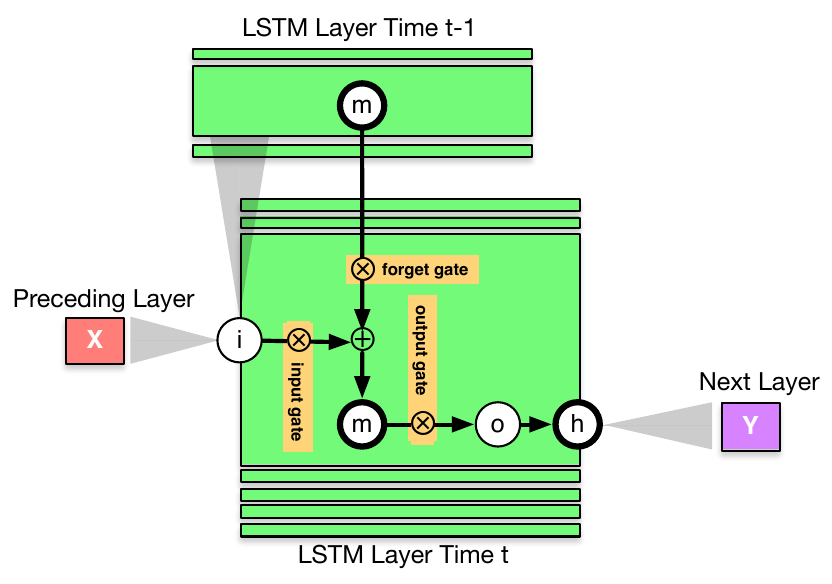}
\end{center}
\caption{A cell in a LSTM neural network. As recurrent neural networks, it receives input from the preceeding layer ($x$) and the hidden layer values from the previous time step $t-1$. The memory state $m$ is updated from the input state $i$ and the previous time's value of the memory state $m^{t-1}$. Various gates channel information flow in the cell towards the output value $o$.}
\label{fig:nn:lstm}
\end{figure}

Furthermore, the read/write/reset operations in a LSTM cell are regulated with a real numbered parameter, which are called {\bf gates}\index{gate} (see Figure~\ref{fig:nn:lstm}).

\begin{itemize}
\item The {\bf input gate}\index{input gate}\index{gate!index}\NoteMarginal{input gate} parameter regulates how much new input changes the memory state.
\item The {\bf forget gate}\index{forget gate}\index{gate!forget}\NoteMarginal{forget gate}  parameter regulates how much of the prior memory state is retained (or forgotten).
\item The {\bf output gate}\index{output gate}\index{gate!output}\NoteMarginal{output gate}  parameter regulates how strongly the memory state is passed on to the next layer.
\end{itemize}

Formally, marking the input, memory, and output values with the time step $t$, we define the flow of information within a cell as follows.
\begin{equation}
\begin{split}
\text{memory}^t &= \text{gate}_\text{input} \times \text{input}^t + \text{gate}_\text{forget} \times \text{memory}^{t-1}\\
\text{output}^t &= \text{gate}_\text{output} \times \text{memory}^t
\end{split}
\end{equation}

The hidden node value $h^t$ passed on to the next layer is the application of an activation function $f$ to the output value.
\begin{equation}
h^t = f( \text{output}^t )
\end{equation}

An LSTM layer consists of a vector of LSTM cells, just as traditional layers consist of a vector of nodes.
The input to LSTM layer is computed in the same way as the input to a recurrent neural network node. Given the node values for the prior layer $x^t$ and the values for the hidden layer from the previous time step $h^{t-1}$, the input value is the typical combination of matrix multiplication with weights $W^x$ and $W^h$ and an activation function $g$.
\begin{equation}
\text{input}^t = g \left( W^x x^t + W^h h^{t-1} \right)
\end{equation}

But how are the gate parameters set? They actually play a fairly important role. In particular contexts, we would like to give preference to recent input ($\text{gate}_\text{input} \simeq 1$), rather retain past memory ($\text{gate}_\text{forget} \simeq 1$), or pay less attention to the cell at the current point in time ($\text{gate}_\text{output} \simeq 0$). Hence, this decision has to be informed by a broad view of the context.

How do we compute a value from such a complex conditioning context? Well, we treat it like a node in a neural network. For each gate $a \in (\text{input},\text{forget},\text{output})$ we define matrices $W^{xa}$, $W^{ha}$, and $W^{ma}$ to compute the gate parameter value by the multiplication of weights and node values in the previous layer $x^t$, the hidden layer  $h^{t-1}$ at the previous time step, and the memory states at the previous time step $\text{memory}^{t-1}$, followed by an activation function $h$.

\begin{equation}
\text{gate}_a = h \left( W^{xa} x^t +  W^{ha} h^{t-1} + W^{ma} \text{memory}^{t-1} \right)
\end{equation}


LSTM are trained the same way as recurrent neural networks, using back-propagation through time or fully unrolling the network. While the operations within a LSTM cell are more complex than in a recurrent neural network, all the operations are still based on matrix multiplications and differentiable activation functions. Hence, we can compute gradients for the objective function with respect to all parameters of the model and compute update functions.

\subsection{Gated Recurrent Units}
LSTM cells add a large number of additional parameters. For each gate alone, multiple weight matrices are added. More parameters lead to longer training times and risk overfitting. As a simpler alternative, {\bf gated recurrent units}\index{gated recurrent unit}\index{GRU} (GRU) have been proposed and used in neural translation models. At the time of writing, LSTM cells seem to make a comeback in neural machine translation, but both are still commonly used.

\begin{figure}
\begin{center}
\includegraphics[scale=0.75]{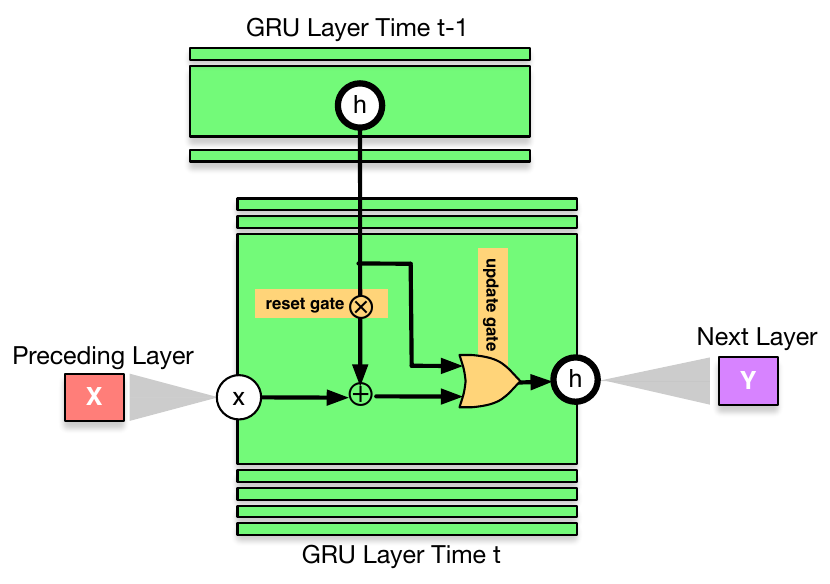}
\end{center}
\caption{Gated Recurrent Unit (GRU): a simplification of long short term memory (LSTM) cells.}
\label{fig:gru}
\end{figure}

See Figure~\ref{fig:gru} for an illustration for GRU cells. There is no separate memory state, just a hidden state that serves both purposes. Also, there are only two gates. These gates are predicted as before from the input and the previous state.

\begin{equation}
\begin{aligned}\text{update}_{t}&=g(W_{\text{update}}\;\text{input}_{t}+U_{\text{update}}\;\text{state}_{t-1}+\text{bias}_{\text{update}})\\
\text{reset}_{t}&=g(W_{\text{reset}}\;\;\;\text{input}_{t}+U_{\text{reset}}\;\;\;\;\text{state}_{t-1}+\text{bias}_{\text{reset}})\\
\end{aligned}
\end{equation}

The first gate is used in the combination of the input and previous state. This is combination is identical to traditional recurrent neural network, except that the previous states impact is scaled by the reset gate. Since the gate's value is between 0 and 1, this may give preference to the current input.

\begin{equation}
\text{combination}_t=f(W\;\text{input}_{t}+U(\text{reset}_{t}\circ \text{state}_{t-1}))
\label{eqn:gru-combination}
\end{equation}

Then, the update gate is used for a interpolation of the previous state and the just computed combination. This is done as a weighted sum, where the update gate balances between the two.

\begin{equation}
\begin{aligned}\text{state}_{t}=&(1-\text{update}_{t})\circ \text{state}_{t-1}\;+\\&\text{update}_{t}\;\;\;\;\;\;\;\;\;\circ\text{combination}_t)+\text{bias}\end{aligned}
\label{eqn:gru-output}
\end{equation}

In one extreme case, the update gate is 0, and the previous state is passed through directly. In another extreme case, the update gate is 1, and the new state is mainly determined from the input, with as much impact from the previous state as the reset gate allows.

It may seem a bit redundant to have two operations with a gate each that combine prior state and input. However, these play different roles. The first operation yielding $\text{combination}_t$ (Equation~\ref{eqn:gru-combination}) is a classic recurrent neural network component that allows more complex computations in the combination of input and output. The second operation yielding the new hidden state and the output of the unit (Equation~\ref{eqn:gru-output}) allows for bypassing of the input, enabling long-distant memory that simply passes through information and, during back-propagation, passes through the gradient, thus enabling long-distance dependencies.

\subsection{Deep Models}\label{sec:deep-rnn}\index{deep model}\index{model!deep}
The currently fashionable name {\bf deep learning} for the latest wave of neural network research has a real motivation. Large gains have been seen in tasks such as vision and speech recognition due to stacking multiple hidden layers together. 

More layers allow for more complex computations, just as having sequences of traditional computation components (Boolean gates) allows for more complex computations such as addition and multiplication of numbers. While this has been generally recognized for a long time, modern hardware finally enabled to train such deep neural networks on real world problems. And we learned from experiments in vision and speech that having a handful, and even dozens of layers does give increasingly better quality.

\begin{figure}
\begin{center}
\includegraphics[width=\textwidth]{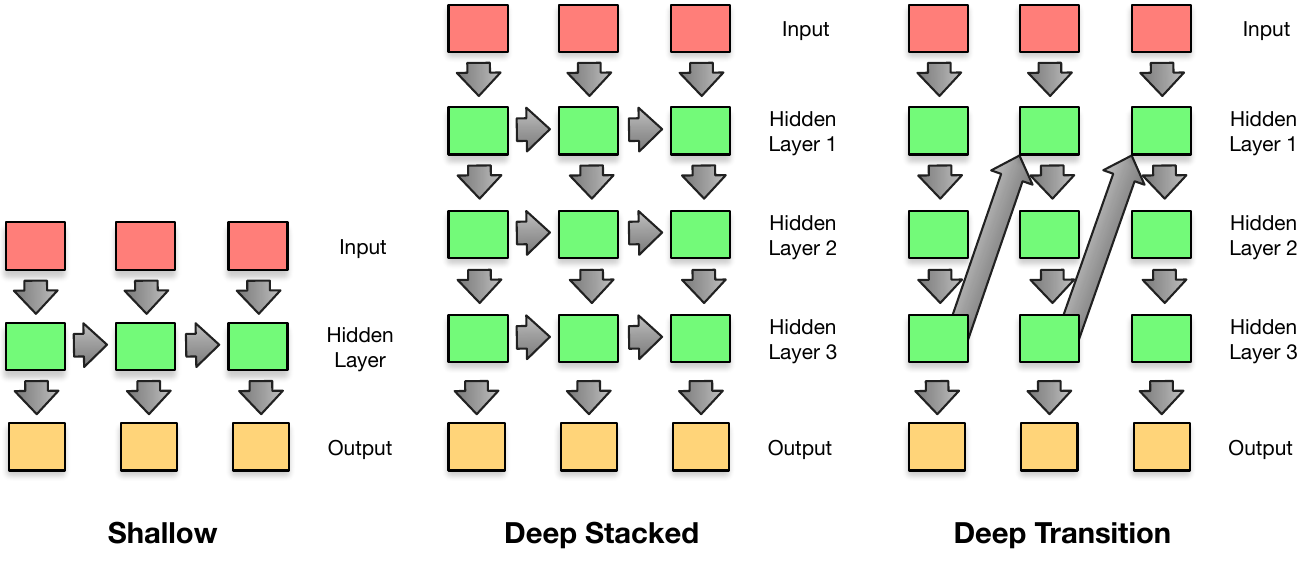}
\end{center}
\caption{Deep recurrent neural networks. The input is passed through a few hidden layers before an output prediction is made. In deep stacked models, the hidden layers are also connected horizontally, i.e., a layer's values at time step $t$ depends on its value at time step $t-1$ as well as the previous layer at time step $t$. In deep transitional models, the layers at any time step $t$ are sequentially connected and first hidden layer is also informed by the last layer at time step $t-1$.}
\label{fig:deep-rnn}
\end{figure}

How does the idea of deep neural networks apply to the sequence prediction tasks common in language? There are several options. Figure~\ref{fig:deep-rnn} gives two examples. In shallow neural networks, the input is passed to a single hidden layer, from which the output is predicted. Now, a sequence of hidden layers is used. These hidden layers $h_{t,i}$ may be {\bf deeply stacked}\index{deeply stacked}\index{stacked}, so that each layer acts like the hidden layer in the shallow recurrent neural network. Its state is conditioned on its value at the previous time step $h_{t-1,i}$  and the value of previous layer in the sequence $h_{t,i-1}$. 

\begin{equation}
\begin{aligned}
h_{t,1} & = f_1(h_{t-1,1}, x_t) & \text{first layer}\\
h_{t,i} & = f_i(h_{t-1,i}, h_{t,i-1}) & \text{for $i>1$}\\
y_t & = f_{i+1}(h_{t,I}) & \text{prediction from last layer $I$}
\end{aligned}
\end{equation}

Or, the hidden layers may be directly connected in {\bf deep transitional}\index{deep transition} networks, where the first hidden layer $h_{t,1}$ is informed by the last hidden layer at the previous time step $h_{t-1,I}$, but all other hidden layers are not connected to values from previous time steps.

 \begin{equation}
\begin{aligned}
h_{t,1} & = f_1(h_{t-1,I}, x_t) & \text{first layer}\\
h_{t,i} & = f_i(h_{t,i-1}) & \text{for $i>1$}\\
y_t & = f_{i+1}(h_{t,I}) & \text{prediction from last layer $I$}
\end{aligned}
\end{equation}

In all these equations, the function $f_i$ may be a feedforward layer (matrix multiplication plus activation function), an LSTM cell or a GRU cell.

Experiments with using neural language models in traditional statistical machine translation have shown benefits with 3--4 hidden layers \citep{luong-kayser-manning:2015:CoNLL}.

While modern hardware allows training of deep models, they do stretch computational resources to their practical limit. Not only are there more computations in the neural network, convergence of training is typically slower. Adding skip connections (linking the input directly to the output or the final hidden layer) sometimes speeds up training, but we still talking about a several times longer training times than shallow networks.

\paragraph{Further Readings}
\furtherreadings{The first vanguard of neural network research tackled language models. A prominent reference for neural language model is \cite{bengio:JMLR:2003}, who implement an n-gram language model as a feed-forward neural network with the history words as input and the predicted word as output. \cite{schwenk-dechelotte-gauvain:2006:POS} introduce such language models to machine translation (also called ``continuous space language models"), and use them in re-ranking, similar to the earlier work in speech recognition. \cite{schwenk:csl:2007} propose a number of speed-ups. They made their implementation available as a open source toolkit \citep{pbml-93-schwenk}, which also supports training on a graphical processing unit (GPU) \citep{schwenk-rousseau-attik:2012:WLM}. 

By first clustering words into classes and encoding words as pair of class and word-in-class bits, \cite{pbml-102-baltescu-blunsom-hoang} reduce the computational complexity sufficiently to allow integration of the neural network language model into the decoder. Another way to reduce computational complexity to enable decoder integration is the use of noise contrastive estimation by \cite{vaswani-EtAl:2013:EMNLP}, which roughly self-normalizes the output scores of the model during training, hence removing the need to compute the values for all possible output words. \cite{baltescu-blunsom:2015:NAACL-HLT} compare the two techniques - class-based word encoding with normalized scores vs. noise-contrastive estimation without normalized scores - and show that the letter gives better performance with much higher speed. 

As another way to allow straightforward decoder integration, \cite{wang-EtAl:2013:EMNLP2} convert a continuous space language model for a short list of 8192 words into a traditional n-gram language model in ARPA (SRILM) format. \cite{wang-EtAl:2014:EMNLP20142} present a method to merge (or ``grow") a continuous space language model with a traditional n-gram language model, to take advantage of both better estimate for the words in the short list and the full coverage from the traditional model.

\cite{finch-dixon-sumita:2012:NEWS2012} use a recurrent neural network language model to rescore n-best lists for a transliteration system. \cite{sundermeyer13:cmp} compare feed-forward with long short-term neural network language models, a variant of recurrent neural network language models, showing better performance for the latter in a speech recognition re-ranking task. \cite{mikolov:phd} reports significant improvements with reranking n-best lists of machine translation systems with a recurrent neural network language model.

Neural language model are not deep learning models in the sense that they use a lot of hidden layers. \cite{luong-kayser-manning:2015:CoNLL} show that having 3-4 hidden layers improves over having just the typical 1 layer.

{\em Language Models in Neural Machine Translation:} Traditional statistical machine translation models have a straightforward mechanism to integrate additional knowledge sources, such as a large out of domain language model. It is harder for end-to-end neural machine translation. \cite{DBLP:journals/corr/GulcehreFXCBLBS15} add a language model trained on additional monolingual data to this model, in form of a recurrently neural network that runs in parallel. They compare the use of the language model in re-ranking (or, re-scoring) against deeper integration where a gated unit regulates the relative contribution of the language model and the translation model when predicting a word.}

\section{Neural Translation Models}\label{nn:sec:neural-machine-translation}
We are finally prepared to look at actual translation models. We have already done most of the work, however, since the most commonly used architecture for neural machine translation is a straightforward extension of neural language models with one refinement, an alignment model.

\subsection{Encoder-Decoder Approach}\index{encoder-decoder}
Our first stab at a neural translation model is a straightforward extension of the language model. Recall the idea of a recurrent neural network to model language as a sequential process. Given all previous words, such a model predicts the next word. When we reach the end of the sentence, we now proceed to predict the translation of the sentence, one word at a time. 

\begin{figure}
\begin{center}
\includegraphics[width=\textwidth]{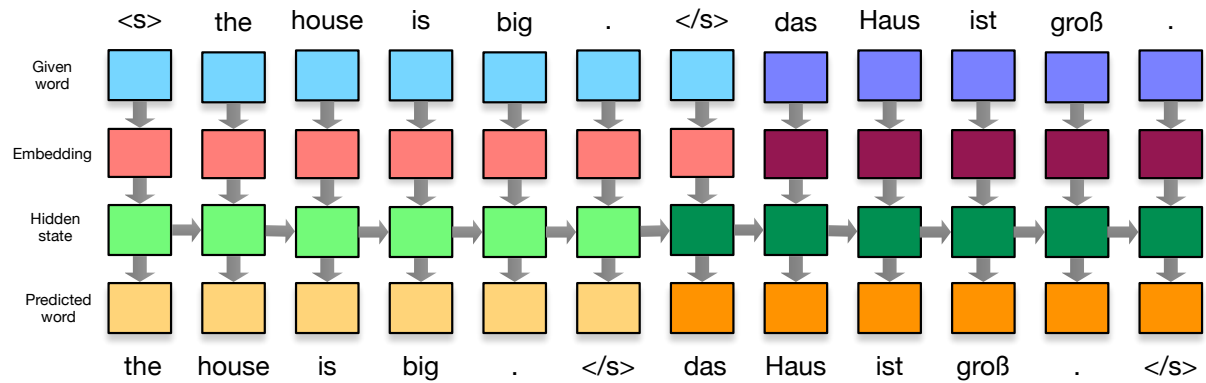}
\end{center}
\caption{Sequence-to-sequence encoder-decoder model: Extending the language model, we concatenate the English input sentence {\em the house is big} with the German output sentence {\em das Haus ist gro\ss}. The first dark green box (after processing the end-of-sentence token {\em $<$/s$>$}) contains the embedding of the entire input sentence .}
\label{fig:seq2seq}
\end{figure}

See Figure~\ref{fig:seq2seq} for an illustration. To train such a model, we simply concatenate the input and output sentences and use the same method as to train a language model. For decoding, we feed in the input sentence, and then go through the predictions of the model until it predicts an end of sentence token.

How does such a network work? Once processing reaches the end of the input sentence (having predicted the end of sentence marker {\em $<$/s$>$}), the hidden state encodes its meaning. In other words, the vector holding the values of the nodes of this final hidden layer is the {\bf input sentence embedding}\index{sentence embedding}\index{embedding!sentence}. This is the {\bf encoder}\index{encoder} phase of the model. Then this hidden state is used to produce the translation in the {\bf decoder}\index{decoder} phase. 

Clearly, we are asking a lot from the hidden state in the recurrent neural network here. During encoder phase, it needs to incorporate all information about the input sentence. It cannot forget the first words towards the end of the sentence. During the decoder phase, not only does it need to have enough information to predict each next word, there also needs to be some accounting for what part of the input sentence has been already translated, and what still needs to be covered.

In practice, the proposed models works reasonable well for short sentences (up to, say, 10--15 words), but fails for long sentences. 
Some minor refinements to this model have been proposed, such using the sentence embedding state as input to all hidden states of the decoder phase of the model. This makes the decoder structurally different from the encoder and reduces some of the load from the hidden state during decoding, since it does not need to remember anymore the input. Another idea is to reverse the order of the output sentence, so that the last words of the input sentences are close to the last words of the output sentence.

However, in the following section, we will embark on a more significant improvement of the model, by explicitly modelling alignment of output words to input words.

\subsection{Adding an Alignment Model}\index{alignment}
At the time of writing, the state of the art in neural machine translation is a sequence-to-sequence encoder-decoder model with attention. That is a mouthful, but it is essentially the model we just described in the previous section, with a explicitly alignment mechanism. In the deep learning world, this alignment is called {\bf attention}\index{attention}, we are using the words alignment and attention interchangeable here. 

Since the attention mechanism does add a bit of complexity to the model, we are now slowly building up to it, by first taking a look at the encoder, then the decoder, and finally the attention mechanism.

\subsubsection{Encoder}
The task of the encoder is to provide a representation of the input sentence. The input sentence is a sequence of words, for which we first consult the embedding matrix. Then, as in the basic language model described previously, we process these words with a recurrent neural network. This results in hidden states that encode each word with its left context, i.e., all the preceding words. To also get the right context, we also build a recurrent neural network that runs right-to-left, or more precisely, from the end of the sentence to the beginning.

\begin{figure}
\begin{center}
\includegraphics[scale=1]{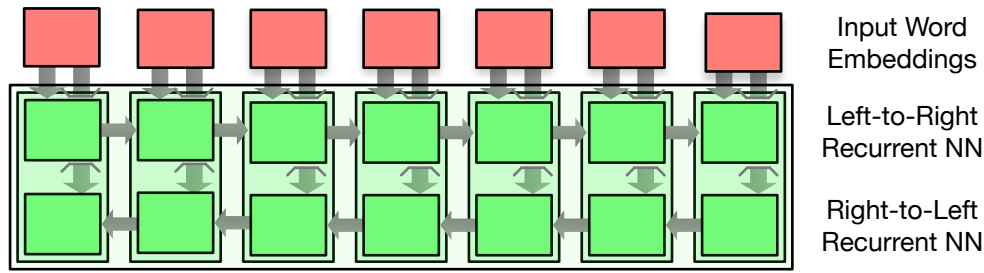}
\end{center}
\caption{Neural machine translation model, part 1: input encoder. It consists of two recurrent neural networks, running right to left and left to right (bidrectional recurrent neural network). The encoder states are the combination of the two hidden states of the recurrent neural networks.}
\label{fig:nn-attention-model-input-encoding}
\end{figure}

Figure~\ref{fig:nn-attention-model-input-encoding} illustrates the model. Having two recurrent neural networks running in two directions is called a {\bf bidirectional recurrent neural network}\index{bidirectional recurrent neural network}\index{neural network!bidirectional recurrent}\index{recurrent neural network!bidirectional}. Mathematically, the encoder consists of the embedding lookup for each input word $x_j$, and the mapping that steps through the hidden states $\overleftarrow{h_j}$ and $\overleftarrow{h_j}$
\begin{eqnarray}
\overleftarrow{h_j} = f(\overleftarrow{h_{j+1}}, \bar{E}\; x_j)\\
\overrightarrow{h_j} = f(\overrightarrow{h_{j-1}}, \bar{E}\; x_j)
\end{eqnarray}

In the equation above, we used a generic function $f$ for a cell in the recurrent neural network. This function may be a typical feed-forward neural network layer --- such as $f(x) = \text{tanh}( A x + b)$ --- or the more complex gated recurrent units (GRUs) or long short term memory cells (LSTMs). The original paper proposing this approached used GRUs, but lately LSTMs have become more popular.

Note that we could train these models by adding a step that predicts the next word in the sequence, but we are actually training it in the context of the full machine translation model. Limiting the description to the decoder, its output is a sequence of word representations that concatenate the two hidden states $(\overleftarrow{h_j},\overrightarrow{h_j})$.

\subsubsection{Decoder}
\begin{figure}
\begin{center}
\includegraphics[scale=1]{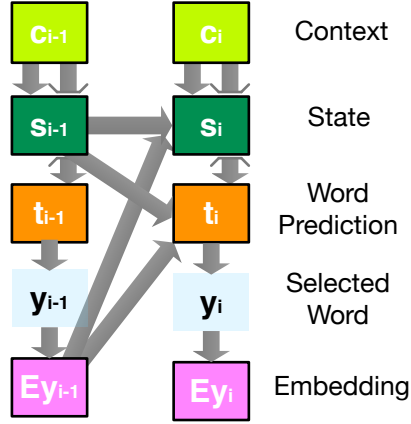}
\end{center}
\caption{Neural machine translation model, part 2: output decoder. Given the context from the input sentence, and the embedding of the previously selected word, new decoder states and word predictions are computed.}
\label{fig:nn-attention-model-with-output-words4-step2-detail}
\end{figure}

The decoder is also a recurrent neural network. It takes some representation of the input context (more on that in the next section on the attention mechanism) and the previous hidden state and output word prediction, and generates a new hidden decoder state and a new output word prediction. See Figure~\ref{fig:nn-attention-model-with-output-words4-step2-detail} for an illustration.

Mathematically, we start with the recurrent neural network that maintains a sequence of hidden states $s_i$ which are computed from the previous hidden state $s_{i-1}$, the embedding of the previous output word $Ey_{i-1}$, and the input context $c_i$ (which we still have to define). 
\begin{equation}
s_i = f(s_{i-1},\; Ey_{i-1},c_i)
\label{eq:nn:neural-translation-final}
\end{equation}

Again, there are several choices for the function $f$ that combines these inputs to generate the next hidden state: linear transforms with activation function, GRUs, LSTMs, etc. Typically, the choice here matches the encoder. So, if we use LSTMs for the encoder, then we also use LSTMs for the decoder.

From the hidden state. we now predict the output word. This prediction takes the form of a probability distribution over the entire output vocabulary. If we have a vocabulary of, say, 50,000 words, then the prediction is a 50,000 dimensional vector, each element corresponding to the probability predicted for one word in the vocabulary. 

The prediction vector $t_i$ is conditioned on the decoder hidden state $s_{i-1}$ and, again, the embedding of the previous output word $Ey_{i-1}$ and the input context $c_i$. 
\begin{equation}
t_i = \text{softmax} \big( W ( U s_{i-1} + V Ey_{i-1} + C c_i ) \big)
\end{equation}

Note that we repeat the conditioning on $Ey_{i-1}$ since we use the hidden state $s_{i-1}$ and not $s_1$. This separates the encoder state progression from $s_{i-1}$ to $s_{i}$ from the prediction of the output word $t_i$.

The softmax is used to convert the raw vector into a probability distribution, where the sum of all values is 1. Typically, the highest value in the vector indicates the output word token $y_i$. Its word embedding $Ey_{i-1}$ informs the next time step of the recurrent neural network.

During training, the correct output word $y_i$ is known, so training proceeds with that word. The training objective is to give as much probability mass as possible to the correct output word. The cost function that drives training is hence the negative log of the probability given to the correct word translation. 
\begin{equation}
\text{cost} = -\text{log} \; t_i[y_i]
\end{equation}

Ideally, we want to give the correct word the probability 1, which would mean a negative log probability of 0, but typically it is a lower probability, hence a higher cost. Note that the cost function is tied to individual words, the overall sentence cost is the sum of all word costs. 

During inference on a new test sentence, we typically chose the word $y_i$ with the highest value in $t_i$ use its embedding $Ey_i$ for the next steps. But we will also explore beam search strategies where the next likely words are selected as $y_i$, creating a different conditioning context for the next words. More on that later.

\subsubsection{Attention Mechanism}
We currently have two loose ends. The decoder gave us a sequence of word representations $h_j = (\overleftarrow{h_j},\overrightarrow{h_j})$ and the decoder expects a context $c_i$ at each step $i$. We now describe the attention mechanism that ties these ends together.

\begin{figure}
\begin{center}
\includegraphics[scale=1]{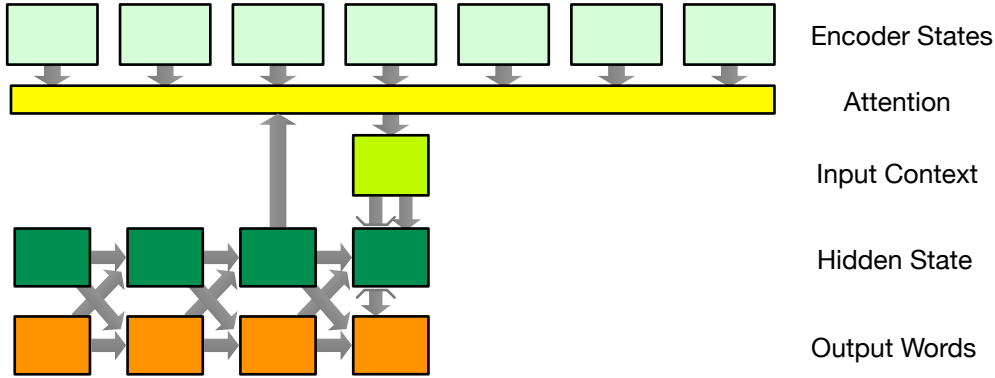}
\end{center}
\caption{Neural machine translation model, part 3: attention model. Associations are computed between the last hidden state of the decoder and the word representations (encoder states). These associations are used to compute a weighted sum of encoder states.}
\label{fig:nn-attention-model-with-output-words4-step5}
\end{figure}

The attention mechamism is hard to visualize using our typical neural network graphs, but Figure~\ref{fig:nn-attention-model-with-output-words4-step5} gives at least an idea what the input and output relations are. The attention mechanism is informed by all input word representations $(\overleftarrow{h_j},\overrightarrow{h_j})$ and the previous hidden state of the decoder $s_{i-1}$, and it produces a context state $c_i$. 

The motivation is that we want to compute an association between the decoder state (which contains information where we are in the output sentence production) and each input word. Based on how strong this association is, or in other words how relevant each particular input word is to produce the next output word, we want to weight the impact of its word representation. 

Mathematically, we first compute this association with a feedforward layer (using weight vectors $w^a$, $u^a$ and bias value $b^a$)
\begin{equation}
a(s_{i-1},h_j) = {w^a}^T s_{i-1} + {u^a}^T h_j + b^a
\label{eqn:attention-model}
\end{equation}
The output of this computation is a scalar value, indicating how important input word $j$ is to produce output word $i$.

We normalize this attention value, so that the attention values across all input words $j$ add up to one, using the softmax.
\begin{equation}
\alpha_{ij} = \frac{\text{exp}(a(s_{i-1},h_j))}{\sum_k \text{exp}(a(s_{i-1},h_k))}
\label{eqn:normalized-attention}
\end{equation}

Now we use the normalized attention value to weigh the contribution of the input word representation $h_j$ to the context vector $c_i$ and we are done.
\begin{equation}
c_i = \sum_j \alpha_{ij} h_j
\end{equation}

Simply adding up word representation vectors (weighted or not) may at first seem an odd and simplistic thing to do. But it is very common practice in deep learning for natural language processing. Researchers have no qualms about using sentence embeddings that are simply the sum of word embeddings and other such schemes.

\subsection{Training}
With the complete model in hand, we can now take a closer look at training. One challenge is that the number of steps in the decoder and the number of steps in the encoder varies with each training example. Sentence pairs consist of sentences of different length, so we cannot have the same computation graph for each training example but instead have to dynamically create the computation graph for each of them. This technique is called {\bf unrolling}\index{unrolling} the recurrent neural networks, and we already discussed it with regard to language models (recall Section~\ref{sec:rnn}).

\begin{figure}
\begin{center}
\includegraphics[scale=0.8]{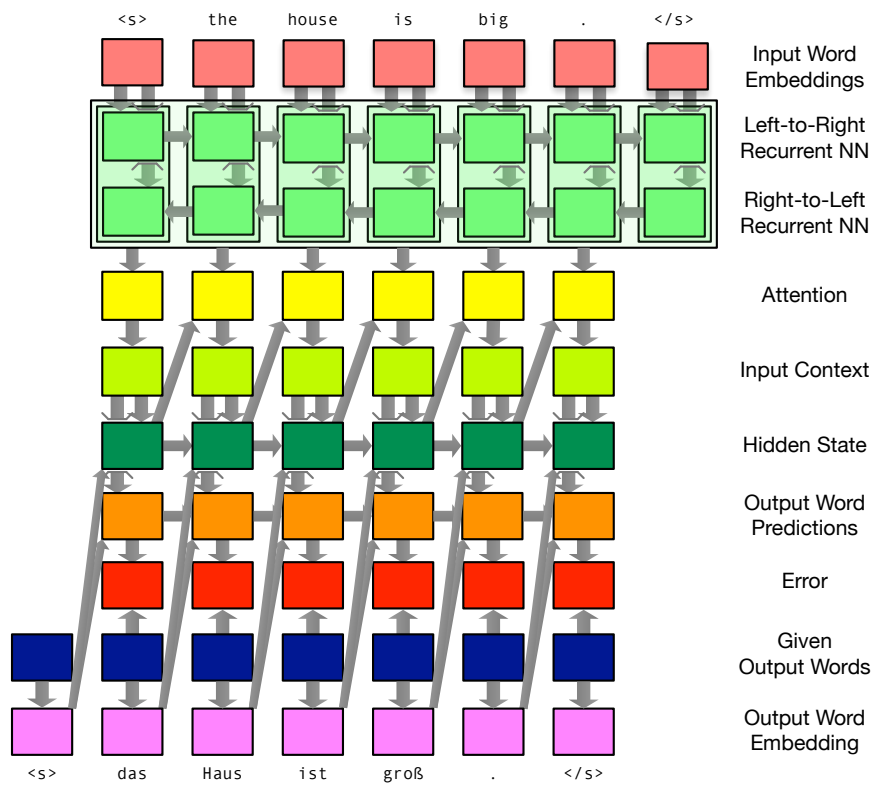}
\end{center}
\caption{Fully unrolled computation graph for training example with 7 input tokens {\em $<$s$>$ the house is big $<$/s$>$} and 6 output tokens {\em das Haus is gro\ss $<$/s$>$}. The cost function (error) is computed for each output word individually, and summed up across the sentence. When walking through the deocder states, the correct previous output words are used as conditioning context.}
\label{fig:nn-attention-model-with-output-words4-full}
\end{figure}

The fully unrolled computation graph for a short sentence pair is shown in Figure~\ref{fig:nn-attention-model-with-output-words4-full}. Note a couple of things. The error computed from this one sentence pair is the sum of the errors computed for each word. When proceeding to the next word prediction, we use the correct word as conditioning context for the decoder hidden state and the word prediction. Hence, the training objective is based on the probability mass given to the correct word, given a perfect context. There have been some attempts to use different training objectives, such as the {\sc bleu} score, but they have not yet been shown to be superior.

Practical training of neural machine translation models requires GPUs which are well suited to the high degree of parallelism inherent in these deep learning models (just think of the many matrix multiplications). To increase parallelism even more, we process several sentence pairs (say, 100) at once. This implies that we increase the dimensionality of all the state tensors.

To given an example. We represent each input word in specific sentence pair with a vector $h_j$. Since we already have a sequence of input words, these are lined up in a matrix. When we process a batch of sentence pairs, we again line up these matrices into a 3-dimensional tensor. 

Similarly, to give another example, the decoder hidden state $s_i$ is a vector for each output word. Since we process a batch of sentences, we line up their hidden states into a matrix. Note that in this case it is not helpful to line up the states for all the output words, since the states are computed sequentially.

Recall the first computation of the attention mechanism
\begin{equation}
a(s_{i-1},h_j) = W^as_{i-1} + U^ah_j + b^a
\end{equation}

We can pass this computation to the GPU with a matrix of encoder states $s_{i-1}$ and a 3-dimensional tensor of input encodings $h_j$, resulting in a matrix of attention values (one dimension for the sentence pairs, one dimension for the input words). Due to the massive re-use of values in $W^a$, $U^a$, and $b^a$ as well as the inherent parallelism of this computation, GPUs can show their true power.

You may feel that we just created a glaring contradiction. First, we argued that we have to process one training example at a time, since sentence pairs typically have different length, and hence computation graphs have different size. Then, we argued for batching, say, 100 sentence pairs together to better exploit parallelism. These are indeed conflicting goals.

\begin{figure}
\begin{center}
\includegraphics[scale=0.41]{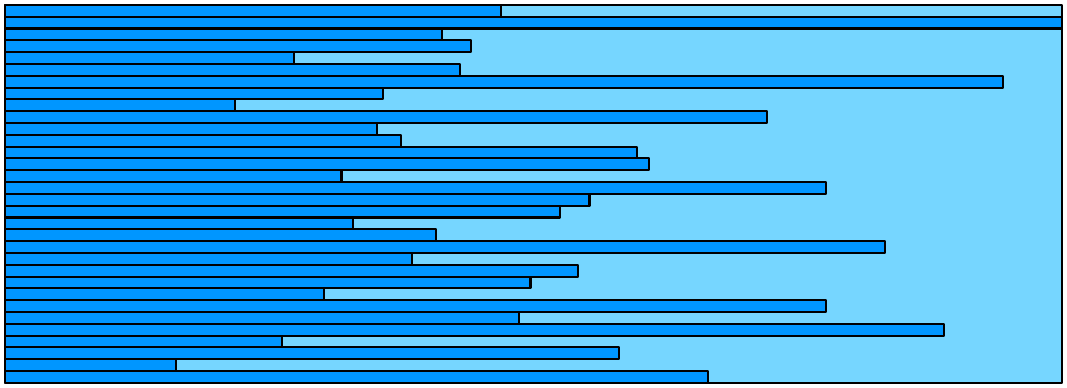}
{\huge $\Rightarrow$}
\includegraphics[scale=0.41]{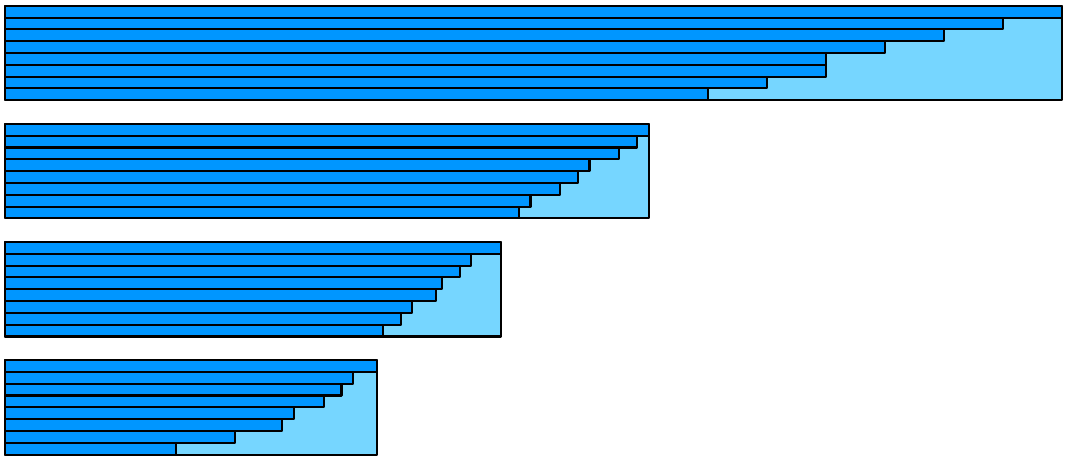}
\end{center}
\caption{To make better use of parallelism in GPUs, we process a batch of training examples (sentence pairs) at a time. Converting a batch of training examples into a set of mini batches that have similar length. This wastes less computation on filler words (light blue).}
\label{fig:batching}
\end{figure}

See Figure~\ref{fig:batching}. When batching training examples together, we have to consider the maximum sizes for input and output sentences in a batch and unroll the computation graph to these maximum sizes. For shorter sentences, we fill the remaining gaps with non-words and keep track of where the valid data is with a {\bf mask}\index{mask}. This means, for instance, that we have to ensure that no attention is given to words beyond the length of the input sentence, and no errors and gradient updates are computed from output words beyond the length of the output sentence.

To avoid wasted computations on gaps, a nice trick is to sort the sentence pairs in the batch by length and break it up into {\bf mini-batches}\index{mini batch} of similar length.\footnote{There is a bit of confusion of the technical terms here. Sometimes, the entire training corpus is called a {\em batch}, as used in the contrast between {\em batch} updating and {\em online} updating. In that context, smaller batches with a subset of the are called {\em mini-batches} (recall Section~\vref{sec:nn:mini-batch}). Here, we use the term {\em batch} (or {\em maxi-batch}) for such a subset, and {\em mini-batch} for a subset of the subset.}

To summarize, training consists of the following steps
\begin{itemize}\itemsep 0mm
\item Shuffle the training corpus (to avoid undue biases due to temporal or topical order)
\item Break up the corpus into maxi-batches
\item Break up each maxi-batch into mini-batches
\item Process each mini-batch, gather gradients
\item Apply all gradients for a maxi-batch to update the parameters
\end{itemize}

Typically, training neural machine translation models takes about 5--15 epochs (passes through entire training corpus). A common stopping criteria is to check progress of the model on a validation set (that is not part of the training data) and halt when the error on the validation set does not improve. Training longer would not lead to any further improvements and may even degrade performance due to overfitting.

\subsection{Beam Search}\index{beam search}

\begin{figure}
\begin{center}
\includegraphics[scale=1]{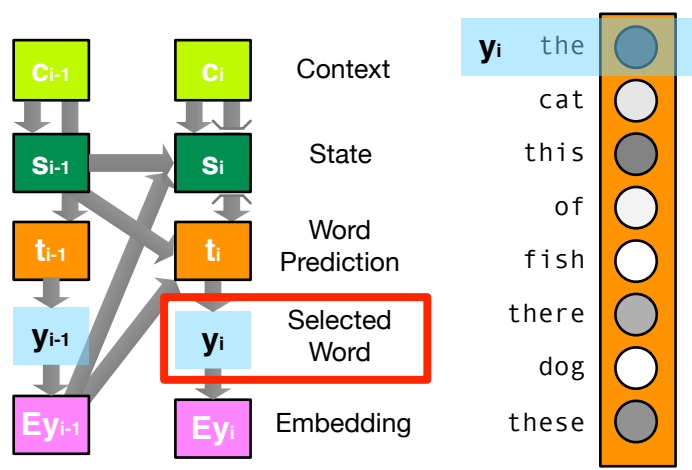}
\end{center}
\caption{Elementary decoding step: The model predicts a word prediction probability distribution. We select the most likely word ({\em the}). Its embedding is part of the conditioning context for the next word prediction (and decoder state).}
\label{fig:nn-attention-model-decoding-step3}
\end{figure}

Translating with neural translation models proceeds one step at a time. At each step, we predict one output word. In our model, we first compute a probability distribution over all words. We then pick the most likely word and move to the next prediction step. Since the model is conditioned on the previous output word (recall Equation~\ref{eq:nn:neural-translation-final}), we use its word embedding in the conditioning context for the next step. 

See Figure~\ref{fig:nn-attention-model-decoding-step3} for an illustration. At each time step, we obtain a probability distribution over words. In practice, this distribution is most often quite spiked, only few words --- or maybe even just one word --- amass almost all of the probability. In the example, the word {\em the} received the highest probability, so we pick it as the output word.

\begin{figure} \small
{\normalsize \bf Input Sentence}\\[2mm]
{\em ich glaube aber auch , er ist clever genug um seine Aussagen vage genug zu halten , so dass sie auf verschiedene Art und Weise interpretiert werden k\"onnen .}
\vspace{7mm}

{\normalsize \bf Output Word Predictions}\\[4mm]
\begin{tabular}{lr|l} 
\bf Best && \bf Alternatives\\ \hline
		{\bf but} & (42.1\%) &
		{\em however} (25.3\%), 
		{\em I} (20.4\%), 
		{\em yet} (1.9\%), 
		{\em and} (0.8\%), 
		{\em nor} (0.8\%), ...\\
		{\bf I} & (80.4\%) &
		{\em also} (6.0\%), 
		{\em ,} (4.7\%), 
		{\em it} (1.2\%), 
		{\em in} (0.7\%), 
		{\em nor} (0.5\%), 
		{\em he} (0.4\%), ...\\
		{\bf also} & (85.2\%) & 
		{\em think} (4.2\%), 
		{\em do} (3.1\%), 
		{\em believe} (2.9\%), 
		{\em ,} (0.8\%), 
		{\em too} (0.5\%), ...\\
		{\bf believe} & (68.4\%) & 
		{\em think} (28.6\%), 
		{\em feel} (1.6\%), 
		{\em do} (0.8\%), ...\\
		{\bf he} & (90.4\%) &
		{\em that} (6.7\%), 
		{\em it} (2.2\%), 
		{\em him} (0.2\%), ...\\
		{\bf is} & (74.7\%) &
		{\em 's} (24.4\%), 
		{\em has} (0.3\%), 
		{\em was} (0.1\%), ...\\
		{\bf clever} & (99.1\%) &
		{\em smart} (0.6\%), ...\\
		{\bf enough} & (99.9\%) & \\
		{\bf to} & (95.5\%) &
		{\em about} (1.2\%), 
		{\em for} (1.1\%), 
		{\em in} (1.0\%), 
		{\em of} (0.3\%), 
		{\em around} (0.1\%), ...\\
		{\bf keep} & (69.8\%) &
		{\em maintain} (4.5\%), 
		{\em hold} (4.4\%), 
		{\em be} (4.2\%), 
		{\em have} (1.1\%), 
		{\em make} (1.0\%), ...\\
		{\bf his} & (86.2\%) &
		{\em its} (2.1\%), 
		{\em statements} (1.5\%), 
		{\em what} (1.0\%), 
		{\em out} (0.6\%), 
		{\em the} (0.6\%), ...\\
		{\bf statements} & (91.9\%) &
		{\em testimony} (1.5\%), 
		{\em messages} (0.7\%), 
		{\em comments} (0.6\%), ...\\
		{\bf vague} & (96.2\%) &
		{\em v@@} (1.2\%), 
		{\em in} (0.6\%), 
		{\em ambiguous} (0.3\%), ...\\
		{\bf enough} & (98.9\%) &
		{\em and} (0.2\%), ...\\
		{\bf so} & (51.1\%) &
		{\em ,} (44.3\%), 
		{\em to} (1.2\%), 
		{\em in} (0.6\%), 
		{\em and} (0.5\%), 
		{\em just} (0.2\%), 
		{\em that} (0.2\%), ...\\
		{\bf they} & (55.2\%) &
		{\em that} (35.3\%), 
		{\em it} (2.5\%), 
		{\em can} (1.6\%), 
		{\em you} (0.8\%), 
		{\em we} (0.4\%), 
		{\em to} (0.3\%), ...\\
		{\bf can} & (93.2\%) &
		{\em may} (2.7\%), 
		{\em could }(1.6\%), 
		{\em are} (0.8\%), 
		{\em will} (0.6\%), 
		{\em might} (0.5\%),  ...\\
		{\bf be} & (98.4\%) &
		{\em have} (0.3\%), 
		{\em interpret} (0.2\%), 
		{\em get }(0.2\%),  ...\\
		{\bf interpreted} & (99.1\%) &
		{\em interpre@@} (0.1\%), 
		{\em constru@@} (0.1\%),  ...\\
		{\bf in} & (96.5\%) &
		{\em on} (0.9\%), 
		{\em differently} (0.5\%), 
		{\em as} (0.3\%), 
		{\em to} (0.2\%), 
		{\em for} (0.2\%), 
		{\em by} (0.1\%),  ...\\
		{\bf different} & (41.5\%) &
		{\em a} (25.2\%), 
		{\em various} (22.7\%), 
		{\em several} (3.6\%), 
		{\em ways} (2.4\%), 
		{\em some} (1.7\%), ...\\
		{\bf ways} & (99.3\%) &
		{\em way} (0.2\%), 
		{\em manner} (0.2\%),  ...\\
		{\bf .} & (99.2\%) &
		{\sc </s>} (0.2\%), 
		{\em ,} (0.1\%), ... \\
		{\bf </s>} & (100.0\%) & \\
\end{tabular}
\vspace{6mm}
\caption{Word predictions of the neural machine translation model. Frequently, most of the probability mass is given to the top choice, but semantically related words may rank high, e.g., {\em believe} (68.4\%) vs. {\em think} (28.6\%). The subword units {\em interpre@@} are explain in Section~\vref{sec:large-vocabulary}.}
\label{fig:decoding-example}
\end{figure}

A real example of how a neural machine translation model translates a German sentence into English is shown in Figure~\ref{fig:decoding-example}. The model tends to give most, if not almost all, probability mass to the top choice, but the sentence translation also indicates word choice ambiguity, such as {\em believe} (68.4\%) vs. {\em think} (28.6\%) or {\em different} (41.5\%) vs. {\em various} (22.7\%). There is also ambiguity about grammatical structure, such as if the sentence should start with the discourse connective {\em but} (42.1\%) or the subject {\em I} (20.4\%).

This process suggests that we perform 1-best greedy search. This makes us vulnerable to the  so-called {\bf garden-path problem}\index{garden path problem}. Sometimes we follow a sequence of words and realize too late that we made a mistake early on. In that case,
the best sequence consists of less probable words initially which are redeemed by subsequent words in the context of the full output. Consider the case of having to produce an idiomatic phrase that is non-compositional. The first words of these phrases may be really odd word choices by themselves (e.g., {\em piece of cake} for {\em easy}). Only once the full phrase is formed, their choice is redeemed.

Note that we are faced with the same problem in traditional statistical machine translation models --- arguable even more so there since we rely on sparser contexts when making predictions for the next words. Decoding algorithms for these models keep a list of the n-best candidate {\bf hypotheses}\index{hypothesis}, expand them and keep the n-best expanded hypotheses. We can do the same for neural translation models.

\begin{figure}
\begin{center}
\includegraphics[scale=1]{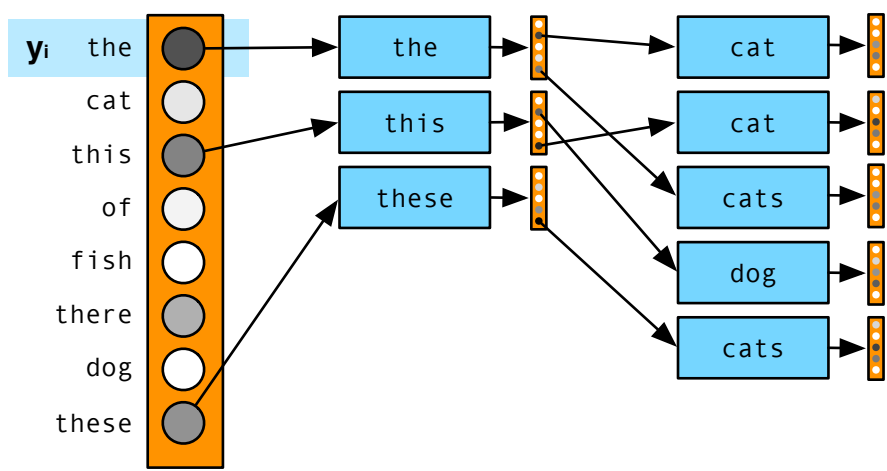}
\end{center}
\caption{Beam search in neural machine translation. After committing to a short list of specific output words (the {\bf beam}), new word predictions are made for each. These differ since the committed output word is part of the conditioning context to make predictions.}
\label{fig:nn-attention-model-decoding-beam}
\end{figure}

When predicting the first word of the output sentence, we keep a {\bf beam}\index{beam} of the top $n$ most likely word choices. They are scored by their probability. Then, we use each of these words in the beam in the conditioning context for the next word. Due to this conditioning, we make different word predictions for each. 
We now multiply the score for the partial translation (at this point just the probability for the first word), and the probabilities from its word predictions. We select the highest scoring word pairs for the next beam. 
See Figure~\ref{fig:nn-attention-model-decoding-beam} for an illustration. 

This process continues. At each time step, we accumulate word translation probabilities, giving us scores for each hypothesis. A sentence translation is complete, when the end of sentence token is produced. At this point, we remove the completed hypothesis from the beam and reduce beam size by 1. Search terminates, when no hypotheses are left in the beam. 

\begin{figure}
\begin{center}
\includegraphics[scale=1]{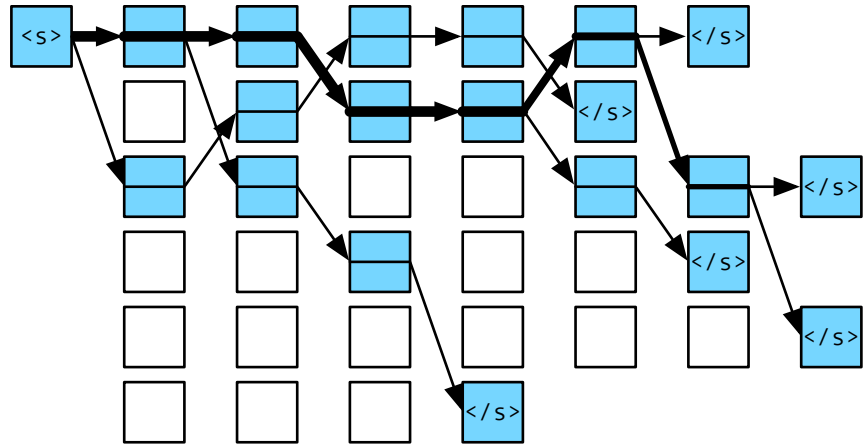}
\end{center}
\caption{Search graph for beam search decoding in neural translation models. At each time step, the $n=6$ best partial translations (called hypotheses) are selected. An output sentence is complete when the end of sentence token {\em $<$/s$>$} is predicted. We reduce the beam after that and terminate when $n$ full sentence translations are completed. Following the back-pointers from the end of sentence tokens allows us to read them off. Empty boxes represent hypotheses that are not part of any complete path.}
\label{fig:nn-attention-model-decoding-graph-paths}
\end{figure}

Search produces a graph of hypotheses, as shown in Figure~\ref{fig:nn-attention-model-decoding-graph-paths}. It starts with the start of sentence symbol  {\em $<$s$>$} and its paths terminate with the end of sentence symbol  {\em $<$/s$>$}. Given the compete graph, the resulting translations can be obtained by following the back-pointers. The complete hypothesis (i.e., one that ended with a  {\em $<$/s$>$} symbol) with the highest score points to the best translation.

When choosing among the best paths, we score each with the product of its word prediction probabilities. In practice, we get better results when we normalize the score by the output length of a translation, i.e., divide by the number of words. We carry out this normalization after search is completed. During search, all translations in a beam have the same length, so the normalization would make no difference.

Note that in traditional statistical machine translation, we were able to combine hypotheses if they share the same conditioning context for future feature functions. This not possible anymore for recurrent neural networks since we condition on the entire output word sequence from the beginning. As a consequence, the search graph is generally less diverse than search graphs in statistical machine translation models. It is really just a search tree where the number of complete paths is the same as the size of the beam.



\paragraph{Further Readings}
\furtherreadings{The attention model has its roots in a sequence-to-sequence model.
\cite{cho-EtAl:2014:SSST-8} use recurrent neural networks for the approach. \cite{NIPS2014_5346} use a LSTM (long short-term memory) network and reverse the order of the source sentence before decoding.

The seminal work by \cite{bahdanau:ICLR:2015} adds an alignment model (so called ``attention mechanism") to link generated output words to source words, which includes conditioning on the hidden state that produced the preceding target word. Source words are represented by the two hidden states of recurrent neural networks that process the source sentence left-to-right and right-to-left. \cite{luong-pham-manning:2015:EMNLP} propose variants to the attention mechanism (which they call ``global" attention model) and also a hard-constraint attention model (``local" attention model) which is restricted to a Gaussian distribution around a specific input word.

To explicitly model the trade-off between source context (the input words) and target context (the already produced target words), \cite{DBLP:journals/corr/TuLLLL16a} introduce an interpolation weight (called ``context gate") that scales the impact of the (a) source context state and (b) the previous hidden state and the last word when predicting the next hidden state in the decoder.

\cite{Tu-EtAl:2017:AAAI} augment the attention model with a reconstruction step. The generated output is translated back into the input language and the training objective is extended to not only include the likelihood of the target sentence but also the likelihood to the reconstructed input sentence.}

\section{Refinements}
The previous section gave a comprehensive description of the currently most commonly used basic neural translation model architecture. It performs fairly well out of the box for many language pairs. Since its conception, a number of refinements have been proposed. We will describe them in this section.

Some of the refinements are fairly general, some target particular use cases or data conditions. To given one example, the best performing system at the recent WMT 2017 evaluation campaign used ensemble decoding (Section~\ref{sec:ensemble}), byte pair encoding to address large vocabularies  (Section~\ref{sec:large-vocabulary}), added synthetic data derived from monolingual target side data  (Section~\ref{sec:monolingual-data}) , and used deeper models  (Section~\ref{sec:deep-models}).

\subsection{Ensemble Decoding}\label{sec:ensemble}
A common technique in machine learning is to not just build one system for your problem, but multiple ones and then combine them. This is called an {\bf ensemble}\index{ensemble} of systems. It is such a successful strategy that various methods have been proposed to systematically build alternative systems, for instance by using different features or different subsets of the data.  For neural networks, one straightforward way is to use different initializations or stop at different points in the training process.

Why does it work? The intuitive argument is that each system makes different mistakes. When two systems agree, then they are more likely both right, rather than both make the same mistake. One can also see the general principle at play in human behavior, such as setting up committees to make decisions or the democratic voting in elections.

Applying ensemble methods to our case of neural machine translation, we have to address two sub-problems: (1) generating alternate systems, and (2) combining their output.

\subsubsection{Generating Alternative Systems}
\begin{figure}
\begin{center}
\begin{tabular}{cc}
\includegraphics[scale=0.65]{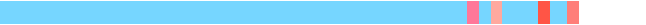} &
\includegraphics[scale=0.65]{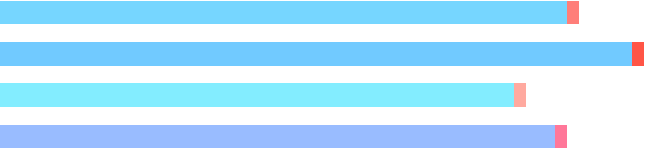}\\[3mm]
Checkpoint ensemble & Multi-run ensemble\\
\end{tabular}
\end{center}
\caption{Two methods to generate alternative systems for ensembling: Checkpoint ensembling uses model dumps from various stages of the training process, while multi-run ensembling starts independent training runs with different initial weights and order of the training data.}
\label{fig:ensemble-generation}
\end{figure}

See Figure~\ref{fig:ensemble-generation} for an illustration of two methods for the first sub-problem, generating alternative system. When training a neural translation model, we iterate through the training data until some stopping criteria is met. This is typically a lack of improvements of the cost function applied to a validation set (measured in cross-entropy), or the translation performance on that validation set (measured in {\sc bleu}).

During training, we dump out the model at fixed intervals (say, every 10,000 iteration of batch processing). Once training is completed, we can look back at the performance of the model these different stages. We then pick the, say, 4 models with the best performance (typically translation quality measured in {\sc bleu}). This is called {\bf checkpoint ensembling}\index{ensemble!checkpoint}\index{checkpoint ensemble} since we select the models at different checkpoints in the training process.

{\bf Multi-run ensembling}\index{ensemble!multi-run}\index{multi-run ensemble} requires building systems in completely different training runs. As mentioned before, this can be accomplished by using different random initialization of weights, which leads training to seek out different local optima. We also randomly shuffle the training data, so using different random order will also lead to different training outcomes. 

Multi-run ensembling usually works a good deal better, but it is also computationally much more expensive. Note that multi-run ensembling can also build on checkpoint ensembling. Instead of combining the end points of training, we first apply checkpoint ensembling to each run, and then combine those ensembles.

\subsubsection{Combine System Output}
Neural translation models allow the combination of several systems fairly deeply. Recall that the model first predicts a probability distribution over possible output words, and then commits to one of the words. This is where we combine the different trained models. Each model predicts a probability distribution and we then combine their predictions. The combination is done by simple averaging over the distributions. The averaged distribution is then the basis for selecting an output word.

\begin{figure}
\begin{center}
\includegraphics[scale=0.8]{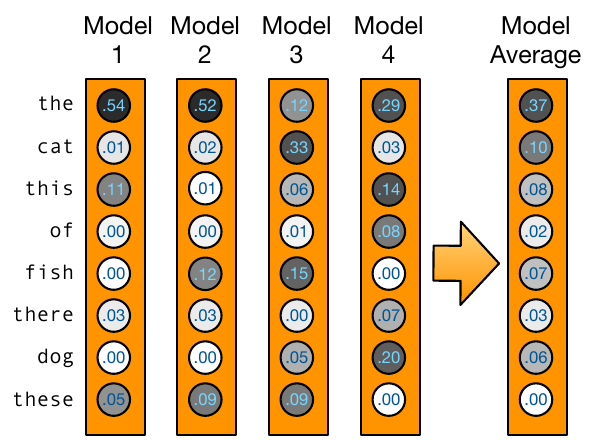}
\end{center}
\caption{Combining predictions from a ensemble of models: Each model independently predicts a probability distribution over output words, which are averaged into a combined distribution.}
\label{fig:ensemble-combination}
\end{figure}

See Figure~\ref{fig:ensemble-combination} for an illustration. There may be some benefit to weighing the different systems differently, although in our way of generating them, they will all have very similar quality, so this is not typically done.

\subsubsection{Reranking with Right-to-Left Decoding}
One more tweak on the idea of ensembling: Instead of building multiple systems with different random initialization, we can also build one set of system as before, and then a second set of system where we reverse the order of the output sentences. The second set of systems are called {\bf right-to-left}\index{right-to-left} systems, although arguably this is not a good name since it makes no sense for languages such as Arabic or Hebrew where the normal writing order is right to left.

The deep integration we described just above does not work anymore for the combination of left-to-right and right-to-left systems, since they produce output in different order. So, we have to resort to {\bf reranking}\index{reranking}. This involves several steps:
\begin{itemize}
\item Use an ensemble of left-to-right systems to generate an n-best list of candidate translations for each input sentence.
\item Score each candidate translation with the individual left-to-right and right-to-left systems.
\item Combine the scores (simple average) of the different models for each candidate, select the candidate with the best score for each input sentence.
\end{itemize}

Scoring a given candidate translation with a right-to-left system does require the require {\bf forced decoding}\index{forced decoding}\index{decoding!forced}, a special mode of running inference on an input sentence, but predicting a given output sentence. This mode is actually much closer to training (where also an output translation is given) the regular inference.


\subsection{Large Vocabularies}\label{sec:large-vocabulary}\index{vocabulary}
Zipf's law tells us that words in a language are very unevenly distribution. So, there is always a large tail of rare words. New words come into the language all the time (e.g., {\em retweeting, website, woke}), and we also have to deal with a very large inventory of names, including company names (e.g., {\em eBay, Yahoo, Microsoft}). 

On the other hand, neural methods are not well equipped to deal with such large vocabularies. The ideal representations for neural networks are continuous space vectors. This is why we first convert discrete objects such as words into such word embeddings. 

However, ultimately the discrete nature of words shows up. On the input side, we need to train an embedding matrix that maps each word into its embedding. On the output side we predict a probability distribution over all output words. The latter is generally the bigger concern, since the amount of computation involved is linear with the size of the vocabulary, making this a very large matrix operation.

Hence, neural translation models typically restrict the vocabulary to, say, 20,000 to 80,000 words. In initial work on neural machine translation, only the most frequent words were used, and all others represented by a {\em unknown} or {\em other} tag. The translation of these rare words was handled with a back-off dictionary.

The more common approach today is to break up rare words into {\bf subword units}\index{subword units}. This may seem a bit crude but is actually very similar to standard approaches in statistical machine translation to handle compounds (recall {\em website} $\rightarrow$ {\em web + site}) and morphology ({\em unfollow} $\rightarrow$ {\em un + follow},  {\em convolutions} $\rightarrow$ {\em convolution + s}). It is even a decent approach to the problem of transliteration of names
which are traditionally handled by a sub-modular letter translation component.

A popular method to create an inventory of subword units and legitimate words is {\bf byte pair encoding}\index{byte pair encoding}. This method is trained on the parallel corpus. First, the words in the corpus are split into characters (marking original spaces with a special space character). Then, the most frequent pair of characters is merged (in English, this may be {\em t} and {\em h} into {\em th}). This step is repeated for a fixed given number of times. Each of these steps increases the vocabulary by one, beyond the original inventory of single characters.

The example mirrors quite well the behavior of the algorithm on real-world data sets. It starts with grouping together with frequent letter combinations ({\em e+r}, {\em t+h}, {\em c+h}) and then joins frequent words ({\em the}, {\em in}, {\em of}). 
At the end of this process, the most frequent words will emerge as single tokens, while rare words consist of still un-merged subwords. See Figure~\ref{fig:bpe-example} for an example, where subword units are indicated with two ``at" symbols (@@). After 49,500 byte pair encoding operations, the vast majority of words are intact, while rarer words are broken up (e.g., {\em critic@@ ises, destabil@@ ising}). Sometimes, the split seem to be morphologically motivated (e.g., {\em im@@ pending}), but mostly they are not (e.g., {\em stra@@ ined}). Note also the decomposition of the relatively rare name {\em Net@@ any@@ ahu}.

\begin{figure}
{\em 
\begin{tabular}{|cp{14.7cm}c|}\hline
&& \\[-1mm]
&Obama receives \textcolor{darkgreen}{Net@@ any@@ ahu}&\\
&& \\[-4mm]
&the relationship between Obama and \textcolor{darkgreen}{Net@@ any@@ ahu} is not exactly
friendly .
the two wanted to talk about the implementation of the international agreement and about Teheran 's \textcolor{darkgreen}{destabil@@ ising} activities in the Middle East .
the meeting was also planned to cover the conflict with the Palestinians and the disputed two state solution .
relations between Obama and \textcolor{darkgreen}{Net@@ any@@ ahu} have been \textcolor{darkgreen}{stra@@ ined} for years .
Washington \textcolor{darkgreen}{critic@@ ises} the continuous building of settlements in Israel and \textcolor{darkgreen}{acc@@ uses} \textcolor{darkgreen}{Net@@ any@@ ahu} of a lack of initiative in the peace process .
the relationship between the two has further deteriorated because of the deal that Obama negotiated on Iran 's atomic programme .
in March , at the invitation of the \textcolor{darkgreen}{Republic@@ ans} , \textcolor{darkgreen}{Net@@ any@@ ahu} made a controversial speech to the US Congress , which was partly seen as an \textcolor{darkgreen}{aff@@ ront} to Obama .
the speech had not been agreed with Obama , who had rejected a meeting with reference to the election that was at that time \textcolor{darkgreen}{im@@ pending} in Israel .&\\[-1mm]
&&\\\hline
\end{tabular}}
\vspace{2mm}
\caption{Byte pair encoding (BPE) applied to English (model used 49,500 BPE operations). Word splits are indicated with @@. Note that the data is also tokenized and true-cased.}
\label{fig:bpe-example}
\end{figure}

\paragraph{Further Readings}
\furtherreadings{A significant limitation of neural machine translation models is the computational burden to support very large vocabularies. To avoid this, typically the vocabulary is reduced to a shortlist of, say, 20,000 words, and the remaining tokens are replaced with the unknown word token ``UNK". To translate such an unknown word, \cite{luong-EtAl:2015:ACL-IJCNLP,jean-EtAl:2015:ACL-IJCNLP} resort to a separate dictionary. 
\cite{arthur-neubig-nakamura:2016:EMNLP2016} argue that neural translation models are worse for rare words and interpolate a traditional probabilistic bilingual dictionary with the prediction of the neural machine translation model. They use the attention mechanism to link each target word to a distribution of source words and weigh the word translations accordingly. 

Source words such as names and numbers may also be directly copied into the target. \cite{gulcehre-EtAl:2016:P16-1} use a so-called switching network to predict either a traditional translation operation or a copying operation aided by a softmax layer over the source sentence. They preprocess the training data to change some target words into word positions of copied source words. Similarly, \cite{gu-EtAl:2016:P16-1} augment the word prediction step of the neural translation model to either translate a word or copy a source word. They observe that the attention mechanism is mostly driven by semantics and the language model in the case of word translation, but by location in case of copying.
 
To speed up training, \cite{mi-wang-ittycheriah:2016:P16-2} use traditional statistical machine translation word and phrase translation models to filter the target vocabulary for mini batches. 

\cite{sennrich-haddow-birch:2016:P16-12} split up all words to sub-word units, using character n-gram models and a segmentation based on the byte pair encoding compression algorithm.}

\subsection{Using Monolingual Data}\label{sec:monolingual-data}\index{monolingual data}\index{data!monolingual}
A key feature of statistical machine translation system are language models, trained on very large monolingual data set. The larger the language models, the higher translation quality. Language models trained on up to a trillion words crawled from the general web have been used. So, it is a surprise that the basic neural translation model does not use any additional monolingual data, its language model aspect (the conditioning of the previous hidden decoder state and the previous output) is trained jointly with the translation model aspect (the conditioning on the input context).

Two main ideas have been proposed to improve neural translation models with monolingual data. One is to transform additional monolingual translation into parallel data by synthesizing the missing half of the data, and the other is to integrate a language model as a component into the neural network architecture.

\subsubsection{Back Translation}\label{sec:backtranslation}
Language models improve fluency of the output. Using larger amounts of monolingual data in the target language give the machine more evidence what are common sequences of words and what are not. 

\begin{figure}
\begin{center}
\includegraphics[scale=1]{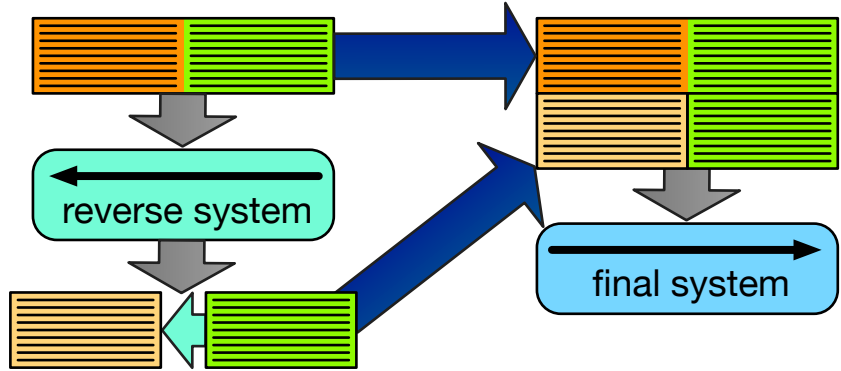}
\end{center}
\caption{Creating synthetic parallel data from target-side monolingual data: (1) train a system in reverse order, (2) use it to translate target-side monolingual data into the source language, (3) combine the generated synthetic parallel data with the true parallel data in final system building.}
\label{fig:backtranslation}
\end{figure}

We cannot use monolingual target side data in our neural translation model training, since it is missing the source side. So, one idea is to just synthesize this data by {\bf back translation}\index{back translation}. See Figure~\ref{fig:backtranslation} for an illustration of the steps involved.

\begin{itemize}
\item Train a reverse system that translates from the intended target language into the source language. We typically use the same neural machine translation setup for this as for our final system, just with source and target flipped. But we may use any system, even traditional phrase-based systems.
\item Use the reverse system to translate target side monolingual data, creating a {\bf synthetic parallel corpus}\index{synthetic parallel corpus}\index{data!synthetic}.
\item Combine the generated synthetic parallel data with the true parallel data when building the final system.
\end{itemize}

There is an open question on how much synthetic parallel data should be used in relation to the amount of existing true parallel data. Typically, there are magnitudes more monolingual data available, but we also do not want to drown out the actual real data. Successful applications of this idea used equal amounts of synthetic and true data. We may also generate much more synthetic parallel data, but then ensure during training that we process equal amounts of each by over-sampling the true parallel data.

\subsubsection{Adding a Language Model}
The other idea is to train a language model as a separate component of the neural translation model.
\cite{DBLP:journals/corr/GulcehreFXCBLBS15} first train the large language model as a recurrent neural network on all available data, including the target side of the parallel corpus. Then, they add this language model to the neural translation model. Since both language model and translation model predict output words, the natural point to connect the two models is joining them at that output prediction node in the network by concatenating their conditioning contexts.

We expand Equation~\ref{eq:nn:neural-translation-final} to add the hidden state of the neural language model $s^\text{\sc lm}_i$ to the hidden state of the neural translation model $s^\text{\sc tm}_i$, the source context $c_i$ and the previous English word $e_{i-1}$.
\begin{equation}
e_i = g(c_i,s^\text{\sc tm}_i,s^\text{\sc lm}_i,e_{i-1})
\label{eq:nn:tm-lm-combination}
\end{equation}

When training the combined model, we leave the parameters of the large neural language model unchanged, and update only the parameters of the translation model and the combination layer. The concern is that otherwise the output side of the parallel corpus would overwrite the memory of the large monolingual corpus. In other words, the language model would overfit to the parallel training data and be less general.

One final question remains: How much weight should be given to the translation model and how much weight should be given to the language model? The above equation considers them in all instances the same way. But there may be output words for which the translation model is more relevant (e.g., the translation of content words with distinct meaning) and output words where the language model is more relevant (e.g., the introduction of relevant function words for fluency).

The balance of the translation model and the language model can be achieved with the type of gated units that we encountered in our discussion of the long short-term memory neural network architecture (Section~\ref{sec:nn:lstm}). Such a gated unit may be predicted solely from the language model state $s^\text{\sc lm}_i$ and then used as a factor that is multiplied with that language model state before it is used in the prediction of Equation~\ref{eq:nn:tm-lm-combination}.

\begin{equation}
\begin{aligned}
\text{gate}^\text{\sc lm}_i &= f(s^\text{\sc lm}_i)\\
\bar{s}^{\text{\sc lm}}_i &= \text{gate}^\text{\sc lm}_i \times s^\text{\sc lm}_i\\
e_i &= g(c_i,s^\text{\sc tm}_i,\bar{s}^{\text{\sc lm}}_i,e_{i-1})\\
\end{aligned}
\end{equation}

\subsubsection{Round Trip Training}
Looking at the backtranslation idea from a strict machine learning perspective, we can see two learning objectives. There is the objective to learn the transformations given by the parallel data, as done traditionally. Then, there is the goal to learn how to convert an output lamguage sentence into the input language, and then back into the output language with the objective to match the traditional sentence. A good machine translation model should be able to preserve the meaning of the output language sentence when mapped into the input language and back.

\begin{figure}
\begin{center}
\includegraphics[scale=0.5]{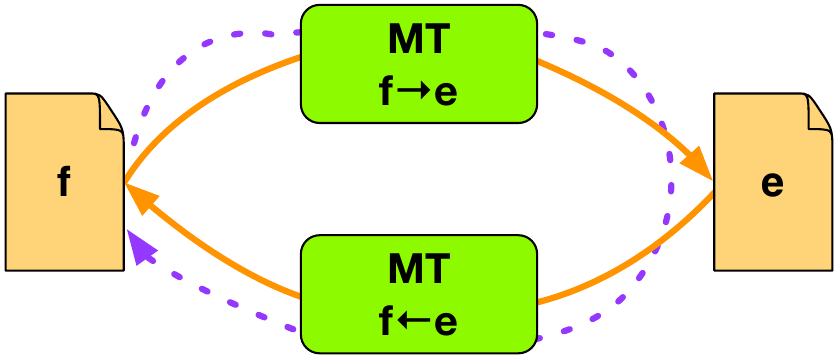}
\end{center}
\caption{Round trip training: In addition to training two models  {\em f$\rightarrow$e} and {\em e$\rightarrow$f}, as done traditionally on parallel data, we also optimize both models to convert a sentence $f$ into $e$ and then restore it back into $f$, using on monolingual data in $f$. We may add a corresponding round trip starting with $e$.}
\label{fig:round-trip-training}
\end{figure}

See Figure~\ref{fig:round-trip-training} for an illustration. There are two machine translation models. One that translates sentences in the language direction {\em f$\rightarrow$e}, the other in the opposite direction {\em e$\rightarrow$f}. These two systems may be trained with traditional means, using a parallel corpus. We can also {\bf round trip}\index{round trip} a sentence $f$ first through the {\em f$\rightarrow$e} and then back through  {\em e$\rightarrow$f}

In this scenario, there are two objectives for model training.
\begin{itemize}
\item The translation {\bf e'} of the given monolingual sentence {\bf f} should be a valid sentence in the language $e$, as measured with a language model $\text{\sc lm}_e(\text{\bf e'})$.
\item The reconstruction of the translation {\bf e'} back into the original language $f$ should be easy, as measured with the translation model $\text{\sc mt}_{e\rightarrow f}(\text{\bf f}|\text{\bf e'})$
\end{itemize}
These two objectives can be used to update model parameters in both translation models $\text{\sc mt}_{f\rightarrow e}$ and $\text{\sc mt}_{e\rightarrow f}$. 

Typical model update is driven by correct predictions of each word. In this round-trip scenario, the translation {\bf e'} has to be computed first, before we can do the usual training of model $\text{\sc mt}_{e\rightarrow f}$ with the given sentence pair  ({\bf e'},{\bf f}). To make better use of the training data, a n-best list of translations $\text{\bf e'}_1,...,\text{\bf e'}_n$ is computed and model updates are computed for each of them.

We can also update the model $\text{\sc mt}_{f\rightarrow e}$ with monolingual data in language $f$ by scaling updates by the language model cost $\text{\sc lm}_e(\text{\bf e'}_i)$ and the forward translation cost $\text{\sc mt}_{f\rightarrow e}(\text{\bf e'}_i|\text{\bf f})$ for each of the translations $\text{\bf e'}_i$ in the n-best list. 

To use monolingual data in language $e$, training is done in the reverse round trip direction. For details of this idea, refer to \cite{DBLP:journals/corr/XiaHQWYLM16}.

\paragraph{Further Readings}
\furtherreadings{\cite{sennrich-haddow-birch:2016:P16-11} back-translate the monolingual data into the input language and use the obtained synthetic parallel corpus as additional training data.
\cite{DBLP:journals/corr/XiaHQWYLM16} use monolingual data in a dual learning\index{dual learning} setup. Machine translation engines are trained in both directions, and in addition to regular model training from parallel data, monolingual data is translated in a round trip ({\em e} to {\em f} to {\em e}) and evaluated with a language model for language {\em f} and reconstruction match back to {\em e} as cost function to drive gradient descent updates to the model.}

\subsection{Deep Models}\label{sec:deep-models}\index{deep model}\index{model!deep}
Learning the lessons from other research fields such as vision or speech recognition, recent work in machine translation has also looked at deeper models. Simply put, this involves adding more intermediate layers into the baseline architecture.

The core components of neural machine translation are the encoder that takes input words and converts them into a sequence of contextualized representations and the decoder that generates a output sequence of words. Both are recurrent neural networks.

Recall that we already discussed how to build deeper recurrent neural networks for language modelling (refer back to Section~\vref{sec:deep-rnn}). We now extend these ideas to the recurrent neural networks in the encoder and the decoder. 

What all these recurrent neural networks have in common is that they process an input sequence into an output sequence, and at each time step $t$ information from a new input $x_t$ is combined with the hidden state from the previous time step $h_{t-1}$ to predict a new hidden state $h_t$. From that hidden state additional predictions may be made (output words $y_t$ in the case of the decoder, the next word in the sequence in the case of language models), or the hidden state is used otherwise (via the attention mechanism in case of the encoder).

\begin{figure}
\begin{center}
\includegraphics[scale=0.8]{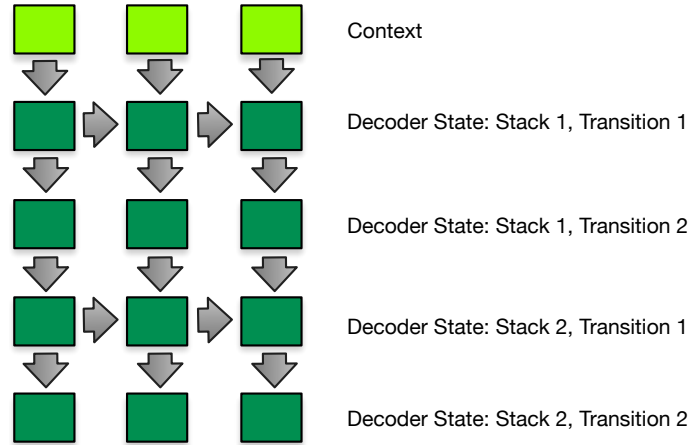}
\end{center}
\caption{Deep Decoder: Instead of a single recurrent neural network (RNN) layer for the decoder state, in a deep model, it consists of several layers. The illustrations shows a combination of a deep transition and stacked RNNs. It omits the word prediction, word selection and output word embedding steps which are identical to the original architecture, shown in Figure~\vref{fig:nn-attention-model-with-output-words4-step2-detail}.}
\label{fig:deep-decoder}
\end{figure}

\paragraph{Decoder}
See Figure~\ref{fig:deep-decoder} for part of the decoder in neural machine translation, using a particular deeper architecture. We see that instead of a single hidden state $h_t$ for a given time step $t$, we now have a sequence of hidden states $h_{t,1}$, $h_{t,2}$, ..., $h_{t,I}$ for a given time step $t$. 

There are various options how the hidden states may be connected. Previously, in Section~\ref{sec:deep-rnn} we presented two ideas. (1) In stacked recurrent neural networks where a hidden state $h_{t,i}$ is conditioned on the hidden state from a previous layer $h_{t,i-1}$ and the hidden state at the same depth from a previous time step $h_{t-1,i}$.  (2) In 
deep transition recurrent neural networks, the first hidden state $h_{t,1}$ is conditioned on the last hidden state from the previous time step $h_{t-1,I}$ and the input, while the other hidden layers $h_{t,I}$ ($i>1$) are just conditioned on the previous previous layer $h_{t,i-1}$.

Figure~\ref{fig:deep-decoder} combines these two ideas. some layers are both stacked (conditioned on the previous time step  $h_{t-1,I}$ and previous layer  $h_{t,i-1}$), while others are deep transitions (conditioned only on the previous layer $h_{t,i-1}$.

Mathematically, we can break this out into the stacked layers $h_{t,i}$:
\begin{equation}
\begin{aligned}
h_{t,1} &= f_1(x_t,h_{t-1,1})\\
h_{t,i} &= f_i(h_{t,i-1},h_{t-1,i}) & \text{for $i>1$}\\[3mm]
\end{aligned}
\end{equation}

and the deep transition layers $v_{i,i,j}$.
\begin{equation}
\begin{aligned}
v_{t,i,1} & = g_{i,1}(\text{in}_{t,i},h_{t-1,i}) & \text{in}_{t,i} \; \text{is either $x_t$ or $h_{t,i-1}$}\\
v_{t,i,j} & = g_{i,j}(v_{t,i,j-1}) & \text{for $j>1$}\\
h_{t,i} &= v_{t,i,J}
\end{aligned}
\end{equation}

The function $f_i(h_{t,i-1},h_{t-1,i})$ is computed as a sequence of function calls $g_{i,j}$.
Each of the functions $g_{i,j}$ may be implemented as feed-forward neural network layer (matrix multiplication plus activation function), long-short term memory cell (LSTM), or gated recurrent unit (GRU). On either case, each function $g_{i,j}$ has its own set of trainable model parameters.

\paragraph{Encoder}
Deep recurrent neural networks for the encoder may draw in the same ideas as the decoder, with one addition: in the baseline neural translation model, we used bidirectional recurrent neural networks to condition on both left and right context. We want to do the same for any deep version of the encoder.

\begin{figure}
\begin{center}
\includegraphics[scale=0.8]{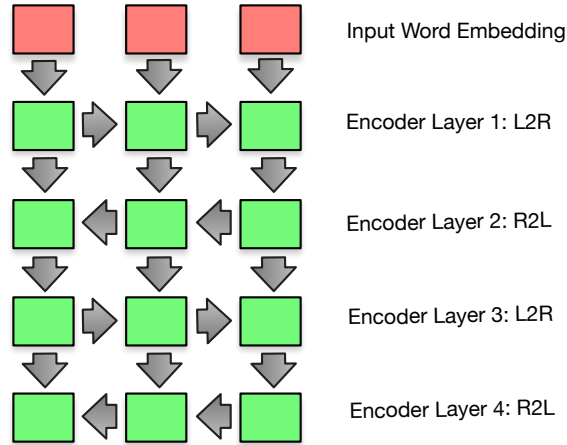}
\end{center}
\caption{Deep alternating encoder: combination of the idea of a bidirectional recurrent neural network previously proposed for neural machine translation (recall Figure~\vref{fig:nn-attention-model-input-encoding}) and the stacked recurrent neural network (recall Figure~\vref{fig:nn-attention-model-input-encoding}). This architecture may be further extended with the idea of deep transitions, as shown for the decoder (previous Figure~\ref{fig:deep-decoder}).}
\label{fig:deep-encoder}
\end{figure}

Figure~\ref{fig:deep-encoder} shows one idea how this could be done, called {\bf alternating recurrent neural network}\index{alternating recurrent neural network}\index{recurrent neural network!alternating}\index{neural network!alternating recurrent}.
It looks basically like a stacked recurrent neural network, with one twist: the hidden states at each layer $h_{t,i}$ are alternately conditioned on the hidden state from the previous time step $h_{t-1,i}$ or the next time step $h_{t+1,i}$.

Mathematically, we formulate this as even numbered hidden states $h_{t,2i}$ being conditioned on the left context $h_{t-1,2i}$ and odd numbered hidden states $h_{t,2i+1}$ conditioned on the right context $h_{t+1,2i}$.
\begin{equation}
\begin{aligned}
h_{t,1} &= f(x_t,h_{t-1,1})\\
h_{t,2i} &= f(h_{t,2i-1},h_{t-1,2i})\\
h_{t,2i+1} &= f(h_{t,2i},h_{t+1,2i+1})
\end{aligned}
\end{equation}

As before in the encoder, we can extend this idea by having deep transitions.

Note that deep models are typically augmented with direct connections from the input to the output. In the case of the encoder, this may mean a direct connection from the embedding to the final encoder layer, or connections at each layer that pass the input directly to the output. Such {\bf residual connections}\index{residual connection} help with training. In early stages, the deep architecture can be skipped. Only when a basic functioning model has been acquired, the deep architecture can be exploited to enrich it. We typically see the benefits of residual connections in early training stages (faster initial reduction of model perplexity), and less so as improvement in the final converged model.

\paragraph{Further Readings}
\furtherreadings{Recent work has shown good results with 4 stacks and 2 deep transitions each for encoder and decoder, as well as alternating networks for the encoder \citep{micelibarone-EtAl:2017:WMT}. There are a large number of variations (including the use of skip connections, the choice of LSTM vs. GRU, number of layers of any type) that still need to be explored empirical for various data conditions.}

\subsection{Guided Alignment Training}\label{sec:guided-alignment}
\begin{figure}
\begin{center} 
\begin{tikzpicture}[scale=0.55]
\node[label=above:\rotatebox{90}{\em relations}] at (0.5,11) {};
\node[label=above:\rotatebox{90}{\em between}] at (1.5,11) {};
\node[label=above:\rotatebox{90}{\em Obama}] at (2.5,11) {};
\node[label=above:\rotatebox{90}{\em and}] at (3.5,11) {};
\node[label=above:\rotatebox{90}{\em Netanyahu}] at (4.5,11) {};
\node[label=above:\rotatebox{90}{\em have}] at (5.5,11) {};
\node[label=above:\rotatebox{90}{\em been}] at (6.5,11) {};
\node[label=above:\rotatebox{90}{\em strained}] at (7.5,11) {};
\node[label=above:\rotatebox{90}{\em for}] at (8.5,11) {};
\node[label=above:\rotatebox{90}{\em years}] at (9.5,11) {};
\node[label=above:\rotatebox{90}{.}] at (10.5,11) {};
\draw (0,10.5) node[anchor=east] {\em die};
\draw (0,9.5) node[anchor=east] {\em Beziehungen};
\draw (0,8.5) node[anchor=east] {\em zwischen};
\draw (0,7.5) node[anchor=east] {\em Obama};
\draw (0,6.5) node[anchor=east] {\em und};
\draw (0,5.5) node[anchor=east] {\em Netanjahu};
\draw (0,4.5) node[anchor=east] {\em sind};
\draw (0,3.5) node[anchor=east] {\em seit};
\draw (0,2.5) node[anchor=east] {\em Jahren};
\draw (0,1.5) node[anchor=east] {\em angespannt};
\draw (0,0.5) node[anchor=east] {.};
\fill[green!56!white] (0,10) rectangle (1,11);
\draw (0.5,10.5) node[align=center] {56};
\fill[green!89!white] (0,9) rectangle (1,10);
\draw (0.5,9.5) node[align=center] {89};
\fill[green!0!white] (0,8) rectangle (1,9);
\fill[green!0!white] (0,7) rectangle (1,8);
\fill[green!1!white] (0,6) rectangle (1,7);
\fill[green!0!white] (0,5) rectangle (1,6);
\fill[green!0!white] (0,4) rectangle (1,5);
\fill[green!0!white] (0,3) rectangle (1,4);
\fill[green!0!white] (0,2) rectangle (1,3);
\fill[green!0!white] (0,1) rectangle (1,2);
\fill[green!0!white] (0,0) rectangle (1,1);
\fill[green!3!white] (1,10) rectangle (2,11);
\fill[green!1!white] (1,9) rectangle (2,10);
\fill[green!72!white] (1,8) rectangle (2,9);
\draw (1.5,8.5) node[align=center] {72};
\fill[green!2!white] (1,7) rectangle (2,8);
\fill[green!2!white] (1,6) rectangle (2,7);
\fill[green!0!white] (1,5) rectangle (2,6);
\fill[green!0!white] (1,4) rectangle (2,5);
\fill[green!0!white] (1,3) rectangle (2,4);
\fill[green!0!white] (1,2) rectangle (2,3);
\fill[green!0!white] (1,1) rectangle (2,2);
\fill[green!0!white] (1,0) rectangle (2,1);
\fill[green!16!white] (2,10) rectangle (3,11);
\draw (2.5,10.5) node[align=center] {16};
\fill[green!0!white] (2,9) rectangle (3,10);
\fill[green!26!white] (2,8) rectangle (3,9);
\draw (2.5,8.5) node[align=center] {26};
\fill[green!96!white] (2,7) rectangle (3,8);
\draw (2.5,7.5) node[align=center] {96};
\fill[green!1!white] (2,6) rectangle (3,7);
\fill[green!0!white] (2,5) rectangle (3,6);
\fill[green!0!white] (2,4) rectangle (3,5);
\fill[green!0!white] (2,3) rectangle (3,4);
\fill[green!0!white] (2,2) rectangle (3,3);
\fill[green!0!white] (2,1) rectangle (3,2);
\fill[green!0!white] (2,0) rectangle (3,1);
\fill[green!0!white] (3,10) rectangle (4,11);
\fill[green!0!white] (3,9) rectangle (4,10);
\fill[green!0!white] (3,8) rectangle (4,9);
\fill[green!0!white] (3,7) rectangle (4,8);
\fill[green!79!white] (3,6) rectangle (4,7);
\draw (3.5,6.5) node[align=center] {79};
\fill[green!0!white] (3,5) rectangle (4,6);
\fill[green!0!white] (3,4) rectangle (4,5);
\fill[green!0!white] (3,3) rectangle (4,4);
\fill[green!0!white] (3,2) rectangle (4,3);
\fill[green!0!white] (3,1) rectangle (4,2);
\fill[green!0!white] (3,0) rectangle (4,1);
\fill[green!2!white] (4,10) rectangle (5,11);
\fill[green!0!white] (4,9) rectangle (5,10);
\fill[green!0!white] (4,8) rectangle (5,9);
\fill[green!0!white] (4,7) rectangle (5,8);
\fill[green!0!white] (4,6) rectangle (5,7);
\fill[green!98!white] (4,5) rectangle (5,6);
\draw (4.5,5.5) node[align=center] {98};
\fill[green!1!white] (4,4) rectangle (5,5);
\fill[green!2!white] (4,3) rectangle (5,4);
\fill[green!0!white] (4,2) rectangle (5,3);
\fill[green!0!white] (4,1) rectangle (5,2);
\fill[green!0!white] (4,0) rectangle (5,1);
\fill[green!2!white] (5,10) rectangle (6,11);
\fill[green!0!white] (5,9) rectangle (6,10);
\fill[green!0!white] (5,8) rectangle (6,9);
\fill[green!0!white] (5,7) rectangle (6,8);
\fill[green!4!white] (5,6) rectangle (6,7);
\fill[green!0!white] (5,5) rectangle (6,6);
\fill[green!42!white] (5,4) rectangle (6,5);
\draw (5.5,4.5) node[align=center] {42};
\fill[green!3!white] (5,3) rectangle (6,4);
\fill[green!0!white] (5,2) rectangle (6,3);
\fill[green!1!white] (5,1) rectangle (6,2);
\fill[green!11!white] (5,0) rectangle (6,1);
\draw (5.5,0.5) node[align=center] {11};
\fill[green!0!white] (6,10) rectangle (7,11);
\fill[green!0!white] (6,9) rectangle (7,10);
\fill[green!0!white] (6,8) rectangle (7,9);
\fill[green!0!white] (6,7) rectangle (7,8);
\fill[green!2!white] (6,6) rectangle (7,7);
\fill[green!0!white] (6,5) rectangle (7,6);
\fill[green!11!white] (6,4) rectangle (7,5);
\draw (6.5,4.5) node[align=center] {11};
\fill[green!2!white] (6,3) rectangle (7,4);
\fill[green!0!white] (6,2) rectangle (7,3);
\fill[green!4!white] (6,1) rectangle (7,2);
\fill[green!14!white] (6,0) rectangle (7,1);
\draw (6.5,0.5) node[align=center] {14};
\fill[green!6!white] (7,10) rectangle (8,11);
\fill[green!4!white] (7,9) rectangle (8,10);
\fill[green!0!white] (7,8) rectangle (8,9);
\fill[green!0!white] (7,7) rectangle (8,8);
\fill[green!4!white] (7,6) rectangle (8,7);
\fill[green!0!white] (7,5) rectangle (8,6);
\fill[green!38!white] (7,4) rectangle (8,5);
\draw (7.5,4.5) node[align=center] {38};
\fill[green!22!white] (7,3) rectangle (8,4);
\draw (7.5,3.5) node[align=center] {22};
\fill[green!0!white] (7,2) rectangle (8,3);
\fill[green!84!white] (7,1) rectangle (8,2);
\draw (7.5,1.5) node[align=center] {84};
\fill[green!23!white] (7,0) rectangle (8,1);
\draw (7.5,0.5) node[align=center] {23};
\fill[green!8!white] (8,10) rectangle (9,11);
\fill[green!1!white] (8,9) rectangle (9,10);
\fill[green!0!white] (8,8) rectangle (9,9);
\fill[green!0!white] (8,7) rectangle (9,8);
\fill[green!1!white] (8,6) rectangle (9,7);
\fill[green!0!white] (8,5) rectangle (9,6);
\fill[green!1!white] (8,4) rectangle (9,5);
\fill[green!54!white] (8,3) rectangle (9,4);
\draw (8.5,3.5) node[align=center] {54};
\fill[green!0!white] (8,2) rectangle (9,3);
\fill[green!0!white] (8,1) rectangle (9,2);
\fill[green!0!white] (8,0) rectangle (9,1);
\fill[green!1!white] (9,10) rectangle (10,11);
\fill[green!0!white] (9,9) rectangle (10,10);
\fill[green!0!white] (9,8) rectangle (10,9);
\fill[green!0!white] (9,7) rectangle (10,8);
\fill[green!0!white] (9,6) rectangle (10,7);
\fill[green!0!white] (9,5) rectangle (10,6);
\fill[green!0!white] (9,4) rectangle (10,5);
\fill[green!10!white] (9,3) rectangle (10,4);
\draw (9.5,3.5) node[align=center] {10};
\fill[green!98!white] (9,2) rectangle (10,3);
\draw (9.5,2.5) node[align=center] {98};
\fill[green!0!white] (9,1) rectangle (10,2);
\fill[green!0!white] (9,0) rectangle (10,1);
\fill[green!1!white] (10,10) rectangle (11,11);
\fill[green!0!white] (10,9) rectangle (11,10);
\fill[green!0!white] (10,8) rectangle (11,9);
\fill[green!0!white] (10,7) rectangle (11,8);
\fill[green!1!white] (10,6) rectangle (11,7);
\fill[green!0!white] (10,5) rectangle (11,6);
\fill[green!2!white] (10,4) rectangle (11,5);
\fill[green!2!white] (10,3) rectangle (11,4);
\fill[green!0!white] (10,2) rectangle (11,3);
\fill[green!7!white] (10,1) rectangle (11,2);
\fill[green!49!white] (10,0) rectangle (11,1);
\draw (10.5,0.5) node[align=center] {49};
\draw[blue,very thick] (0,9) rectangle (1,10);
\draw[blue,very thick] (1,8) rectangle (2,9);
\draw[blue,very thick] (2,7) rectangle (3,8);
\draw[blue,very thick] (3,6) rectangle (4,7);
\draw[blue,very thick] (4,5) rectangle (5,6);
\draw[blue,very thick] (5,3) rectangle (6,4);
\draw[blue,very thick] (6,4) rectangle (7,5);
\draw[blue,very thick] (6,3) rectangle (7,4);
\draw[blue,very thick] (7,1) rectangle (8,2);
\draw[blue,very thick] (8,3) rectangle (9,4);
\draw[blue,very thick] (9,2) rectangle (10,3);
\draw[blue,very thick] (10,0) rectangle (11,1);
\end{tikzpicture}
\end{center}
\caption{Alignment vs. Attention: In this example, alignment points from traditional word alignment methods are shown as squares, and attention states as shaded boxes depending on the alignment value (shown as percentage). They generally match up well, but note for instance that the prediction of the output auxiliary verb {\em sind} pays attention to the entire verb group {\em have been strained}.} 
\label{fig:alignment-vs-attention}
\end{figure}
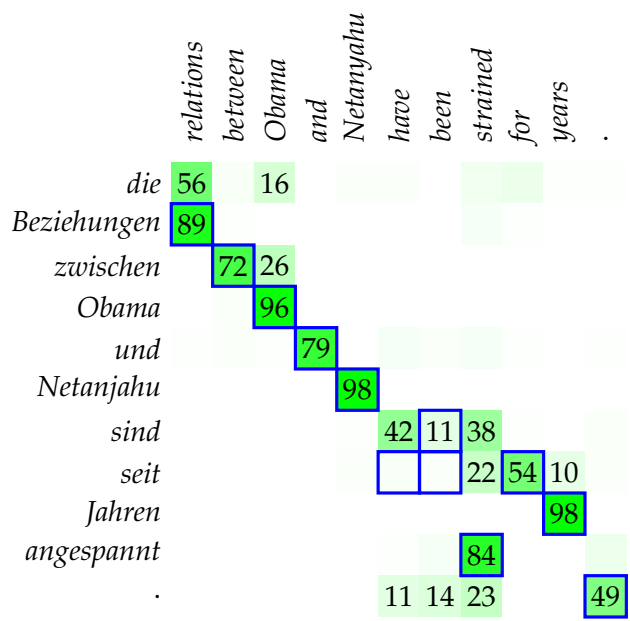

\index{guided alignment training}
The attention mechanism in neural machine translation models is motivated by the need to align output words to input words. Figure~\ref{fig:alignment-vs-attention} shows an example of attention weights given to English input words for each German output word during the translation of a sentence.

The attention values typically match up pretty well with word alignment used in traditional statistical machine translation, obtained with tools such as GIZA++ or fast-align which implement variants of the IBM Models. 

There are several good uses for word alignments beyond their intrinsic value of improving the quality of translations. For instance in the next section, we will look at using the attention mechanism to explicitly track coverage of the input. We may also want to override preferences of the neural machine translation model with pre-specified translations of certain terminology or expressions such as numbers, dates, or measurements that are better handled by rule-based components; this requires to know when the neural model is about to translate a specific source word. But also the end user may be interested in alignment information, such as translators using machine translation in a computer aided translation tool may want to check where an output word originates from.

Hence, instead of trusting the attention mechanism to implicitly acquire the role as word alignmer, we may enforce this role. The idea is to provide not just the parallel corpus as training data, but also pre-computed word alignments using traditional means. Such additional information may even benefit training of models to converge faster or overcome data sparsity under low resource conditions.

A straightforward way to add such given word alignment to the training process is to not change the model at all, but to just modify the training objective. Typically, the goal of training neural machine translation models is to generate the correct output words. We can add to this goal to also match the given word alignment.

Formally, we assume to have access to an alignment matrix $A$ that specifies alignment points $A_{ij}$ input words $j$ and output words $i$ in a way that $\sum_j A_{ij} = 1$, i.e., each output word's alignment scores add up to 1. The model estimates attention scores $\alpha_{ij}$ that also add up to 1 for each output word: $\sum_j \alpha_{ij} = 1$ (recall Equation~\vref{eqn:normalized-attention}). The mismatch between given alignment scores $A_{ij}$ and computed attention scores $\alpha_{ij}$ can be measured in several ways, such as cross entropy
\begin{equation}
\text{cost}_\text{\sc ce} = - \frac{1}{I} \sum_{i=1}^I \sum_{j=1}^J A_{ij} \; \text{log} \; \alpha_{ij}
\end{equation}

or mean squared error
\begin{equation}
\text{cost}_\text{\sc mse} = - \frac{1}{I} \sum_{i=1}^I \sum_{j=1}^J (A_{ij} - \alpha_{ij})^2
\end{equation}

This cost is added to the training objective and may be weighted.

\paragraph{Further Readings}
\furtherreadings{\cite{DBLP:journals/corr/ChenMKP16,liu-EtAl:2016:COLING} add supervised word alignment information (obtained with traditional statistical word alignment methods) to training. They augment the objective function to also optimize matching of the attention mechanism to the given alignments.}

\subsection{Modeling Coverage}\label{sec:coverage}
One impressive aspect of neural machine translation models is how well they are able to translate the entire input sentence, even when a lot of reordering is involved. But this as aspect is not perfect, occasionally the model translates some input words multiple times, and sometimes it misses to translate them.

\begin{figure}
\begin{center}
\begin{tikzpicture}[scale=0.43]
\input{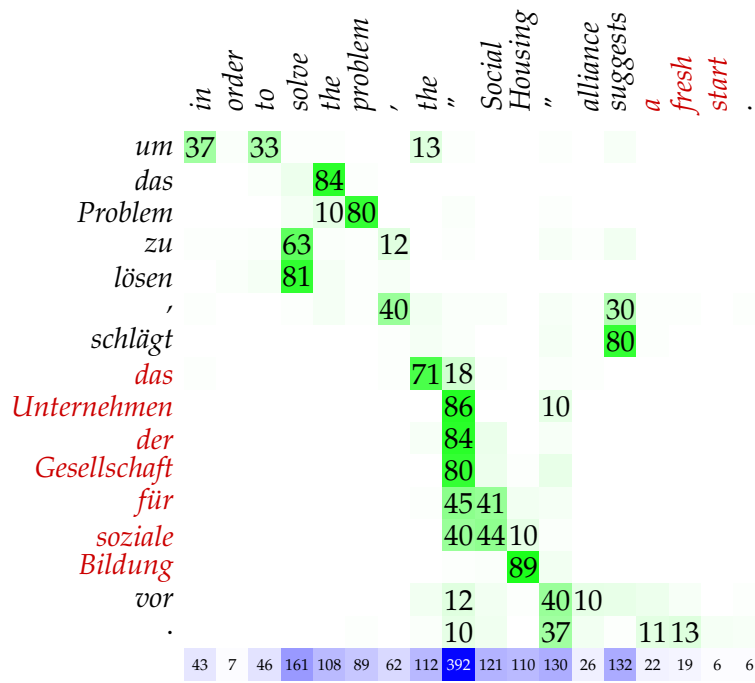}
\end{tikzpicture}
\end{center}
\caption{Example for over-generation and under-generation: the input tokens around {\em Social Housing} are attended too much, leading to hallucinated output words ({\em das Unternehmen}, English: {\em the company}), while the end of the sentence {\em a fresh start} is not attended and untranslated.}
\label{fig:over-under-attention}
\end{figure}

See Figure~\ref{fig:over-under-attention} for an example. The translation has two flaws related to mis-allocation of attention. The beginning of the phrase {\em ``Social Housing" alliance} receives too much attention, resulting in a faulty translation with hallucinated words: {\em das Unternehmen der Gesellschaft f\"ur soziale Bildung}, or {\em the company of the society for social education}. At the end of the input sentence, the phrase {\em a fresh start} does not receive any attention and is hence untranslated in the output.

Hence, an obvious idea is to more strictly model {\bf coverage}\index{coverage}. Given the attention model, a reasonable way to define coverage is by adding up the attention states. In a complete sentence translation, we roughly expect that each input word receives a similar amount of attention. If some input words never receive attention or too much attention that signals a problem with the translation.

\paragraph{Enforcing Coverage during Inference} We may restrict the enforcing of proper coverage to the decoder. When considering multiple hypothesis in beam search, then we should discourage the ones that pay too much attention to some input words. And, once hypotheses are completed, we can penalize those that paid only little attention to some of the input. There are various ways to come up with scoring functions for over-generation and under-generation.

\begin{equation}
\begin{aligned}
\text{coverage}(j) &= \sum_i \sum_k \alpha_{i,k}\\
\text{over-generation} &= \text{max} \Big(0, \sum_j \text{coverage}(j) - 1 \Big)\\
\text{under-generation} &= \text{min} \Big(1,  \sum_j \text{coverage}(j) \Big)\\
\end{aligned}
\end{equation}

The use of multiple scoring functions in the decoder is common practice in traditional statistical machine translation. For now, it is not in neural machine translation. A challenge is to give proper weight to the different scoring functions. If there are only two or three weights, these can be optimized with grid search over possible values. For more weights, we may borrow methods such as MERT or MIRA from statistical machine translation.

\paragraph{Coverage Models}
The vector that accumulates coverage of input words may be directly used to inform the attention model. Previously, the attention given to a specific input word $j$ was conditioned on the previous state of the decoder $s_{i-1}$ and the representation of the input word $h_j$. Now, we also add as conditioning context the accumulated attention given to the word (compare to Equation~\vref{eqn:attention-model}).

\begin{equation}
a(s_{i-1},h_j) = W^as_{i-1} + U^ah_j + V^a \text{coverage}(j) + b^a
\end{equation}

Coverage tracking may also integrated into the training objective. Taking a page from the guided alignment training (recall the previous Section~\ref{sec:guided-alignment}), we augment the training objective function with a coverage penalty with some weight $\lambda$.
\begin{equation}
\log \sum_i P(y_i|x) + \lambda \sum_j (1-\text{coverage}(j))^2
\end{equation}
 
Note that in general, it is problematic to add such additional functions to the learning objective, since it does distract from the main goal of producing good translations. 
 
\paragraph{Fertility} 
So far, we described coverage as the need to cover all input words roughly evenly. However, even the earliest statistical machine translation models considered the {\bf fertility}\index{fertility} of words, i.e., the number of output words that are generated from each input word. Consider the English {\em do not} construction: most other language do not require an equivalent of {\em do} when negating a verb. Meanwhile, other words are translated into multiple output words. For instance, the German {\em nat{\"u}rlich} may be translated as {\em of course}, thus generating 2 output words.

We may augment models of coverage by adding a fertility components that predicts the number of output words for each input words. Here one example for a model that predicts the fertility $\Phi_j$ for each input word, and uses it to normalize the coverage statistics.
\begin{equation}
\begin{aligned}
\Phi_j &= N \sigma(W_jh_j)\\
\text{coverage}(j) &= \frac{1}{\Phi_j} \sum_i \sum_k \alpha_{i,k}\\
\end{aligned}
\end{equation}
Fertility $\Phi_j$ is predicted with a neural network layer that is conditioned on the input word representation $h_j$ and uses a sigmoid activation function (thus resulting in values from 0 to 1), which is scaled to a pre-defined maximum fertility of $N$.

\paragraph{Feature Engineering versus Machine Learning}
The work on modeling coverage in neural machine translation models is a nice example to contrast between the engineering approach and the belief in generic machine learning techniques. From an engineering perspective, a good way to improve a system is to analyze its performance, find weak points and consider changes to overcome them. Here, we notice over-generation and under-generation with respect to the input, and add components to the model to overcome this problem. On the other hand, proper coverage is one of the features of a good translation that machine learning should be able to get from the training data. If it is not able to do that, it may need deeper models, more robust estimation techniques, ways to fight over-fitting or under-fitting, or other adjustments to give it just the right amount of power needed for the problem. 

It is hard to carry out the analysis needed to make generic machine learning adjustments, given the complexity of a task like machine translation. Still, the argument for deep learning is that it does not require feature engineering, such as adding coverage models. It remains to be seen how neural machine translation evolves over the next years, and if it moves more into a engineering or machine learning direction.

\paragraph{Further Readings}
\furtherreadings{To better model coverage,  \cite{tu-EtAl:2016:P16-1} add coverage states for each input word by either (a) summing up attention values, scaled by a fertility value predicted from the input word in context, or (b) learning a coverage update function as a feed-forward neural network layer. This coverage state is added as additional conditioning context for the prediction of the attention state.
\cite{feng-EtAl:2016:COLING3} condition the prediction of the attention state also on the previous context state and also introduce a coverage state (initialized with the sum of input source embeddings) that aims to subtract covered words at each step. Similarly, \cite{meng-EtAl:2016:COLING} separate hidden states that keep track of source coverage and hidden states that keep track of produced output.
\cite{cohn-EtAl:2016:N16-1} add a number of biases to model coverage, fertility, and alignment inspired by traditional statistical machine translation models. They condition the prediction of the attention state on absolute word positions, the attention state of the previous output word in a limited window, and coverage (added attention state values) over a limited window. They also add a fertility model and add coverage in the training objective.}

\subsection{Adaptation}
Text may differ by style, topic, degree of formality, and so on.
A common problem in the practical development of machine translation systems is that most of the available training data is different from the data relevant to a chosen use case. For instance, if your goal is to translate chat room dialogs, you will realize that there is very little translated chat room data available. There are massive quantities of official publications from international organizations, random translations crawled from the web, and maybe somewhat relevant movie subtitle translations.

This problem is generally framed as a problem of {\bf domain adaptation}\index{adaptation}\index{domain adaptation}. In the simplest form, you have one set of data relevant to your use case --- the {\bf in-domain data}\index{in-domain data}\index{data!in-domain} --- and another set that is less relevant --- the {\bf out-of-domain data}\index{out-of-domain data}\index{data!out-of-domain}. 

In traditional statistical machine translation, a vast number of methods for domain adaptation have been proposed. Models may be interpolated, we may back-off from in-domain to out-of-domain models, we may over-sample in-domain data during training or sub-sample out-of-domain data, etc.

\begin{figure}
\begin{center}
\includegraphics[scale=0.9]{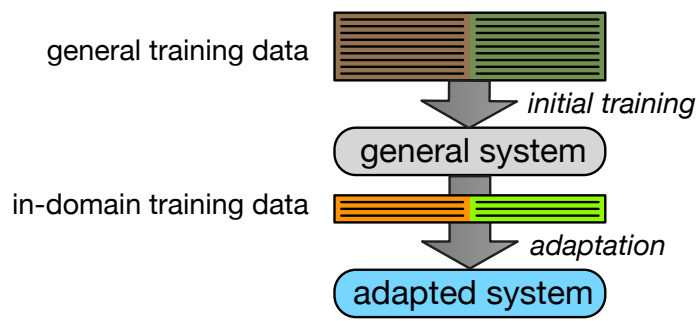}
\end{center}
\caption{Online training of neural machine translation models allows a straightforward domain adaptation method: Having a general domain translation system trained on general-purpose data, a handful of additional training epochs on in-domain data allows for a domain-adapted system.}
\label{fig:neural-adaptation}
\end{figure}

For neural machine translation, a fairly straightforward method is currently the most popular (see Figure~\ref{fig:neural-adaptation}). This method divides training up into two stages. First, we train the model on all available data until convergence. Then, we run a few more iterations of training on the in-domain data only and stop training when performance on the in-domain validate set peaks. This way, the final model benefits from all the training data, but is still specialized to the in-domain data.

Practical experience with this method shows that the second in-domain training stage may converge very quickly. The amount if in-domain data is typically relatively small, and only a handful of training epochs are needed.

Another, less commonly used method draws on the idea of ensemble decoding (Section~\ref{sec:ensemble}). If we train separate models on different sets of data, we may combine their predictions, just as we did for ensemble decoding. In this case, we do want to choose weights for each model, although how to choose these weights is not a trivial task. If there is just an in-domain and out-of-domain model, however, this may be simply done by line search over possible values.

Let us now look at a few special cases that arise in practical use. 

\paragraph{Subsample in-domain data from large collections}
A common problem is that the amount of available in-domain data is very small, so just training on this data, even in a secondary adaptation stage, risks overfitting --- very good performance on the seen data but poor performance on everything else.

Large random collections of parallel text often contain data that closely matches the in-domain data. So, we may want to extract this in-domain data from the large collections of mainly out-of-domain data. The general idea behind a variety of methods is to build two detectors: one in-domain detector trained on in-domain data, and one out-of-domain detector trained on out-of-domain data. We then score each sentence pair in the out-of-domain data with both detectors and select sentence pairs that are preferred (or judged relatively relevant) by the in-domain detector.

The classic detectors are language models trained on the source and target side of the in-domain and out-of-domain data, resulting in a total of 4 language models: the source side in-domain model $\text{LM}^\text{in}_f$,  the target side in-domain model $\text{LM}^\text{in}_e$,  the source side out-of-domain model $\text{LM}^\text{out}_f$, and  the target side out-of-domain model $\text{LM}^\text{out}_e$. Any given sentence pair from the out-of-domain data is then scored based on these models:
\begin{equation}
\text{relevance}_{e,f} = \Big( \text{LM}^\text{in}_e(e) - \text{LM}^\text{out}_e(e) \Big) + \Big( \text{LM}^\text{in}_f(f) - \text{LM}^\text{out}_f(f) \Big)
\end{equation}

We may use traditional n-gram language models or neural recurrent language models. Some work suggests to replace open class words (nouns, verbs, adjectives, adverbs) with part-of-speech tags or word clusters. More sophisticated models not only consider domain-relevance but noisiness of the training data (e.g., misaligned or mistranslated data). We may even use in-domain and out-of-domain neural translation models to score sentence pairs instead of source and target side sentences in isolation.

The subsampled data may be used in several ways. We may only train on this data to build our domain-specific system. Or, we use it in a secondary adaptation stage as outlined above.

\paragraph{Only monolingual in-domain data}
What if we have no parallel data in the domain of our use case? Two main ideas have been explored. Firstly, we may still use the monolingual data, may it be in the source or target language or both, for subsampling parallel data from a large pile of general data, as outline above.

Another idea is to use existing parallel data to train an out-of-domain model, then back-translate out-of-domain data (recall Section~\ref{sec:backtranslation}) to generate a synthetic in-domain corpus, and then use this data to adapt the initial model. In traditional statistical machine translation, much adaptation success has been achieved with just interpolating the language model, and this idea is the neural translation equivalent to that.

\paragraph{Multiple domains}
Sometimes, we have multiple collections of data that are clearly identified by domain --- typically categories such as information technology, medical, law, etc. We can use the techniques described above to build specialized translation models for each of these domains. 

For a given test sentence, we then select the appropriate model. If we do not know the domain of the test sentence, we first have to build a classifier that allows us to automatically make this determination. The classifier may be based on the methods for domain detectors described above. Given the decision of the classifier, we then select the most appropriate model.

But we do not have to commit to a single domain. The classifier may instead provide a distribution of relevance of the specific domain models (say, 50\% domain A, 30\% domain B, 20\% domain C) which are then used as weights in an ensemble of domain-specific models.

The domain classification may done based on a whole document instead of each individual sentence, which brings in more context to make a more robust decision.

As a final remark, it is hard to give conclusive advice on how to handle adaptation challenges, since it is such a broad topic. The style of the text may be more relevant than its content. Data may differ narrowly (e.g., official publications from the United Nations vs. official announcements from the European Union) or dramatically (e.g., chat room dialogs vs. published laws). The amounts of in-domain and out-of-domain data differs. The data may be cleanly separated by domain or just come in a massive disorganized pile. Some of the data may be of higher translation quality than other, which may be polluted by noise such as mistranslations, misalignments, or even generated by some other machine translation system.

\paragraph{Further Readings}
\furtherreadings{There is often a domain mismatch between the bulk (or even all) of the training data for a translation and its test data during deployment. There is rich literature in traditional statistical machine translation on this topic. A common approach for neural models is to first train on all available training data, and then run a few iterations on in-domain data only \citep{IWSLT-2015-Luong}, as already pioneered in neural language model adaption \citep{tersarkisov-EtAl:2015:CVSC}. \cite{DBLP:journals/corr/ServanCS16} demonstrate the effectiveness of this adaptation method with small in-domain sets consisting of as little as 500 sentence pairs. 

\cite{chu-dabre-kurohashi:2017:Short} argue that given small amount of in-domain data leads to overfitting and suggest to mix in-domain and out-of-domain data during adaption. \cite{Freitag:2016:unpublished} identify the same problem and suggest to use an ensemble of baseline models and adapted models to avoid overfitting. \cite{DBLP:journals/corr/PerisCC17} consider alternative training methods for the adaptation phase but do not find consistently better results than the traditional gradient descent training. Inspired by domain adaptation work in statistical machine translation on sub-sampling and sentence weighting, \cite{chen-EtAl:2017:NMT} build an in-domain vs. out-of-domain classifier for sentence pairs in the training data, and then use its prediction score to reduce the learning rate for sentence pairs that are out of domain.

\cite{farajian-EtAl:2017:EACLshort} show that traditional statistical machine translation outperforms neural machine translation when training general-purpose machine translation systems on a collection data, and then tested on niche domains. The adaptation technique allows neural machine translation to catch up.

A multi-domain model may be trained and informed at run-time about the domain of the input sentence. \cite{Kobus:2016:unpublished} apply an idea initially proposed by \cite{sennrich-haddow-birch:2016:N16-1} - to augment input sentences for register with a politeness feature token - to the domain adaptation problem. They add a domain token to each training and test sentence. \cite{DBLP:journals/corr/ChenMKP16} report better results over the token approach to adapt to topics by encoding the given topic membership of each sentence as an additional input vector to the conditioning context of word prediction layer.}

\subsection{Adding Linguistic Annotation}\index{linguistic annotation}
One of the big debates in machine translation research is the question if the key to progress is to develop better, relatively generic, machine learning methods that implicitly learn the important features of language, or to use linguistic insight to augment data and models. 

Recent work in statistical machine translation has demonstrated the benefits of linguistically motivated models. The best statistical machine translation systems in major evaluation campaigns for language pairs such as Chinese--English and German--English are syntax-based. While they translate sentences, they also build up the syntactic structure of the output sentence. There have been serious efforts to move towards deeper semantics in machine translation.

The turn towards neural machine translation was at first hard swing back towards better machine learning while ignoring much linguistic insights. Neural machine translation views translation as a generic sequence to sequence task, which just happens to involve sequences of words in different languages. Methods such as byte pair encoding or character-based translation models even put the value of the concept of a word as a basic unit into doubt. 

However, recently there have been also attempts to add linguistic annotation into neural translation models, and steps towards more linguistically motivated models. We will take a look at successful efforts to integrate (1) linguistic annotation to the input sentence, (2) linguistic annotation to the output sentence, and (3) build linguistically structured models.

\paragraph{Linguistic annotation of the input}
One of the great benefits of neural networks is their ability to cope with rich context. In the neural machine translation models we presented, each word prediction is conditioned on the entire input sentence and all previously generated output words. Even if, as it is typically the case, a specific input sequence and partially generated output sequence has never been observed before during training, the neural model is able to generalize the training data and draw from relevant knowledge. In traditional statistical models, this required carefully chosen independence assumptions and back-off schemes.

So, adding more information to the conditioning context in neural translation models can be accommodated rather straightforwardly. First, what information would be like to add? The typical linguistic treasure chest contains part-of-speech tags, lemmas, morphological properties of words, syntactic phrase structure, syntactic dependencies, and maybe even some semantic annotation.

All of these can be formatted as annotations to individual input words. Sometimes, this requires a bit more work, such as syntactic and semantic annotation that spans multiple words. See Figure~\ref{fig:linguistic-input-annotation} for an example. To just walk through the linguistic annotation of the word {\em girl} in the sentence:

\begin{figure}
\begin{center}
\begin{tabular}{l|ccccccc}
Words & \em the &  \em girl &  \em  watched &  \em attentively &  \em the & \em  beautiful &  \em fireflies\\ \hline
Part of speech & {\small \sc det} & {\small \sc nn} & {\small \sc adv} & {\small \sc vfin} & {\small \sc det} & {\small \sc jj} & {\small \sc nns} \\
Lemma & \em  the &  \em girl &  \em watch &  \em attentive &  \em the &  \em beautiful &  \em firefly \\
Morphology & - & {\small \sc sing.} & {\small \sc past} & - & - & {\small \sc plural}\\
Noun phrase & {\small \sc begin} & {\small \sc cont} & {\small \sc other} & {\small \sc other} & {\small \sc begin} & {\small \sc cont} & {\small \sc cont}  \\
Verb phrase & {\small \sc other} & {\small \sc other} & {\small \sc begin} & {\small \sc cont} &{\small \sc cont} &{\small \sc cont} &{\small \sc cont}\\
Synt. dependency &  \em girl &  \em watched & - &  \em watched &  \em fireflies &  \em fireflies &  \em watched \\
Depend. relation & {\small \sc det} & {\small \sc subj} & - & {\small \sc adv} & {\small \sc det} & {\small \sc adj} & {\small \sc obj}\\ 
Semantic role & - & {\small \sc actor} & - & {\small \sc manner} & - & {\small \sc mod} & {\small \sc patient}\\
Semantic type & - & {\small \sc human} & {\small \sc view} & - & - & - & {\small \sc animate}\\
\hline
\end{tabular}
\end{center}
\caption{Linguistic annotation of a sentence, formatted as word-level factored representation}
\label{fig:linguistic-input-annotation}
\end{figure}

\begin{itemize}\itemsep 0mm
\item Part of speech is {\sc nn}, a noun.
\item Lemma is {\em girl}, the same as the surface form. The lemma differs for {\em watched} / {\em watch}.
\item Morphology is singular.
\item The word is the continuation ({\sc cont}) of the noun phrase that started with {\em the}.
\item The word is not part of a verb phrase ({\sc other}).
\item Its syntactic head is {\em watched}.
\item The dependency relationship to the head is subject ({\sc subj}).
\item Its semantic role is {\sc actor}.
\item There are many schemes of semantic types. For instance {\em girl} could be classified as {\sc human}.
\end{itemize}

Note how phrasal annotations are handled. The first noun phrase is {\em the girl}. It is common to use an annotation scheme that tags individual words in a phrasal annation as {\sc begin} and {\sc continuation} (or {\sc intermediate}), while labelling words outside such phrases as {\sc other}. 

How do we encode the word-level factored representation? Recall that words are initially represented as 1-hot vectors. We can encode each factor in the factored representation as a 1-hot vector. The concatenation of these vectors is then used as input to the word embedding. Note that mathematically this means, that each factor of the representation is mapped to a embedding vector, and the final word embedding is the sum of the factor embeddings. 

Since the input to the neural machine translation system is still a sequence of word embeddings, we do not have to change anything in the architecture of the neural machine translation model. We just provide richer input representations and hope that the model is able to learn how to take advantage of it.

Coming back to the debate about linguistics versus machine learning. All the linguistic annotation proposed here can arguable be learned automatically as part of the word embeddings (or contextualized word embeddings in the hidden encoder states). This may or may not be true. But it does provide additional knowledge that comes from the tools that produce the annotation and that is particularly relevant if there is not enough training data to automatically induce it. Also, why make the job harder for the machine learning algorithm than needed? In other words, why force the machine learning to discover features that can be readily provided? Ultimately, these questions will be resolved empirically by demonstrating what actually works in specific data conditions.

\paragraph{Linguistic annotation of the output}
What we have done for input words could be done also for output words. Instead of discussing the fine points about what adjustments need to made (e.g., separate softmax for each output factor), let us take a look at another annotation scheme for the output that has been successfully applied to neural machine translation.

Most syntax-based statistical machine translation models have focused on adding syntax to the output side. Traditional n-gram language models are good at promoting fluency among neighboring words, they are not powerful enough to ensure overall grammaticality of each output sentence. By designing models that also produce and evaluate the syntactic parse structure for each output sentence, syntax-based models give the means to promote grammatically correct output.

The word-level annotation of phrase structure syntax suggested in Figure~\ref{fig:linguistic-input-annotation} is rather crude. The nature of language is recursive, and annotating nested phrases cannot be easily handled with a {\sc begin}/{\sc cont}/{\sc other} scheme. Instead, typically tree structures are used to represent syntax.

\begin{figure}
\begin{center}
\begin{tabular}{l|p{12cm}}
Sentence & \em the girl watched attentively the beautiful fireflies\\ \hline
Syntax tree & 
\tikzset{level distance=20pt}
\sc \Tree [.s [.np [.det {\em the} ] [.nn {\em girl} ] ] [.vp [.vfin {\em watched} ] [.advp [.adv {\em attentively} ] ] [.np [.det {\em the} ] [.jj {\em beautiful} ] [.nns {\em fireflies} ] ] ] ]
\\ \hline
Linearized & \em {\sc (s (np (det} the {\sc ) (nn} girl {\sc ) ) (vp (vfin } watched {\sc ) (advp (adv } attentively {\sc ) ) (np (det } the {\sc ) (jj} beautiful {\sc ) (nns} fireflies {\sc ) ) ) )}\\
\end{tabular}
\end{center}
\caption{Linearization of phrase structure grammar tree into a sequence of words --- e.g., {\em girl, watched} --- and tags --- e.g., {\sc (s, (np, )}}
\label{fig:syntax-annotation}
\end{figure}

See Figure~\ref{fig:syntax-annotation} for an example. It shows the phrase structure syntactic parse tree for our example sentence {\em The girl watched attentively the beautiful fireflies}. Generating a tree structures is generally a quite different process than generating a sequence. It is typically built recursively bottom-up with algorithms such as chart parsing. 

However, we can {\bf linearize}\index{linearization of parse tree} the parse tree into a sequence of words and structural tokens that indicate the beginning  --- e.g., {\sc ``(np"} --- and end --- closing parenthesis {\sc ``)"} ---  of syntactic phrases. So, forcing syntactic parse tree annotations into our sequence-to-sequence neural machine translation model may be done by encoding the parse structure with additional output tokens. To be perfectly clear, the idea is to produce as the output of the neural translation system not just a sequence of words, but a sequence of a mix of output words and special tokens. 

The hope is that forcing the neural machine translation model to produce syntactic structure (even in a linearized form) encourages it to produce syntactically well-formed output. There is some evidence to support this hope, despite the simplicity of the approach.

\paragraph{Linguistically structured models}
The field of syntactic parsing has not been left untouched by the recent wave of neural networks. The previous section suggests that syntactic parsing may be done as simply as framing it as a sequence to sequence with additional output tokens.

However, the best-performing syntactic parsers use model structures that take the recursive nature of language to heart. They are either inspired by convolutional networks and build parse trees bottom-up, or are neural versions of left-to-right push-down automata that maintain a stack of opened phrases that any new word may extend or close, or be pushed down the stack to start a new phrase.

There is some early work on integrating syntactic parsing and machine translation into a unified framework but no consensus on best practices has emerged yet. At the time of writing, this is clearly still a challenge for future work.

\paragraph{Further Readings}
\furtherreadings{\cite{iwslt12:Wu} propose to use factored representations of words (using lemma, stem, and part of speech), with each factor encoded in a one-hot vector, in the input to a recurrent neural network language model. \cite{sennrich-haddow:2016:WMT} use such representations in the input and output of neural machine translation models, demonstrating better translation quality.}

\subsection{Multiple Language Pairs}\index{multiple language pairs}
There are more than two languages in the world. And we also have training data for many language pairs, sometimes it is highly overlapping (e.g., European Parliament proceedings in 24 languages), sometimes it is unique (e.g. Canadian Hansards in French and English). For some language pairs, a lot of training data is available (e.g., French--English). But for most language pairs, there is only very little, including commercially interesting language pairs such as Chinese--German or Japanese--Spanish.

There is a long history of moving beyond specific languages and encode meaning language-independent, sometimes called interlingua. In machine translation, the idea is to map the input language first into an interlingua, and then map the interlingua into the output language. In such a system, we have to build just one mapping step into and one step out of the interlingua for each language. Then we can translate between it and all the other languages for which we have done the same.

Researchers in deep learning often do not hesitate to claim that intermediate states in neural translation models encode semantics or meaning. So, can we train a neural machine translation system that accepts text in any language as input and translates it into any other language?

\paragraph{Multiple Input Languages}
Let us say, we have two parallel corpora, one for German--English, and one for French--English. We can train a neural machine translation model on both corpora at the same time by simply concatenating them. The input vocabulary contains both German and French words. Any input sentence will be quickly recognized as being either German or French, due to the sentence context, disambiguating words such as {\em du} ({\em you} in German, {\em of} in French).

The combined model trained on both data sets has one advantage over two separate models. It is exposed to both English sides of the parallel corpora and hence can learn a better language model. There may be also be general benefits to having diversity in the data, leading to more robust models.

\paragraph{Multiple Output Languages}
We can do the same trick for the output language, by concatenating, say, a French--English and a French--Spanish corpus. But given a French input sentence during inference, how would the system know which output language to generate? A crude but effective way to signal this to the model is by adding a tag like {\sc [spanish]} as first token of the input sentence.

\begin{quotation} \em
\noindent
{\sc [english]} N'y a-t-il pas ici deux poids, deux mesures?\\
\phantom{.} \hfill  $\Rightarrow$
Is this not a case of double standards?\\[3mm]
{\sc [spanish]} N'y a-t-il pas ici deux poids, deux mesures?\\
\phantom{.} \hfill  $\Rightarrow$
¿No puede verse con toda claridad que estamos utilizando un doble rasero?
\end{quotation}

If we train a system on the three corpora mentioned (German--English, French--English, and French--Spanish) we can also use it translate a sentence from German to Spanish --- without having ever presented a sentence pair as training data to the system.

\begin{quotation} \em
\noindent
{\sc [spanish]} Messen wir hier nicht mit zweierlei Ma\ss?\\
\phantom{.} \hfill  $\Rightarrow$
¿No puede verse con toda claridad que estamos utilizando un doble rasero?
\end{quotation}

For this to work, there has to be some representation of the meaning of the input sentence that is not tied to the input language and the output language. Surprisingly, experiments show that this actually does work, somewhat. To achieve good quality, however, some parallel data in the desired language pair is needed, but much less than for a standalone model \citep{DBLP:journals/corr/JohnsonSLKWCTVW16}.

\begin{figure}
\begin{center}
\includegraphics[scale=0.55]{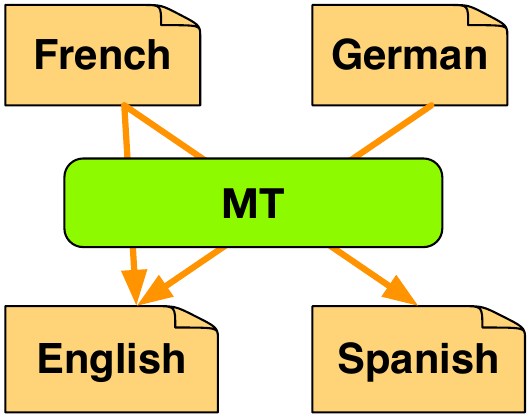}
\end{center}
\caption{Multi-language machine translation system trained on one language pair at a time, rotating through many of them. After training on French--English, French--Spanish, and German--English, it is even able to translate from German to Spanish.}
\label{fig:multiple-languages}
\end{figure}

Figure~\ref{fig:multiple-languages} summarizes this idea. A single neural machine translation is trained on various parallel corpora in turn, resulting in a system that may translate between any seen input and output language.
It is likely that increasingly deeper models (recall Section~\ref{sec:deep-models}) may better serve as multi-language translators, since their deeper layer compute more abstract representations of language.

The idea of marking the output language with a token such as  {\sc [spanish]} has been explored more widely in the context of systems for a single language pair. Such tokens may represent the domain of the input sentence \citep{Kobus:2016:unpublished}, or the required level of politeness of the output sentence \citep{sennrich-haddow-birch:2016:N16-1}.

\paragraph{Sharing Components}
Instead of just throwing data at a generic neural machine translation model, we may want to more carefully consider which components may be shared among language-pair-specific models. The idea is to train one model per language pair, but some of the components are identical in these unique models.
\begin{itemize}
\item The encoder may be shared in models that have the same input language.
\item The decoder may be shared in models that have the same output language.
\item The attention mechanism may be shared in all models for all language pairs.
\end{itemize}
Sharing components means is that the same parameter values (weight matrices, etc.) are used in these separate models. Updates to them when training a model for one language pair then also changes them for in the model for the other language pairs.
There is no need to mark the output language, since each model is trained for a specific language pair. 

The idea of shared training of components can also be pushed further to exploit monolingual data. The encoder may be trained on monolingual input language data, but we will need to add a training objective (e.g., language model cross-entropy). Also, the decoder may be trained in isolation with monolingual language model data. However, since there are no context states available, these have to be blanked out, which may lead it to learn to ignore the input sentence and function only as a target side language model.


\paragraph{Further Readings}
\furtherreadings{
\cite{DBLP:journals/corr/JohnsonSLKWCTVW16} explore how well a single canonical neural translation model is able to learn from multiple to multiple languages, by simultaneously training on on parallel corpora for several language pairs. They show small benefits for several input languages with the same output languages, mixed results for translating into multiple output languages (indicated by an additional input language token). The most interesting result is the ability for such a model to translate in language directions for which no parallel corpus is provided, thus demonstrating that some interlingual meaning representation is learned, although less well than using traditional pivot methods.

\cite{firat-cho-bengio:2016:N16-1} support multi-language input and output by training language-specific encoders and decoders and a shared attention mechanism.
}

\section{Alternate Architectures}
Most of neural network research has focused on the use of recurrent neural networks with attention. But this is by no means the only architecture for neural networks. Arguable, a disadvantage of using recurrent neural networks on the input side is that it requires a long sequential process that consumes each input word in one step. This also prohibits the ability to parallelize the processing of all words at once, thus limiting the use of the capabilities of GPUs.

There have been a few alternate suggestions for the architecture of neural machine translation models. We will briefly present some of them in this section. It remains to be seen, if they are a curiosity or conquer the field.

\subsection{Convolutional Neural Networks}
The first end-to-end neural machine translation model of the modern era \citep{kalchbrenner-blunsom:2013:EMNLP} was actually not based on recurrent neural networks, but based on {\bf convolutional neural networks}\index{convolutional neural network}\index{neural network!convolutional}. These had been shown to be very successful in image processing, thus looking for other applications was a natural next step.

\begin{figure}
\begin{center}
\includegraphics[scale=0.8]{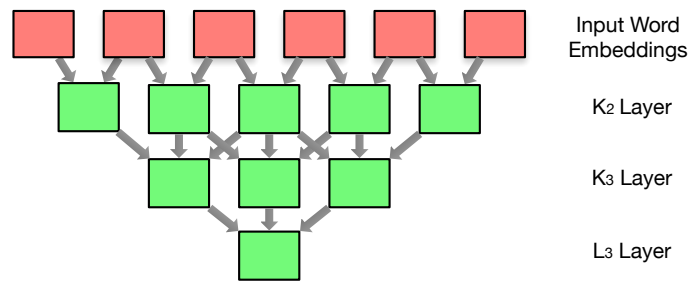}
\end{center}
\caption{Encoding a sentence with a convolutional neural network. By always using two convolutional layers, the size of the convolutions differ (here $K_2$ and $K_3$). Decoding reverses this process.}
\label{fig:convolutional-neural-network}
\end{figure}

See Figure~\ref{fig:convolutional-neural-network} for an illustration of a convolutional network that encodes an input sentence. The basic building block of these networks is a convolution. It merges the representation of $i$ input words into a single representation by using a matrix $K_i$. 
Applying the convolution to every sequence of input words reduces the length of the sentence representation by $i-1$. 
Repeating this process leads to a sentence representation in a single vector. 

The illustration shows an architecture with two convolutional $K_i$ layers, followed by a final $L_i$ layer that merges the sequence of phrasal representations into a single sentence representation. The size of the convolutional kernels $K_i$ and $L_i$ depends on the length of the sentences. The example shows a 6-word sentence and a sequence of $K_2$, $K_3$, and $L_3$ layers. For longer sentences, bigger kernels are needed.

The hierarchical process of building up a sentence representation bottom-up is well grounded in linguistic insight in the recursive nature of language. It is similar to chart parsing, except that we are not committing to a single hierarchical structure. On the other hand, we are asking an awful lot from the resulting sentence embedding to represents the meaning of an entire sentence of arbitrary length.

Generating the output sentence translation reverses the bottom-up process. One problem for the decoder is to decide the length of the output sentence. One option to address this problem is to add a model that predicts output length from input length. This then leads to the selection of the size of the reverse convolution matrices.

\begin{figure}
\begin{center}
\includegraphics[scale=0.8]{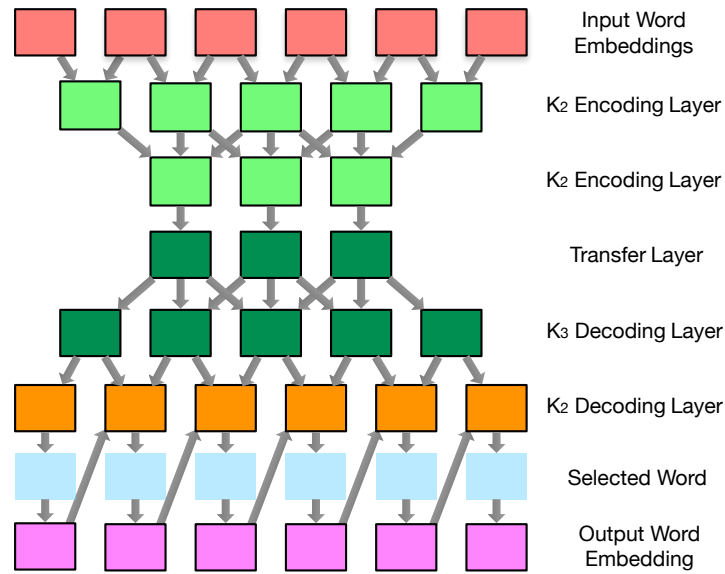}
\end{center}
\caption{Refinement of the convolutional neural network model. Convolutions do not result in a single sentence embedding but a sequence. The encoder is also informed by a recurrent neural network (connections from output word embeddings to final decoding layer.}
\label{fig:convolutional-neural-network-generation}
\end{figure}

See Figure~\ref{fig:convolutional-neural-network-generation} for an illustration of a variation of this idea. The shown architecture always uses a $K_2$ and a $K_3$ convolutional layer, resulting in a sequence of phrasal representations, not a single sentence embedding. There is an explicit mapping step from phrasal representations of input words to phrasal representations of output words, called transfer layer. 

The decoder of the model includes a recurrent neural network on the output side. Sneaking in a recurrent neural network here does undermine a bit the argument about better parallelization. However, the claim still holds true for encoding the input, and a sequential language model is just a too powerful tool to disregard.

While the just-described convolutional neural machine translation model helped to set the scene for neural network approaches for machine translation, it could not be demonstrated to achieve competitive results compared to traditional approaches. The compression of the sentence representation into a single vector is especially a problem for long sentences. However, the model was used successfully in reranking candidate translations generated by traditional statistical machine translation systems.

\subsection{Convolutional Neural Networks With Attention}
\cite{DBLP:journals/corr/GehringAGYD17} propose an architecture for neural networks that combines the ideas of convolutional neural networks and the attention mechanism. It is essentially the sequence-to-sequence attention that we described as the canonical neural machine translation approach, but with the recurrent neural networks replaced by convolutional layers.

We introduced convolutions in the previous section. The idea is to combine a short sequence of neighboring words into a single representation. To look at it in another way, a convolution encodes a word with its left and right context, in a limited window. Let us now describe in more detail what this means for the encoder and the decoder in the neural model.

\begin{figure}
\begin{center}
\includegraphics[scale=0.8]{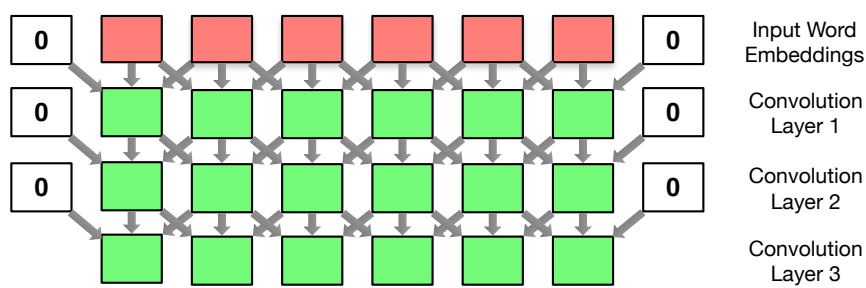}
\end{center}
\caption{Encoder using stacked convolutional layers. Any number of layers may be used.}
\label{fig:facebook-convolutional2017-encoder}
\end{figure}

\paragraph{Encoder} See Figure~\ref{fig:facebook-convolutional2017-encoder} for an illustration of the convolutional layers used in the encoder. For each input word, the state at each layer is informed by the corresponding state in the previous layer and its two neighbors. 
Note that these convolutional layers do not shorten the sequence, because we have a convolution centered around each word, using padding (vectors with zero values) for word positions that are out of bounds.

Mathematically, we start with the input word embeddings $E x_j$ and progress through a sequence of layer encodings $h_{d,j}$ at different depth $d$ until a maximum depth $D$.
\begin{equation}
\begin{aligned}
h_{0,j} &= E \; x_j\\
h_{d,j} &= f(h_{d-1,j-k}, ..., h_{d-1,j+k}) & \text{for $d>0$, $d \leq D$}
\end{aligned}
\end{equation}

The function $f$ is a feed-forward layer, with a residual connection from the corresponding previous layer state $h_{d-1,j}$.

Note that even with a few convolutional layers, the final representation of a word $h_{D,j}$ may only be informed by partial sentence context --- in contrast to the bi-directional recurrent neural networks in the canonical model. However, relevant context words in the input sentence that help with disambiguation may be outside this window. 

On the other hand, there are significant computational advantages to this idea. All words at one depth can be processed in parallel, even combined into one massive tensor operation that can be efficiently parallelized on a GPU.

\paragraph{Decoder} The decoder in the canonical model also has at its core a recurrent neural network. Recall its state progression defined in Equation~\vref{eq:nn:neural-translation-final}:
\begin{equation}
s_i = f(s_{i-1},\; Ey_{i-1},c_i)
\end{equation}
where $s_i$ is the encoder state, $Ey_{i-1}$ the embedding of the previous output word, and $c_i$ the input context.

The convolutional version of this does not have recurrent decoder states, i.e., the computation does not depend on the previous state $s_{i-1}$, but is conditioned on the sequence of the $\kappa$ most recent previous words.
\begin{equation}
s_i = f(Ey_{i-\kappa},..., Ey_{i-1},c_i)
\end{equation}

Furthermore, these decoder convolutions may be stacked, just as the encoder convolutional layers.
\begin{equation}
\begin{aligned}
s_{1,i} &= f(Ey_{i-\kappa},..., Ey_{i-1},c_i)\\
s_{d,i} &= f(s_{d-1,i-\kappa-1},..., s_{d-1,i},c_i) & \text{for $d>0$, $d \leq \hat{D}$}
\end{aligned}
\end{equation}

\begin{figure}
\begin{center}
\includegraphics[scale=0.8]{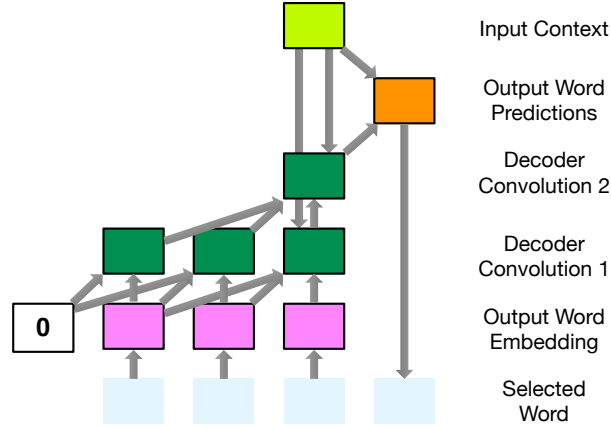}
\end{center}
\caption{Decoder in convolutional neural network with attention. The decoder state is computed as a sequence of convolutional layers (here: 2) over the already predicted output words. Each convolutional state is also informed by the input context computed from the input sentence and attention.}
\label{fig:facebook-convolutional2017-decoder}
\end{figure}

See Figure~\ref{fig:facebook-convolutional2017-decoder} for an illustration of these equations. The main difference between the canonical neural machine translation model and this architecture is the conditioning of the states of the decoder. They are computed in a sequence of convolutional layers, and also always the input context.

\paragraph{Attention} The attention mechanism is essentially unchanged from the canonical neural translation model. Recall that is is based on an association $a(s_{i-1},h_j)$ between the word representations computed by the encoder $h_j$ and the previous state of the decoder $s_{i-1}$ (refer back to Equation~\vref{eqn:attention-model}).

Since we still have such encoder and decoder states ($h_{D,j}$ and $s_{\hat{D},{i-1}}$), we use the same here. These association scores are normalized and used to compute a weighted sum of the input word embeddings (i.e., the encoder states $h_{D,j}$). A refinement is that the encoder state $h_{D,j}$ and the input word embedding $x_j$ is combined via addition when computing the context vector. This is the usual trick of using residual connections to assist training with deep neural networks. 

\subsection{Self-Attention}
The critique of the use of recurrent neural networks is that they require a lengthy walk-through, word by word, of the entire input sentence, which is time-consuming and limits parallelization. The previous sections replaced the recurrent neural networks in our canonical model with convolutions. However, these have a limited context window to enrich representations of words. What we would like is some architectural component that allows us to use wide context and can be highly parallelized. What could that be?

In fact, we already encountered it: the attention mechanism. It considers associations between every input word and any output word, and uses it to build a vector representation of the entire input sequence. The idea behind {\bf self-attention}\index{self-attention}\index{attention!self-attention} is to extend this idea to the encoder. Instead of computing the association between an input and an output word, self-attention computes the association between any input word and any other input word. One way to view it is that this mechanism refines the representation of each input word by enriching it with context words that help to disambiguate it.

\paragraph{Computing Self-Attention}
\cite{DBLP:journals/corr/VaswaniSPUJGKP17} define self attention for a sequence of vectors $h_j$ (of size $|h|$), packed into a matrix $H$, as
\begin{equation}
\text{self-attention}(H) = \text{softmax} \Big( \frac{HH^T}{\sqrt{|h|}} \Big) H
\label{eqn:self-attention}
\end{equation}

Let us look at this equation in detail. The association between every word representation $h_j$ any other context word $h_k$ is done via the dot product between the packed matrix $H$ and its transpose $H^T$, resulting in a vector of {\em raw association} values $HH^T$. The values in this vector are first scaled by the size of the word representation vectors $|h|$, and then by the softmax, so that their values add up to 1. The resulting vector of {\em normalized association} values is then used to weigh the context words.

Another way to put Equation~\ref{eqn:self-attention} without the matrix $H$ notation but using word representation vectors $h_j$:
\begin{equation}
\begin{aligned}
a_{jk} &= \frac{1}{|h|} h_jh_k^T & \text{raw association $\Big(\frac{HH^T}{\sqrt{|h|}}\Big)$}\\
\alpha_{jk} & = \frac{\text{exp}(a_{jk})}{\sum_\kappa \text{exp}(a_{j\kappa})} & \text{normalized association (softmax)}\\
\text{self-attention}(h_j) &= \sum_k \alpha_{j\kappa} h_k & \text{weighted sum}
\end{aligned}
\end{equation}

\paragraph{Self-Attention Layer}
The self-attention step described above is only one step in the self-attention layer used to encode the input sentence. There are four more steps that follow it.
\begin{itemize}
\item We combine self-attention with residual connections that pass the word representation through directly
\begin{equation}
\text{self-attention}(h_j) + h_j
\end{equation}
\item Next up is a layer normalization step (described in Section~\vref{sec:layer-normalization}).
\begin{equation}
\hat{h}_j = \text{layer-normalization}(\text{self-attention}(h_j) + h_j)
\end{equation}
\item A standard feed-forward step with ReLU activation function is applied.
\begin{equation}
\text{relu}(W \hat{h}_j + b)
\end{equation}
\item This is also augmented with residual connections and layer normalization.
\begin{equation}
\text{layer-normalization}(\text{relu}(W \hat{h}_j + b) + \hat{h}_j)
\end{equation}
\end{itemize}

Taking a page from deep models, we now stack several such layers (say, $D=6$) on top of each other.

\begin{equation}
\begin{aligned}
h_{0,j} & = Ex_j & \text{start with input word embedding}\\
h_{d,j} &= \text{self-attention-layer}(h_{d-1,j}) & \text{for $d>0$, $d \leq D$}
\end{aligned}
\end{equation}

The deep modeling is the reason behind the residual connections in the self-attention layer --- such residual connections help with training since they allow a shortcut to the input which may be utilized in early stages of training, before it can take advantage of the more complex interdependencies that deep models enable. The layer normalization step is one standard training trick that also helps especially with deep models.

\paragraph{Attention in the Decoder}
Self-attention is also used in the decoder, now between output words. The decoder also has more traditional attention. In total there are 3 sub layers.

\begin{itemize}
\item {\em Self attention:} Output words are initially encoded by word embeddings $s_i = Ey_i$. We perform exactly the same self-attention computation as described in Equation~\ref{eqn:self-attention}. However, the association of a word $s_i$ is limited to words $s_k$ with $k \leq i$, i.e., just the previously produced output words. Let us denote the result of this sub layer for output word $i$ as $\tilde{s_i}$
\item {\em Attention:} The attention mechanism in this model follows very closely self-attention. The only difference is that, previously, we compute self attention between the hidden states $H$ and themselves. Now, we compute attention between the decoder states $\tilde{S}$ and the final encoder states $H$.
\begin{equation}
\text{attention}(\tilde{S},H) = \text{softmax} \Big( \frac{\tilde{S}H^T}{\sqrt{|h|}} \Big) H
\end{equation}

Using the same more detailed exposition as above for self-attention:

\begin{equation}
\begin{aligned}
a_{ik} &= \frac{1}{|h|} \tilde{s}_ih_k^T & \text{raw association $\Big(\frac{\tilde{S}H^T}{\sqrt{|h|}}\Big)$}\\
\alpha_{ik} & = \frac{\text{exp}(a_{ik})}{\sum_\kappa \text{exp}(a_{i\kappa})} & \text{normalized association (softmax)}\\
\text{attention}(\tilde{s}_i) &= \sum_k \alpha_{j\kappa} h_k & \text{weighted sum}
\end{aligned}
\end{equation}

This attention computation is augmented by adding in residual connections, layer normalization, and an additional ReLU layer, just like the self-attention layer described above.

It is worth noting that, the output of the attention computation is a weighted sum over input word representations $\sum_k \alpha_{j\kappa} h_k$. To this, we add  the (self-attended) representation of the decoder state $\tilde{s}_i$ via a residual connection. This allows skipping over the deep layers, thus speeding up training.

\item {\em Feed-forward layer:} This sub layer is identical to the encoder, i.e., $\text{relu}(W_s \hat{s}_i + b_s)$
\end{itemize}

Each of the sub-layers is followed by the add-and-norm step of first using residual connections and then layer normalization (as noted in the description of the attention sub layer).

\begin{figure}
\begin{center}
\includegraphics[scale=0.8]{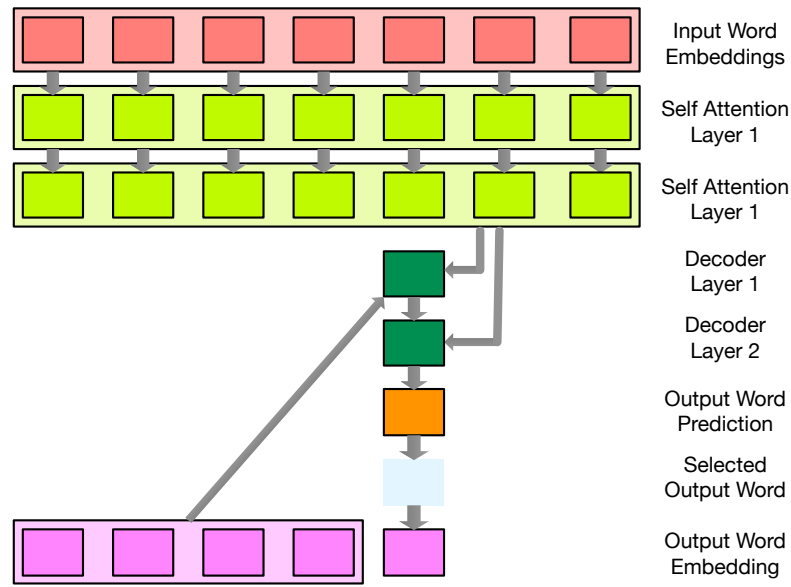}
\end{center}
\caption{Attention-based machine translation model: the input is encoded with several layers of self-attention. The decoder computes attention-based representations of the input in several layers, initialized with the previous word embeddings.}
\label{fig:self-attention}
\end{figure}

The entire model is shown in Figure~\ref{fig:self-attention}, 

\paragraph{Further Readings} 
\furtherreadings{\cite{kalchbrenner-blunsom:2013:EMNLP} build a comprehensive machine translation model by first encoding the source sentence with a convolutional neural network, and then generate the target sentence by reversing the process. A refinement of this was proposed by \cite{DBLP:journals/corr/GehringAGYD17} who use multiple convolutional layers in the encoder and the decoder that do not reduce the length of the encoded sequence but incorporate wider context with each layer.

\cite{DBLP:journals/corr/VaswaniSPUJGKP17} replace the recurrent neural networks used in attentional sequence-to-sequence models with multiple self-attention layers, both for the encoder as well as the decoder. There are a number of additional refinements of this model: so-called multi-head attention, encoding of sentence positions of words, etc.}



\section{Current Challenges}

Neural machine translation has emerged as the most promising machine translation approach in recent years, showing superior performance on public benchmarks \citep{bojar-EtAl:2016:WMT1} and rapid adoption in deployments by, e.g., Google \citep{DBLP:journals/corr/WuSCLNMKCGMKSJL16}, Systran \citep{DBLP:journals/corr/CregoKKRYSABCDE16}, and WIPO \citep{IWSLT-2016-Junczys-Dowmunt}. But there have also been reports of poor performance, such as the systems built under low-resource conditions in the DARPA LORELEI program.\footnote{\tt https://www.nist.gov/itl/iad/mig/lorehlt16- evaluations}

Here, we examine a number of challenges to neural machine translation and give empirical results on how well the technology currently holds up, compared to traditional statistical machine translation.
We show that, despite its recent successes, neural machine translation still has to overcome various challenges, most notably performance out-of-domain and under low resource conditions. 

What a lot of the problems have in common is that the neural translation models do not show robust behavior when confronted with conditions that differ significantly from training conditions --- may it be due to limited exposure to training data, unusual input in case of out-of-domain test sentences, or unlikely initial word choices in beam search. The solution to these problems may hence lie in a more general approach of training that steps outside optimizing single word predictions given perfectly matching prior sequences.

Another challenge that we do not examine empirically: neural machine translation systems are much less interpretable. The answer to the question of why the training data leads these systems to decide on specific word choices during decoding is buried in large matrices of real-numbered values. There is a clear need to develop better analytics for neural machine translation.

We use common toolkits for neural machine translation (Nematus) and traditional phrase-based statistical machine translation (Moses) with common data sets, drawn from WMT and OPUS.
Unless noted otherwise, we use default settings, such as beam search and single model decoding. The training data is processed with byte-pair encoding \citep{sennrich-haddow-birch:2016:P16-11} into subwords to fit a 50,000 word vocabulary limit.

Our statistical machine translation systems are trained using Moses\footnote{\tt http://www.stat.org/moses/} \citep{koehn-EtAl:2007:PosterDemo}. We build phrase-based systems using standard features that are commonly used in recent system submissions to WMT \citep{williams-EtAl:2016:WMT,ding-EtAl:2016:WMT}. 
While we consider here only phrase-based systems, we note that there are other statistical machine translation approaches such as hierarchical phrase-based models \citep{Chiang:CL:2007} and syntax-based models \citep{Galley:2004,galley-EtAl:2006:COLACL} that have been shown to give superior performance for language pairs such as Chinese--English and German--English.

We carry out our experiments on English--Spanish and German--English. For these language pairs, large training data sets are available.  We use datasets from the shared translation task organized alongside the Conference on Machine Translation (WMT)\footnote{\tt http://www.statmt.org/wmt17/}. For the domain experiments, we use the OPUS corpus\footnote{\tt http://opus.lingfil.uu.se/} \citep{TIEDEMANN12.463.L12-1246}.

Except for the domain experiments, we use the WMT test sets composed of news stories, which are characterized by a broad range of topic, formal language, relatively long sentences (about 30 words on average), and high standards for grammar, orthography, and style.

\subsection{Domain Mismatch}
A known challenge in translation is that in different domains,\footnote{We use the customary definition of domain in machine translation: a {\em domain} is defined by a corpus from a specific source, and may differ from other {\em domains} in topic, genre, style, level of formality, etc.} words have different translations and meaning is expressed in different styles. Hence, a crucial step in developing machine translation systems targeted at a specific use case is domain adaptation. We expect that methods for domain adaptation will be developed for neural machine translation. A currently popular approach is to train a general domain system, followed by training on in-domain data for a few epochs \citep{IWSLT-2015-Luong,Freitag:2016:unpublished}.

Often, large amounts of training data are only available out of domain, but we still seek to have robust performance. To test how well neural machine translation and statistical machine translation hold up, we trained five different systems using different corpora obtained from OPUS \citep{TIEDEMANN12.463.L12-1246}. An additional system was trained on all the training data. Statistics about corpus sizes are shown in Table~\ref{tab:domain-corpora}. Note that these domains are quite distant from each other, much more so than, say, Europarl, TED Talks, News Commentary, and Global Voices.

\begin{table}
\small
\begin{center}
\begin{tabular}{lrrc}
\bf Corpus & \bf Words & \bf Sentences & \bf W/S \\ \hline
Law (Acquis) &   18,128,173 &  715,372 & 25.3 \\
Medical (EMEA) &  14,301,472 & 1,104,752 & 12.9 \\
IT &   3,041,677 &  337,817 & 9.0 \\
Koran (Tanzil) &   9,848,539 & 480,421 & 20.5 \\
Subtitles & 114,371,754 & 13,873,398 & 8.2 \\
\end{tabular}
\end{center}
\caption{Corpora used to train domain-specific systems, taken from the OPUS repository. IT corpora are GNOME, KDE, PHP, Ubuntu, and OpenOffice.}
\label{tab:domain-corpora}
\end{table}

We trained both statistical machine translation and neural machine translation systems for all domains.
All systems were trained for German-English, with tuning and test sets sub-sampled from the data (these were not used in training). A common byte-pair encoding is used for all training runs.

\begin{figure}
\begin{center}

\newcommand{\barchart}[2]{\begin{tikzpicture}[scale=0.02]
\fill[green] (0,0) rectangle (55,#1);
\fill[blue] (60,0) rectangle (115,#2);
\draw (27.5,0) node[anchor=north] {#1};
\draw (87.5,0) node[anchor=north] {#2};
\end{tikzpicture}}
\newcommand{\rowlabel}[1]{\begin{tikzpicture}[scale=0.02]\fill[white] (0,0) rectangle (2,2);\draw (1,0) node[anchor=north] {\bf #1}; \end{tikzpicture}}

\begin{tabular}{l|c|c|c|c|c}
\bf System $\downarrow$ & \bf Law & \bf Medical& \bf IT& \bf Koran& \bf Subtitles \\ \hline \hline
\rowlabel{All Data} & \barchart{30.5}{32.8} & \barchart{45.1}{42.2} & \barchart{35.3}{44.7} & \barchart{17.9}{17.9} & \barchart{26.4}{20.8}\\\hline \hline
\rowlabel{Law} & \cellcolor{lightgray}\barchart{31.1}{34.4} & \barchart{12.1}{18.2} & \barchart{3.5}{6.9} & \barchart{1.3}{2.2} & \barchart{2.8}{6.0}\\\hline
\rowlabel{Medical} & \barchart{3.9}{10.2} & \cellcolor{lightgray}\barchart{39.4}{43.5} & \barchart{2.0}{8.5} & \barchart{0.6}{2.0} & \barchart{1.4}{5.8}\\\hline
\rowlabel{IT} & \barchart{1.9}{3.7} & \barchart{6.5}{5.3} & \cellcolor{lightgray}\barchart{42.1}{39.8} & \barchart{1.8}{1.6} & \barchart{3.9}{4.7}\\\hline
\rowlabel{Koran} & \barchart{0.4}{1.8} & \barchart{0.0}{2.1} & \barchart{0.0}{2.3} & \cellcolor{lightgray}\barchart{15.9}{18.8} & \barchart{1.0}{5.5}\\\hline
\rowlabel{Subtitles} & \barchart{7.0}{9.9} & \barchart{9.3}{17.8} & \barchart{9.2}{13.6} & \barchart{9.0}{8.4} & \cellcolor{lightgray}\barchart{25.9}{22.1}\\\hline
\end{tabular}

\end{center}
\caption{Quality of systems (BLEU), when trained on one domain (rows) and tested on another domain (columns). Comparably, neural machine translation systems (left bars) show more degraded performance out of domain.}
\label{tab:domain-results}
\end{figure}

See Figure~\ref{tab:domain-results} for results. 
While the in-domain neural and statistical machine translation systems are similar (neural machine translation is better for IT and Subtitles, statistical machine translation is better for Law, Medical, and Koran), the out-of-domain performance for the neural machine translation systems is worse in almost all cases, sometimes dramatically so. For instance the Medical system leads to a BLEU score of 3.9 (neural machine translation) vs. 10.2 (statistical machine translation) on the Law test set. 

Figure~\ref{fig:domain-mismatch-examples} displays an example. When translating the sentence {\em Schaue um dich herum.} (reference: {\em Look around you.}) from the Subtitles corpus, we see mostly non-sensical and completely unrelated output from the neural machine translation system. For instance, the translation from the IT system is {\em Switches to paused.} 

Note that the output of the neural machine translation system is often quite fluent (e.g., {\em Take heed of your own souls.}) but completely unrelated to the input, while the statistical machine translation output betrays its difficulties with coping with the out-of-domain input by leaving some words untranslated (e.g., {\em Schaue by dich around.}). This is of particular concern when MT is used for information gisting --- the user will be mislead by hallucinated content in the neural machine translation output.

\begin{figure}
\small
\begin{center}
\begin{tabular}{l|p{10cm}}
\bf Source & \em Schaue um dich herum.\\ \hline
\bf Reference & \em Look around you. \\ \hline \hline
\bf All & NMT: \em Look around you.\\
& SMT: \em Look around you.\\\hline
\bf Law & NMT: \em Sughum gravecorn.\\
& SMT: \em In order to implement dich Schaue .\\\hline
\bf Medical & NMT: \em EMEA / MB / 049 / 01-EN-Final Work progamme for 2002\\
& SMT: \em Schaue by dich around .\\\hline
\bf IT & NMT: \em Switches to paused.\\
& SMT: \em To Schaue by itself . \textbackslash t \textbackslash t\\\hline
\bf Koran & NMT: \em Take heed of your own souls.\\
& SMT: \em And you see. \\\hline
\bf Subtitles & NMT: \em Look around you.\\
& SMT: \em Look around you .
\end{tabular}
\end{center}
\caption{Examples for the translation of a sentence from the Subtitles corpus, when translated with systems trained on different corpora. Performance out-of-domain is dramatically worse for neural machine translation.}
\label{fig:domain-mismatch-examples}
\end{figure}

\subsection{Amount of Training Data}\label{sec:learning-curve}\index{amount of training data}\index{size of training data}\index{data!size}
A well-known property of statistical systems is that increasing amounts of training data lead to better results. In statistical machine translation systems, we have previously observed that doubling the amount of training data gives a fixed increase in BLEU scores. This holds true for both parallel and monolingual data \citep{turchi-debie-cristianini:2008:WMT,irvine-callisonburch:2013:WMT}.

\begin{figure}
\begin{center}
\small\noindent\begin{tikzpicture}
\input{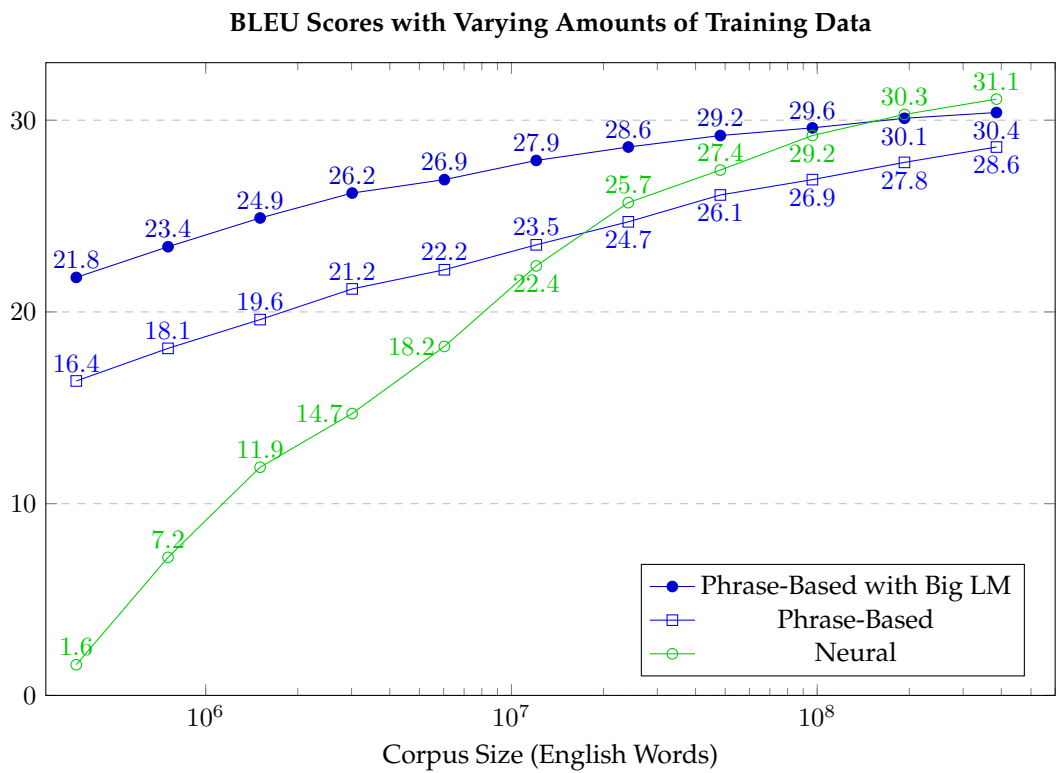}
\end{tikzpicture}
\end{center}
\caption{BLEU scores for English-Spanish systems trained on 0.4 million to 385.7 million words of parallel data. Quality for neural machine translation starts much lower, outperforms statistical machine translation at about 15 million words, and even beats a statistical machine translation system with a big 2 billion word in-domain language model under high-resource conditions.}
\label{fig:learning-curve}
\end{figure}

How do the data needs of statistical machine translation and neural machine translation compare? Neural machine translation promises both to generalize better (exploiting word similarity in embeddings) and condition on larger context (entire input and all prior output words).

We built English-Spanish systems on WMT data,\footnote{Spanish was last represented in 2013, we used data from \tt http://statmt.org/wmt13/translation-task.html} about 385.7 million English words paired with Spanish. To obtain a learning curve, we used $\frac{1}{1024}$, $\frac{1}{512}$, ..., $\frac{1}{2}$, and all of the data. For statistical machine translation, the language model was trained on the Spanish part of each subset, respectively. In addition to a neural and statistical machine translation system trained on each subset, we also used all additionally provided monolingual data for a big language model in contrastive statistical machine translation systems.

Results are shown in Figure~\ref{fig:learning-curve}. Neural machine translation exhibits a much steeper learning curve, starting with abysmal results (BLEU score of 1.6 vs. 16.4 for $\frac{1}{1024}$ of the data), outperforming statistical machine translation 25.7 vs. 24.7 with $\frac{1}{16}$ of the data (24.1 million words), and even beating the statistical machine translation system with a big language model with the full data set (31.1 for neural machine translation, 28.4 for statistical machine translation, 30.4 for statistical with a big language model).

The contrast between the neural and statistical machine translation learning curves is quite striking. While neural machine translation is able to exploit increasing amounts of training data more effectively, it is unable to get off the ground with training corpus sizes of a few million words or less. 

To illustrate this, see Figure~\ref{fig:learning-curve-example}. With $\frac{1}{1024}$ of the training data, the output is completely unrelated to the input, some key words are properly translated with $\frac{1}{512}$ and $\frac{1}{256}$ of the data ({\em estrategia} for {\em strategy}, {\em elecci{\'o}n} or {\em elecciones} for {\em election}), and starting with $\frac{1}{64}$ the translations become respectable.

\begin{figure}
\begin{center}
\small
\begin{tabular}{c|r|p{12cm}}
\bf Ratio & \bf Words & {\bf Source:} \em A Republican strategy to counter the re-election of Obama\\\hline\hline
$\frac{1}{1024}$ & 0.4 million & \em Un {\'o}rgano de coordinaci{\'o}n para el anuncio de libre determinaci{\'o}n\\ \hline
$\frac{1}{512}$ & 0.8 million & \em Lista de una estrategia para luchar contra la elecci{\'o}n de hojas de Ohio\\\hline
$\frac{1}{256}$ &1.5 million  &\em Explosi{\'o}n realiza una estrategia divisiva de luchar contra las elecciones de autor\\\hline
$\frac{1}{128}$ & 3.0 million & \em Una estrategia republicana para la eliminaci{\'o}n de la reelecci{\'o}n de Obama\\\hline
$\frac{1}{64}$ & 6.0 million & \em Estrategia siria para contrarrestar la reelecci{\'o}n del Obama .\\\hline
$\frac{1}{32}+$ & 12.0 million &  \em Una estrategia republicana para contrarrestar la reelecci{\'o}n de Obama\\
\end{tabular}
\end{center}
\caption{Translations of the first sentence of the test set using neural machine translation system trained on varying amounts of training data. Under low resource conditions, neural machine translation produces fluent output unrelated to the input.}
\label{fig:learning-curve-example}
\end{figure}

\subsection{Noisy Data}
Statistical machine translation is fairly robust to {\bf noisy data}\index{data!noisy}\index{noisy data}. The quality of systems holds up fairly well, even if large parts of the training data are corrupted in various ways, such as misaligned sentences, content in wrong languages, badly translated sentences, etc. Statistical machine translation models are built on probability distributions estimated from many occurrences of words and phrases. Any unsystematic noise in the training only affects the tail end of the distribution.

Is this still the case for neural machine translation? \cite{chen:AMTA:2016} considered one kind of noise: misaligned sentence pairs in an experiments with a large English--French parallel corpus. They shuffle the target side of part of the training corpus, so that these sentence pairs are mis-aligned.

Table~\ref{tab:noise} shows the result. Statistical machine translation systems hold up fairly well. Even with 50\% of the data perturbed, the quality only drops from 32.7 to 32.0 {\sc bleu} points, about what is to be expected with half the valid training data. However, the neural machine translation system degrades severely, from 35.4 to 30.1 {\sc bleu} points, a drop of 5.3 points, compared to the 0.7 point drop for statistical systems.

\begin{table}
\begin{center}
\begin{tabular}{l|c|c|c|c} 
Ratio shuffled & 0\% & 10\% & 20\% & 50\% \\ \hline
SMT \sc (bleu) & 32.7 & 32.7 (--0.0) & 32.6  (--0.1) & 32.0 (--0.7) \\ 
NMT \sc (bleu) & 35.4 & 34.8 (--0.6) & 32.1  (--3.3) & 30.1 (--5.3) \\
\end{tabular}
\end{center}
\caption{Impact of noise in the training data, with parts of the training corpus shuffled to contain mis-aligned sentence pairs. Neural machine translation degrades severely, while statistical machine translation holds up fairly well.}
\label{tab:noise}
\end{table}

A possible explanation for this poor behavior of neural machine translation models is that its prediction has to find a good balance between language model and input context as the main driver. When training observes increasing ratios of training example, for which the input sentence is a meaningless distraction, it may generally learn to rely more on the output language model aspect, hence hallucinating fluent by inadequate output.


\subsection{Word Alignment}\index{word alignment}\index{alignment!evaluation}
The key contribution of the attention model in neural machine translation \citep{bahdanau:ICLR:2015} was the imposition of an alignment of the output words to the input words. This takes the shape of a probability distribution over the input words which is used to weigh them in a bag-of-words representation of the input sentence.

Arguably, this attention model does not functionally play the role of a word alignment between the source in  the target, at least not in the same way as its analog in statistical machine translation. While in both cases, alignment is a latent variable that is used to obtain probability distributions over words or phrases, arguably the attention model has a broader role. For instance, when translating a verb, attention may also be paid to its subject and object since these may disambiguate it. To further complicate matters, the word representations are products of bidirectional gated recurrent neural networks that have the effect that each word representation is informed by the entire sentence context.

But there is a clear need for an alignment mechanism between source and target words. For instance, prior work used the alignments provided by the attention model to interpolate word translation decisions with traditional probabilistic dictionaries \citep{arthur-neubig-nakamura:2016:EMNLP2016}, for the introduction of coverage and fertility models \citep{tu-EtAl:2016:P16-1}, etc.

But is the attention model in fact the proper means? To examine this, we compare the soft alignment matrix (the sequence of attention vectors) with word alignments obtained by traditional word alignment methods. We use incremental fast-align \citep{dyer-chahuneau-smith:2013:NAACL-HLT} to align the input and output of the neural machine system. 

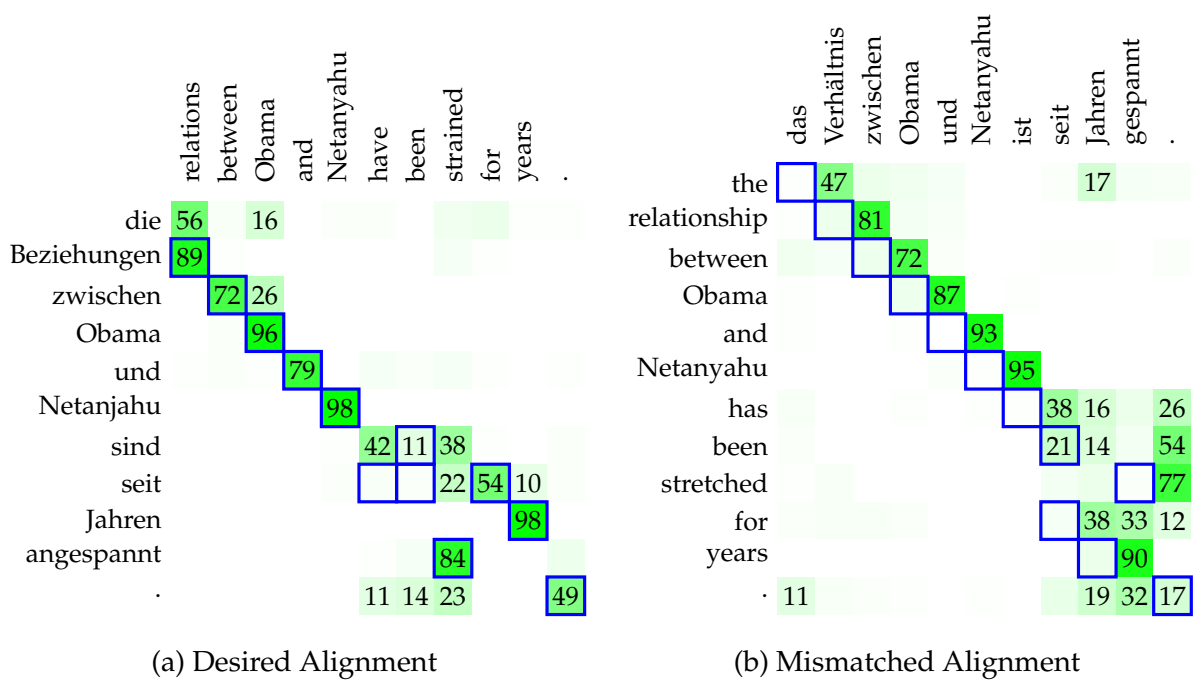
\begin{figure}
\begin{tabular}{cc}
\small
\begin{tikzpicture}[scale=0.5]
\node[label=above:\rotatebox{90}{relations}] at (0.5,11) {};
\node[label=above:\rotatebox{90}{between}] at (1.5,11) {};
\node[label=above:\rotatebox{90}{Obama}] at (2.5,11) {};
\node[label=above:\rotatebox{90}{and}] at (3.5,11) {};
\node[label=above:\rotatebox{90}{Netanyahu}] at (4.5,11) {};
\node[label=above:\rotatebox{90}{have}] at (5.5,11) {};
\node[label=above:\rotatebox{90}{been}] at (6.5,11) {};
\node[label=above:\rotatebox{90}{strained}] at (7.5,11) {};
\node[label=above:\rotatebox{90}{for}] at (8.5,11) {};
\node[label=above:\rotatebox{90}{years}] at (9.5,11) {};
\node[label=above:\rotatebox{90}{.}] at (10.5,11) {};
\draw (0,10.5) node[anchor=east] {die};
\draw (0,9.5) node[anchor=east] {Beziehungen};
\draw (0,8.5) node[anchor=east] {zwischen};
\draw (0,7.5) node[anchor=east] {Obama};
\draw (0,6.5) node[anchor=east] {und};
\draw (0,5.5) node[anchor=east] {Netanjahu};
\draw (0,4.5) node[anchor=east] {sind};
\draw (0,3.5) node[anchor=east] {seit};
\draw (0,2.5) node[anchor=east] {Jahren};
\draw (0,1.5) node[anchor=east] {angespannt};
\draw (0,0.5) node[anchor=east] {.};
\fill[green!56!white] (0,10) rectangle (1,11);
\draw (0.5,10.5) node[align=center] {56};
\fill[green!89!white] (0,9) rectangle (1,10);
\draw (0.5,9.5) node[align=center] {89};
\fill[green!0!white] (0,8) rectangle (1,9);
\fill[green!0!white] (0,7) rectangle (1,8);
\fill[green!1!white] (0,6) rectangle (1,7);
\fill[green!0!white] (0,5) rectangle (1,6);
\fill[green!0!white] (0,4) rectangle (1,5);
\fill[green!0!white] (0,3) rectangle (1,4);
\fill[green!0!white] (0,2) rectangle (1,3);
\fill[green!0!white] (0,1) rectangle (1,2);
\fill[green!0!white] (0,0) rectangle (1,1);
\fill[green!3!white] (1,10) rectangle (2,11);
\fill[green!1!white] (1,9) rectangle (2,10);
\fill[green!72!white] (1,8) rectangle (2,9);
\draw (1.5,8.5) node[align=center] {72};
\fill[green!2!white] (1,7) rectangle (2,8);
\fill[green!2!white] (1,6) rectangle (2,7);
\fill[green!0!white] (1,5) rectangle (2,6);
\fill[green!0!white] (1,4) rectangle (2,5);
\fill[green!0!white] (1,3) rectangle (2,4);
\fill[green!0!white] (1,2) rectangle (2,3);
\fill[green!0!white] (1,1) rectangle (2,2);
\fill[green!0!white] (1,0) rectangle (2,1);
\fill[green!16!white] (2,10) rectangle (3,11);
\draw (2.5,10.5) node[align=center] {16};
\fill[green!0!white] (2,9) rectangle (3,10);
\fill[green!26!white] (2,8) rectangle (3,9);
\draw (2.5,8.5) node[align=center] {26};
\fill[green!96!white] (2,7) rectangle (3,8);
\draw (2.5,7.5) node[align=center] {96};
\fill[green!1!white] (2,6) rectangle (3,7);
\fill[green!0!white] (2,5) rectangle (3,6);
\fill[green!0!white] (2,4) rectangle (3,5);
\fill[green!0!white] (2,3) rectangle (3,4);
\fill[green!0!white] (2,2) rectangle (3,3);
\fill[green!0!white] (2,1) rectangle (3,2);
\fill[green!0!white] (2,0) rectangle (3,1);
\fill[green!0!white] (3,10) rectangle (4,11);
\fill[green!0!white] (3,9) rectangle (4,10);
\fill[green!0!white] (3,8) rectangle (4,9);
\fill[green!0!white] (3,7) rectangle (4,8);
\fill[green!79!white] (3,6) rectangle (4,7);
\draw (3.5,6.5) node[align=center] {79};
\fill[green!0!white] (3,5) rectangle (4,6);
\fill[green!0!white] (3,4) rectangle (4,5);
\fill[green!0!white] (3,3) rectangle (4,4);
\fill[green!0!white] (3,2) rectangle (4,3);
\fill[green!0!white] (3,1) rectangle (4,2);
\fill[green!0!white] (3,0) rectangle (4,1);
\fill[green!2!white] (4,10) rectangle (5,11);
\fill[green!0!white] (4,9) rectangle (5,10);
\fill[green!0!white] (4,8) rectangle (5,9);
\fill[green!0!white] (4,7) rectangle (5,8);
\fill[green!0!white] (4,6) rectangle (5,7);
\fill[green!98!white] (4,5) rectangle (5,6);
\draw (4.5,5.5) node[align=center] {98};
\fill[green!1!white] (4,4) rectangle (5,5);
\fill[green!2!white] (4,3) rectangle (5,4);
\fill[green!0!white] (4,2) rectangle (5,3);
\fill[green!0!white] (4,1) rectangle (5,2);
\fill[green!0!white] (4,0) rectangle (5,1);
\fill[green!2!white] (5,10) rectangle (6,11);
\fill[green!0!white] (5,9) rectangle (6,10);
\fill[green!0!white] (5,8) rectangle (6,9);
\fill[green!0!white] (5,7) rectangle (6,8);
\fill[green!4!white] (5,6) rectangle (6,7);
\fill[green!0!white] (5,5) rectangle (6,6);
\fill[green!42!white] (5,4) rectangle (6,5);
\draw (5.5,4.5) node[align=center] {42};
\fill[green!3!white] (5,3) rectangle (6,4);
\fill[green!0!white] (5,2) rectangle (6,3);
\fill[green!1!white] (5,1) rectangle (6,2);
\fill[green!11!white] (5,0) rectangle (6,1);
\draw (5.5,0.5) node[align=center] {11};
\fill[green!0!white] (6,10) rectangle (7,11);
\fill[green!0!white] (6,9) rectangle (7,10);
\fill[green!0!white] (6,8) rectangle (7,9);
\fill[green!0!white] (6,7) rectangle (7,8);
\fill[green!2!white] (6,6) rectangle (7,7);
\fill[green!0!white] (6,5) rectangle (7,6);
\fill[green!11!white] (6,4) rectangle (7,5);
\draw (6.5,4.5) node[align=center] {11};
\fill[green!2!white] (6,3) rectangle (7,4);
\fill[green!0!white] (6,2) rectangle (7,3);
\fill[green!4!white] (6,1) rectangle (7,2);
\fill[green!14!white] (6,0) rectangle (7,1);
\draw (6.5,0.5) node[align=center] {14};
\fill[green!6!white] (7,10) rectangle (8,11);
\fill[green!4!white] (7,9) rectangle (8,10);
\fill[green!0!white] (7,8) rectangle (8,9);
\fill[green!0!white] (7,7) rectangle (8,8);
\fill[green!4!white] (7,6) rectangle (8,7);
\fill[green!0!white] (7,5) rectangle (8,6);
\fill[green!38!white] (7,4) rectangle (8,5);
\draw (7.5,4.5) node[align=center] {38};
\fill[green!22!white] (7,3) rectangle (8,4);
\draw (7.5,3.5) node[align=center] {22};
\fill[green!0!white] (7,2) rectangle (8,3);
\fill[green!84!white] (7,1) rectangle (8,2);
\draw (7.5,1.5) node[align=center] {84};
\fill[green!23!white] (7,0) rectangle (8,1);
\draw (7.5,0.5) node[align=center] {23};
\fill[green!8!white] (8,10) rectangle (9,11);
\fill[green!1!white] (8,9) rectangle (9,10);
\fill[green!0!white] (8,8) rectangle (9,9);
\fill[green!0!white] (8,7) rectangle (9,8);
\fill[green!1!white] (8,6) rectangle (9,7);
\fill[green!0!white] (8,5) rectangle (9,6);
\fill[green!1!white] (8,4) rectangle (9,5);
\fill[green!54!white] (8,3) rectangle (9,4);
\draw (8.5,3.5) node[align=center] {54};
\fill[green!0!white] (8,2) rectangle (9,3);
\fill[green!0!white] (8,1) rectangle (9,2);
\fill[green!0!white] (8,0) rectangle (9,1);
\fill[green!1!white] (9,10) rectangle (10,11);
\fill[green!0!white] (9,9) rectangle (10,10);
\fill[green!0!white] (9,8) rectangle (10,9);
\fill[green!0!white] (9,7) rectangle (10,8);
\fill[green!0!white] (9,6) rectangle (10,7);
\fill[green!0!white] (9,5) rectangle (10,6);
\fill[green!0!white] (9,4) rectangle (10,5);
\fill[green!10!white] (9,3) rectangle (10,4);
\draw (9.5,3.5) node[align=center] {10};
\fill[green!98!white] (9,2) rectangle (10,3);
\draw (9.5,2.5) node[align=center] {98};
\fill[green!0!white] (9,1) rectangle (10,2);
\fill[green!0!white] (9,0) rectangle (10,1);
\fill[green!1!white] (10,10) rectangle (11,11);
\fill[green!0!white] (10,9) rectangle (11,10);
\fill[green!0!white] (10,8) rectangle (11,9);
\fill[green!0!white] (10,7) rectangle (11,8);
\fill[green!1!white] (10,6) rectangle (11,7);
\fill[green!0!white] (10,5) rectangle (11,6);
\fill[green!2!white] (10,4) rectangle (11,5);
\fill[green!2!white] (10,3) rectangle (11,4);
\fill[green!0!white] (10,2) rectangle (11,3);
\fill[green!7!white] (10,1) rectangle (11,2);
\fill[green!49!white] (10,0) rectangle (11,1);
\draw (10.5,0.5) node[align=center] {49};
\draw[blue,very thick] (0,9) rectangle (1,10);
\draw[blue,very thick] (1,8) rectangle (2,9);
\draw[blue,very thick] (2,7) rectangle (3,8);
\draw[blue,very thick] (3,6) rectangle (4,7);
\draw[blue,very thick] (4,5) rectangle (5,6);
\draw[blue,very thick] (5,3) rectangle (6,4);
\draw[blue,very thick] (6,4) rectangle (7,5);
\draw[blue,very thick] (6,3) rectangle (7,4);
\draw[blue,very thick] (7,1) rectangle (8,2);
\draw[blue,very thick] (8,3) rectangle (9,4);
\draw[blue,very thick] (9,2) rectangle (10,3);
\draw[blue,very thick] (10,0) rectangle (11,1);
\end{tikzpicture}
&
\small
\begin{tikzpicture}[scale=0.5]
\node[label=above:\rotatebox{90}{das}] at (0.5,12) {};
\node[label=above:\rotatebox{90}{Verh\"altnis}] at (1.5,12) {};
\node[label=above:\rotatebox{90}{zwischen}] at (2.5,12) {};
\node[label=above:\rotatebox{90}{Obama}] at (3.5,12) {};
\node[label=above:\rotatebox{90}{und}] at (4.5,12) {};
\node[label=above:\rotatebox{90}{Netanyahu}] at (5.5,12) {};
\node[label=above:\rotatebox{90}{ist}] at (6.5,12) {};
\node[label=above:\rotatebox{90}{seit}] at (7.5,12) {};
\node[label=above:\rotatebox{90}{Jahren}] at (8.5,12) {};
\node[label=above:\rotatebox{90}{gespannt}] at (9.5,12) {};
\node[label=above:\rotatebox{90}{.}] at (10.5,12) {};
\draw (0,11.5) node[anchor=east] {the};
\draw (0,10.5) node[anchor=east] {relationship};
\draw (0,9.5) node[anchor=east] {between};
\draw (0,8.5) node[anchor=east] {Obama};
\draw (0,7.5) node[anchor=east] {and};
\draw (0,6.5) node[anchor=east] {Netanyahu};
\draw (0,5.5) node[anchor=east] {has};
\draw (0,4.5) node[anchor=east] {been};
\draw (0,3.5) node[anchor=east] {stretched};
\draw (0,2.5) node[anchor=east] {for};
\draw (0,1.5) node[anchor=east] {years};
\draw (0,0.5) node[anchor=east] {.};
\fill[green!1!white] (0,11) rectangle (1,12);
\fill[green!1!white] (0,10) rectangle (1,11);
\fill[green!6!white] (0,9) rectangle (1,10);
\fill[green!1!white] (0,8) rectangle (1,9);
\fill[green!1!white] (0,7) rectangle (1,8);
\fill[green!0!white] (0,6) rectangle (1,7);
\fill[green!3!white] (0,5) rectangle (1,6);
\fill[green!2!white] (0,4) rectangle (1,5);
\fill[green!1!white] (0,3) rectangle (1,4);
\fill[green!2!white] (0,2) rectangle (1,3);
\fill[green!0!white] (0,1) rectangle (1,2);
\fill[green!11!white] (0,0) rectangle (1,1);
\draw (0.5,0.5) node[align=center] {11};
\fill[green!47!white] (1,11) rectangle (2,12);
\draw (1.5,11.5) node[align=center] {47};
\fill[green!7!white] (1,10) rectangle (2,11);
\fill[green!4!white] (1,9) rectangle (2,10);
\fill[green!0!white] (1,8) rectangle (2,9);
\fill[green!0!white] (1,7) rectangle (2,8);
\fill[green!0!white] (1,6) rectangle (2,7);
\fill[green!0!white] (1,5) rectangle (2,6);
\fill[green!0!white] (1,4) rectangle (2,5);
\fill[green!3!white] (1,3) rectangle (2,4);
\fill[green!3!white] (1,2) rectangle (2,3);
\fill[green!0!white] (1,1) rectangle (2,2);
\fill[green!3!white] (1,0) rectangle (2,1);
\fill[green!8!white] (2,11) rectangle (3,12);
\fill[green!81!white] (2,10) rectangle (3,11);
\draw (2.5,10.5) node[align=center] {81};
\fill[green!7!white] (2,9) rectangle (3,10);
\fill[green!0!white] (2,8) rectangle (3,9);
\fill[green!0!white] (2,7) rectangle (3,8);
\fill[green!0!white] (2,6) rectangle (3,7);
\fill[green!0!white] (2,5) rectangle (3,6);
\fill[green!0!white] (2,4) rectangle (3,5);
\fill[green!0!white] (2,3) rectangle (3,4);
\fill[green!3!white] (2,2) rectangle (3,3);
\fill[green!0!white] (2,1) rectangle (3,2);
\fill[green!2!white] (2,0) rectangle (3,1);
\fill[green!6!white] (3,11) rectangle (4,12);
\fill[green!5!white] (3,10) rectangle (4,11);
\fill[green!72!white] (3,9) rectangle (4,10);
\draw (3.5,9.5) node[align=center] {72};
\fill[green!7!white] (3,8) rectangle (4,9);
\fill[green!1!white] (3,7) rectangle (4,8);
\fill[green!0!white] (3,6) rectangle (4,7);
\fill[green!1!white] (3,5) rectangle (4,6);
\fill[green!0!white] (3,4) rectangle (4,5);
\fill[green!0!white] (3,3) rectangle (4,4);
\fill[green!2!white] (3,2) rectangle (4,3);
\fill[green!0!white] (3,1) rectangle (4,2);
\fill[green!1!white] (3,0) rectangle (4,1);
\fill[green!4!white] (4,11) rectangle (5,12);
\fill[green!3!white] (4,10) rectangle (5,11);
\fill[green!2!white] (4,9) rectangle (5,10);
\fill[green!87!white] (4,8) rectangle (5,9);
\draw (4.5,8.5) node[align=center] {87};
\fill[green!0!white] (4,7) rectangle (5,8);
\fill[green!2!white] (4,6) rectangle (5,7);
\fill[green!0!white] (4,5) rectangle (5,6);
\fill[green!0!white] (4,4) rectangle (5,5);
\fill[green!0!white] (4,3) rectangle (5,4);
\fill[green!0!white] (4,2) rectangle (5,3);
\fill[green!0!white] (4,1) rectangle (5,2);
\fill[green!0!white] (4,0) rectangle (5,1);
\fill[green!0!white] (5,11) rectangle (6,12);
\fill[green!0!white] (5,10) rectangle (6,11);
\fill[green!0!white] (5,9) rectangle (6,10);
\fill[green!0!white] (5,8) rectangle (6,9);
\fill[green!93!white] (5,7) rectangle (6,8);
\draw (5.5,7.5) node[align=center] {93};
\fill[green!1!white] (5,6) rectangle (6,7);
\fill[green!1!white] (5,5) rectangle (6,6);
\fill[green!0!white] (5,4) rectangle (6,5);
\fill[green!0!white] (5,3) rectangle (6,4);
\fill[green!0!white] (5,2) rectangle (6,3);
\fill[green!0!white] (5,1) rectangle (6,2);
\fill[green!1!white] (5,0) rectangle (6,1);
\fill[green!0!white] (6,11) rectangle (7,12);
\fill[green!0!white] (6,10) rectangle (7,11);
\fill[green!0!white] (6,9) rectangle (7,10);
\fill[green!1!white] (6,8) rectangle (7,9);
\fill[green!0!white] (6,7) rectangle (7,8);
\fill[green!95!white] (6,6) rectangle (7,7);
\draw (6.5,6.5) node[align=center] {95};
\fill[green!1!white] (6,5) rectangle (7,6);
\fill[green!0!white] (6,4) rectangle (7,5);
\fill[green!0!white] (6,3) rectangle (7,4);
\fill[green!0!white] (6,2) rectangle (7,3);
\fill[green!0!white] (6,1) rectangle (7,2);
\fill[green!0!white] (6,0) rectangle (7,1);
\fill[green!2!white] (7,11) rectangle (8,12);
\fill[green!0!white] (7,10) rectangle (8,11);
\fill[green!1!white] (7,9) rectangle (8,10);
\fill[green!0!white] (7,8) rectangle (8,9);
\fill[green!0!white] (7,7) rectangle (8,8);
\fill[green!0!white] (7,6) rectangle (8,7);
\fill[green!38!white] (7,5) rectangle (8,6);
\draw (7.5,5.5) node[align=center] {38};
\fill[green!21!white] (7,4) rectangle (8,5);
\draw (7.5,4.5) node[align=center] {21};
\fill[green!5!white] (7,3) rectangle (8,4);
\fill[green!4!white] (7,2) rectangle (8,3);
\fill[green!0!white] (7,1) rectangle (8,2);
\fill[green!9!white] (7,0) rectangle (8,1);
\fill[green!17!white] (8,11) rectangle (9,12);
\draw (8.5,11.5) node[align=center] {17};
\fill[green!0!white] (8,10) rectangle (9,11);
\fill[green!1!white] (8,9) rectangle (9,10);
\fill[green!0!white] (8,8) rectangle (9,9);
\fill[green!0!white] (8,7) rectangle (9,8);
\fill[green!0!white] (8,6) rectangle (9,7);
\fill[green!16!white] (8,5) rectangle (9,6);
\draw (8.5,5.5) node[align=center] {16};
\fill[green!14!white] (8,4) rectangle (9,5);
\draw (8.5,4.5) node[align=center] {14};
\fill[green!8!white] (8,3) rectangle (9,4);
\fill[green!38!white] (8,2) rectangle (9,3);
\draw (8.5,2.5) node[align=center] {38};
\fill[green!8!white] (8,1) rectangle (9,2);
\fill[green!19!white] (8,0) rectangle (9,1);
\draw (8.5,0.5) node[align=center] {19};
\fill[green!4!white] (9,11) rectangle (10,12);
\fill[green!0!white] (9,10) rectangle (10,11);
\fill[green!0!white] (9,9) rectangle (10,10);
\fill[green!0!white] (9,8) rectangle (10,9);
\fill[green!0!white] (9,7) rectangle (10,8);
\fill[green!0!white] (9,6) rectangle (10,7);
\fill[green!8!white] (9,5) rectangle (10,6);
\fill[green!5!white] (9,4) rectangle (10,5);
\fill[green!2!white] (9,3) rectangle (10,4);
\fill[green!33!white] (9,2) rectangle (10,3);
\draw (9.5,2.5) node[align=center] {33};
\fill[green!90!white] (9,1) rectangle (10,2);
\draw (9.5,1.5) node[align=center] {90};
\fill[green!32!white] (9,0) rectangle (10,1);
\draw (9.5,0.5) node[align=center] {32};
\fill[green!3!white] (10,11) rectangle (11,12);
\fill[green!0!white] (10,10) rectangle (11,11);
\fill[green!2!white] (10,9) rectangle (11,10);
\fill[green!0!white] (10,8) rectangle (11,9);
\fill[green!0!white] (10,7) rectangle (11,8);
\fill[green!0!white] (10,6) rectangle (11,7);
\fill[green!26!white] (10,5) rectangle (11,6);
\draw (10.5,5.5) node[align=center] {26};
\fill[green!54!white] (10,4) rectangle (11,5);
\draw (10.5,4.5) node[align=center] {54};
\fill[green!77!white] (10,3) rectangle (11,4);
\draw (10.5,3.5) node[align=center] {77};
\fill[green!12!white] (10,2) rectangle (11,3);
\draw (10.5,2.5) node[align=center] {12};
\fill[green!0!white] (10,1) rectangle (11,2);
\fill[green!17!white] (10,0) rectangle (11,1);
\draw (10.5,0.5) node[align=center] {17};
\draw[blue,very thick] (0,11) rectangle (1,12);
\draw[blue,very thick] (1,10) rectangle (2,11);
\draw[blue,very thick] (2,9) rectangle (3,10);
\draw[blue,very thick] (3,8) rectangle (4,9);
\draw[blue,very thick] (4,7) rectangle (5,8);
\draw[blue,very thick] (5,6) rectangle (6,7);
\draw[blue,very thick] (6,5) rectangle (7,6);
\draw[blue,very thick] (7,4) rectangle (8,5);
\draw[blue,very thick] (7,2) rectangle (8,3);
\draw[blue,very thick] (8,1) rectangle (9,2);
\draw[blue,very thick] (9,3) rectangle (10,4);
\draw[blue,very thick] (10,0) rectangle (11,1);
\end{tikzpicture}\\[2mm]
(a) Desired Alignment & (b) Mismatched Alignment
\end{tabular}
\vspace{3mm}
\caption{Word alignment for English--German: comparing the attention model states (green boxes with probability in percent if over 10) with alignments obtained from fast-align (blue outlines).}
\label{fig:good-and-flakyword-alignment-matrix}
\end{figure}

See Figure~\ref{fig:good-and-flakyword-alignment-matrix}a for an illustration. We compare the word attention states (green boxes) with the word alignments obtained with fast align (blue outlines). For most words, these match up pretty well. Both attention states and fast-align alignment points are a bit fuzzy around the function words {\em have-been/sind}.

However, the attention model may settle on alignments that do not correspond with our intuition or alignment points obtained with fast-align. See Figure~\ref{fig:good-and-flakyword-alignment-matrix}b for the reverse language direction, German--English. All the alignment points appear to be off by one position. We are not aware of any intuitive explanation for this divergent behavior --- the translation quality is high for both systems.

We measure how well the soft alignment (attention model) of the neural machine translation system match the alignments of fast-align with two metrics:
\begin{itemize}
\item a {\bf match score}\index{match score (alignment)} that checks for each output if the aligned input word according to fast-align is indeed the input word that received the highest attention probability, and
\item a {\bf probability mass score}\index{probability mass score (alignment)} that sums up the probability mass given to each alignment point obtained from fast-align.
\end{itemize}
In these scores, we have to handle byte pair encoding and many-to-many alignments\footnote{(1) neural machine translation operates on subwords, but fast-align is run on full words. (2) If an input word is split into subwords by byte pair encoding, then we add their attention scores. (3) If an output word is split into subwords, then we take the average of their attention vectors. (4) The match scores and probability mass scores are computed as average over output word-level scores. (5) If an output word has no fast-align alignment point, it is ignored in this computation. (6) If an output word is fast-aligned to multiple input words, then (6a) for the match score: count it as correct if the $n$ aligned words among the top $n$ highest scoring words according to attention and (6b) for the probability mass score: add up their attention scores.}

In out experiment, we use the neural machine translation models provided by Edinburgh\footnote{\tt https://github.com/rsennrich/wmt16-scripts} \citep{sennrich-haddow-birch:2016:WMT}. We run fast-align on the same parallel data sets to obtain alignment models and used them to align the input and output of the neural machine translation system.
Table~\ref{tab:word-alignment-scores} shows alignment scores for the systems. The results suggest that, while drastic, the divergence for German--English is an outlier. We note, however, that we have seen such large a divergence also under different data conditions.

\begin{table}
\begin{center}
\begin{tabular}{l|c|c}
\bf Language Pair & Match & Prob. \\ \hline
German--English & 14.9\% & 16.0\%\\
English--German & 77.2\% & 63.2\% \\
Czech--English & 78.0\% & 63.3\% \\
English--Czech & 76.1\% & 59.7\% \\
Russian--English & 72.5\% & 65.0\% \\
English--Russian & 73.4\% & 64.1\% \\
\end{tabular}
\end{center}
\caption{Scores indicating overlap between attention probabilities and alignments obtained with fast-align.}
\label{tab:word-alignment-scores}
\end{table}

Note that the attention model may produce better word alignments by guided alignment training \citep{DBLP:journals/corr/ChenMKP16,liu-EtAl:2016:COLING} where supervised word alignments (such as the ones produced by fast-align) are provided to model training.

\subsection{Beam Search}\index{beam search!evaluation}
\begin{figure}
\input{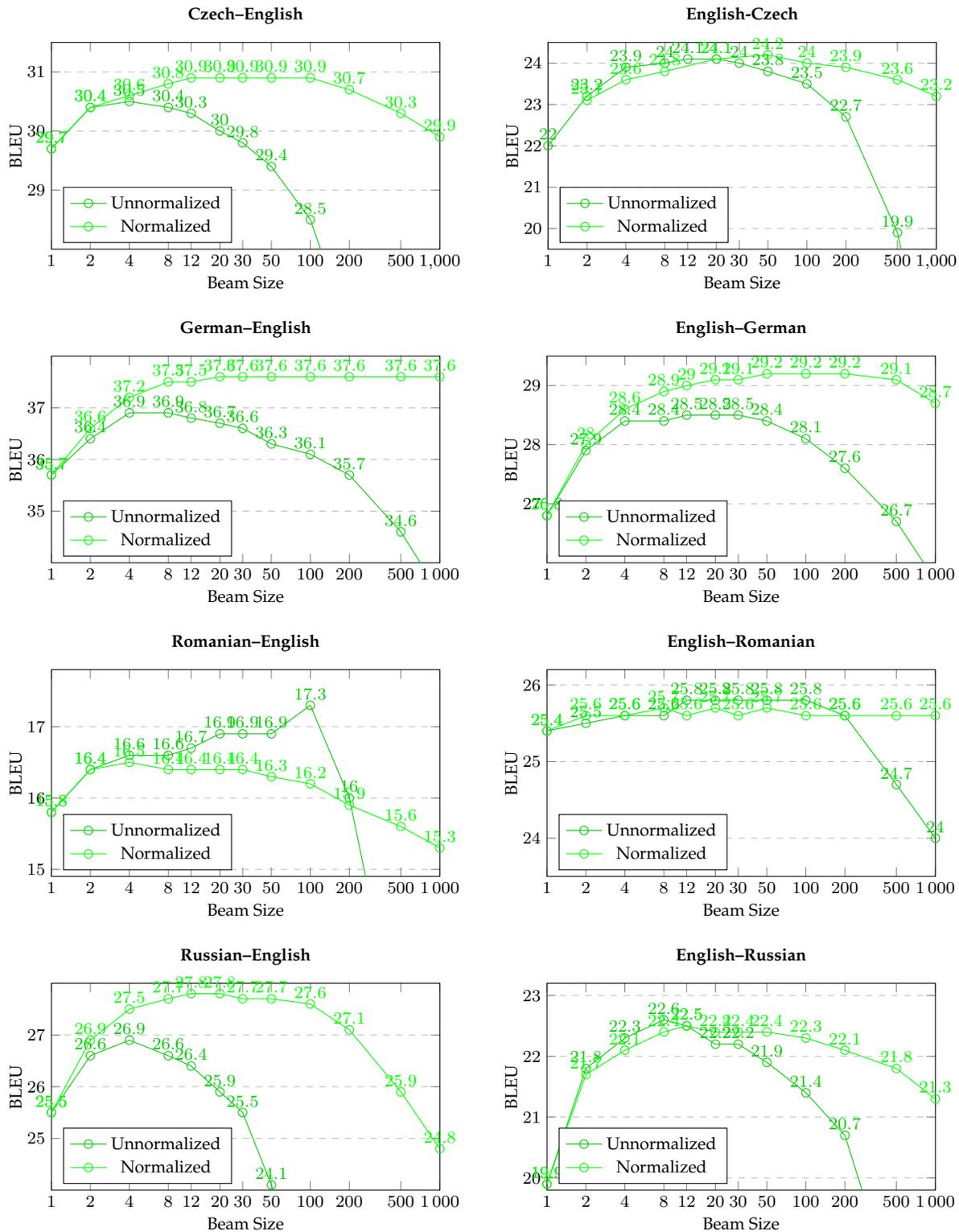}
\vspace{2mm}
\caption{Translation quality with varying beam sizes. For large beams, quality decreases, especially when not normalizing scores by sentence length.}
\label{fig:beam-search}
\end{figure}

The task of decoding is to find the full sentence translation with the highest probability. In statistical machine translation, this problem has been addressed with heuristic search techniques that explore a subset of the space of possible translation. A common feature of these search techniques is a beam size parameter that limits the number of partial translations maintained per input word.

There is typically a straightforward relationship between this beam size parameter and the model score of resulting translations and also their quality score (e.g., BLEU). While there are diminishing returns for increasing the beam parameter, typically improvements in these scores can be expected with larger beams.

Decoding in neural translation models can be set up in similar fashion. When predicting the next output word, we may not only commit to the highest scoring word prediction but also maintain the next best scoring words in a list of partial translations. We record with each partial translation the word translation probabilities (obtained from the softmax), extend each partial translation with subsequent word predictions and accumulate these scores. Since the number of partial translation explodes exponentially with each new output word, we prune them down to a beam of highest scoring partial translations.

As in traditional statistical machine translation decoding, increasing the beam size allows us to explore a larger set of the space of possible translation and hence find translations with better model scores.

However, as Figure~\ref{fig:beam-search} illustrates, increasing the beam size does not consistently improve translation quality. In fact, in almost all cases, worse translations are found beyond an optimal beam size setting (we are using again Edinburgh's WMT 2016 systems). The optimal beam size varies from 4 (e.g., Czech--English) to around 30 (English--Romanian).

Normalizing sentence level model scores by length of the output alleviates the problem somewhat and also leads to better optimal quality in most cases (5 of the 8 language pairs investigated). Optimal beam sizes are in the range of 30--50 in almost all cases, but quality still drops with larger beams. The main cause of deteriorating quality are shorter translations under wider beams.

\subsection{Further Readings}

Other studies have looked at the comparable performance of neural and statistical machine translation systems. \citet{bentivogli-EtAl:2016:EMNLP2016} considered different linguistic categories for English--German and \citet{toral-sanchezcartagena:2017:EACLlong} compared different broad aspects such as fluency and reordering for nine language directions.

\section{Additional Topics}

\furtherreadings{
Especially early work on neural networks for machine translation was aimed at building neural components to be used in traditional statistical machine translation systems.

\paragraph{Translation Models} By including aligned source words in the conditioning context, \cite{devlin-EtAl:2014:P14-1} enrich a feed-forward neural network language model with source context
\cite{JiajunZhang:2015:ijcai} add a sentence embedding to the conditional context of this model, which are learned using a variant of convolutional neural networks and mapping them across languages. \cite{meng-EtAl:2015:ACL-IJCNLP} use a more complex convolutional neural network to encode the input sentence that uses gated layers and also incorporates information about the output context.

\paragraph{Reordering Models} Lexicalized reordering models struggle with sparse data problems when conditioned on rich context. \cite{li-EtAl:2014:Coling3} show that a neural reordering model can be conditioned on current and previous phrase pair (encoded with a recursive neural network auto-encoder) to make the same classification decisions for orientation type.

\paragraph{Pre-Ordering} Instead of handing reordering within the decoding process, we may pre-order the input sentence into output word order. 
\cite{degispert-iglesias-byrne:2015:NAACL-HLT} use an input dependency tree to learn a model that swaps children nodes and implement it using a feed-forward neural network. \cite{micelibarone-attardi:2015:ACL-IJCNLP} formulate a top-down left-to-right walk through the dependency tree and make reordering decisions at any node. They model this process with a recurrent neural network that includes past decisions in the conditioning context.

\paragraph{N-Gram Translation Models} An alternative view of the phrase based translation model is to break up phrase translations into minimal translation units, and employing a n-gram model over these units to condition each minimal translation units on the previous ones. \cite{schwenk-rcostajussa-rfonollosa:2007:EMNLP-CoNLL2007} treat each minimal translation unit as an atomic symbol and train a neural language model over it. Alternatively, \citep{hu-EtAl:2014:EACL} represent the minimal translation units as bag of words, \citep{wu-EtAl:2014:EMNLP2014} break them even further into single input words, single output words, or single input-output word pairs, and \cite{yu-zhu:2015:ACL-IJCNLP} use phrase embeddings leaned with an auto-encoder.}

\addcontentsline{toc}{section}{Bibliography}
\small
\bibliographystyle{acl_natbib} 
\bibliography{mt,more} 

\newpage
\chapter*{Author Index}
\addcontentsline{toc}{section}{Author Index}
\markboth{Author Index}{Author Index}
{\small 
\renewcommand{\baselinestretch}{0.95}
\begin{multicols}{2}
\printauthorindex
\end{multicols} }
\renewcommand{\baselinestretch}{1.1}

\addcontentsline{toc}{section}{Index}
\printindex

\end{document}